\documentclass{article}
\pdfoutput=1
\usepackage{iclr2027_conference,times}

\usepackage{amsmath,amsfonts,bm}

\def\eqref#1{equation~\ref{#1}}

\def\1{\bm{1}}

\DeclareMathAlphabet{\mathsfit}{\encodingdefault}{\sfdefault}{m}{sl}
\SetMathAlphabet{\mathsfit}{bold}{\encodingdefault}{\sfdefault}{bx}{n}

\newcommand{\Ptrain}{\hat{P}_{\rm{data}}}

\newcommand{\chg}[1]{\textcolor{black}{#1}}

\usepackage{hyperref}
\usepackage{url}
\usepackage{graphicx}
\usepackage{wrapfig}
\usepackage{cleveref}
\makeatletter
\AddToHook{cmd/appendix/before}{\crefalias{section}{appendix}}
\makeatother

\usepackage{multirow}
\usepackage{enumitem}
\usepackage{wrapfig}
\usepackage{booktabs}
\usepackage{tikz}
\usetikzlibrary{arrows.meta}
\usetikzlibrary{matrix}

\definecolor{mynavy}{HTML}{1D6996}

\newcommand{\Molmo}{\texttt{MolmoSpaces}}

\hypersetup{
    colorlinks=true,
    citecolor=mynavy,
    linkcolor=mynavy,
    urlcolor=mynavy,
    filecolor=black
}

\title{Reinforcement Learning Enables Scaling \\ Generalist Robot Policy Improvement}

\author{%
\mbox{%
Michal Korniak$^{1,*}$\hspace{1.5em}%
Marco Bagatella$^{1,2,*}$\hspace{1.5em}%
Ralf R\"omer$^{3,*}$} \BIGAND%
\mbox{%
Michal Nauman$^{4,5}$\hspace{1.5em}%
Yarden As$^{1}$\hspace{1.5em}%
Pranav Vignesh$^{6}$\hspace{1.5em}%
Manan Agarwal}$^{6}$ \AND
\mbox{%
Leander Diaz-Bone$^{1,2}$\hspace{1.5em}%
Piotr Zalewski$^{5}$\hspace{1.5em}%
Miko{\l}aj Czarnecki$^{4}$} \AND
\mbox{%
Lenart Treven$^{1}$\hspace{1.5em}%
Shashwat Saxena$^{6}$\hspace{1.5em}%
Sreyas Venkataraman$^{6}$} \AND
\mbox{%
Sarvesh Patil$^{6}$\hspace{1.5em}%
Zhiyuan Zhou$^{7}$\hspace{1.5em}%
Daniel Marta$^{1}$} \BIGAND
\mbox{%
Oier Mees$^{1, 8, \dagger}$\hspace{1.5em}%
Nur Muhammad Mahi Shafiullah$^{7,\dagger}$\hspace{1.5em}%
Carmelo Sferrazza$^{7, \dagger}$} \AND
\mbox{%
Bhavya Sukhija$^{1, \dagger}$\hspace{1.5em}%
Angela P. Schoellig$^{3, \dagger}$\hspace{1.5em}%
Andreas Krause$^{1,\dagger}$} \BIGAND
\mbox{%
Max Simchowitz$^{6, 9, \ddagger}$\hspace{1.5em}%
Andrew Wagenmaker$^{7, 10, \ddagger}$} \\
\\
\small
$^{1}$ETH Zurich \hspace{2em} 
$^{2}$MPI for Intelligent Systems, Tübingen \hspace{2em} 
$^{3}$Technical University of Munich \\ \small 
$^{4}$University of Warsaw \hspace{2em} 
$^{5}$Nomagic \hspace{2em} 
$^{6}$Carnegie Mellon University \hspace{2em}
$^{7}$UC Berkeley \hspace{2em} \\ \small 
$^{8}$Microsoft \hspace{2em}
$^{9}$Amazon FAR \hspace{2em}
$^{10}$Physical Intelligence
\\ \\
\small $^{*}$Co-First Authors, order by coin flip \hspace{1em} $^{\dagger}$Co-Principal Investigators \hspace{1em} $^{\ddagger}$Lead Principal Investigators \\ \\
Project page: \href{https://lasgroup.github.io/generalist-improvement}{https://lasgroup.github.io/generalist-improvement}
 \\
Repository: \href{https://github.com/lasgroup/generalist-improvement}{https://github.com/lasgroup/generalist-improvement}
}

\newcommand{\condon}{\,|\,}

\iclrfinalcopy
\begin{document}

\maketitle

\begin{abstract}
The introduction of generalist robot policies has enabled a single, large model to learn to perform a variety of tasks from a vast pre-training corpus.
While these models often exhibit powerful zero-shot generalization capabilities, they typically require an adaptation phase to achieve sufficient performance in deployment.
Iterated self-improvement via reinforcement learning (RL) has demonstrated promise in enabling this adaptation, yet such approaches have been largely explored in \emph{single-task} settings -- where the goal is to simply improve on a narrow deployment task -- which lies in stark contrast to the \emph{multi-task} capabilities of the base policy.
In this work we ask whether generalist robot policies are capable of multi-task, \emph{generalist improvement in deployment} -- targeting many diverse tasks simultaneously -- and what algorithmic recipe enables this.
To answer this question, we first develop a set of structured evaluations to test generalist improvement capabilities, and systematically evaluate algorithmic and practical design choices regarding value estimation, policy extraction and resource allocation.
We show that not only can generalist policies be improved online across diverse tasks, but that the right recipe can enable \emph{generalization} and increase performance on tasks not considered during online training.
We characterize when this generalization occurs and show that both the training task distribution and the use of value-based RL are critical.
To the best of our knowledge, this is \chg{among} the first evidence for online generalist improvement for robot foundation models.
\end{abstract}

\newpage
\section{Introduction}

\begin{wrapfigure}[22]{r}{0.38\linewidth}
  \vspace{-11mm}
  \includegraphics[width=\linewidth]{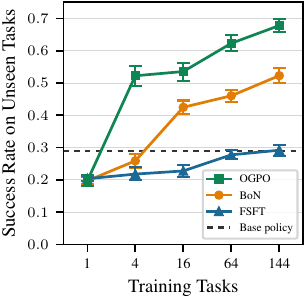}
  \vspace{-8mm}
  \caption{\textbf{Scaling the number of RL training tasks enables generalist improvement of generalist robot policies.} As training task diversity increases, critic-guided \chg{policy extraction (OGPO) or} refinement (BoN) prevent forgetting and improve performance on unseen tasks on \texttt{MolmoSpaces}, unlike filtered supervised fine-tuning (FSFT).\looseness -1}
  \label{fig:hook}
\end{wrapfigure}

\looseness=-1 Generalist robot policies, particularly Vision-Language-Action Models (VLAs), have shown promise in enabling general-purpose robotic capabilities, allowing a single policy to be deployed across diverse tasks~\citep{brohan2023rt1roboticstransformerrealworld, octo_2023, black2024pi_0, gr00tn1_2025}.
However, despite encoding diverse task-solving behaviors, such policies often lack sufficient reliability for robust deployment, and a degree of adaptation is typically necessary for effective performance~\citep{kim2025finetuningvisionlanguageactionmodelsoptimizing, intelligence2025pi_, reuss2025flower}. While enabling this adaptation through human supervision is often prohibitively expensive, autonomous adaptation through reinforcement learning (RL) has emerged as a promising approach to scale policy improvement, enabling robots to learn from their own experience in deployment \citep{intelligence2025pi}.
Yet, such autonomous improvement is typically limited to narrow \emph{single-task} settings. \chg{As such, while these approaches enable generalist policies to improve, they may also specialize this improvement to the particular deployment task considered, and achieving improved performance on additional tasks of interest generally requires further online interaction and fine-tuning.}\looseness -1

A generalist policy, however, can produce a wide range of behaviors across diverse tasks, even if these behaviors may be far from optimal. Ideally we might hope that, by deploying on a diverse range of tasks, we could improve the performance of the policy \emph{across all tasks simultaneously} -- boosting its performance not just on a single deployment task, but on the entire distribution of tasks of interest.
\chg{In particular, we might hope to enable truly \emph{generalist improvement}: improvement on both the diverse set of tasks we deploy on, as well as \emph{generalization} beyond these deployment tasks and improved performance on future novel deployment settings.}
This motivates the central question we consider in this work: \textbf{\emph{how can we leverage interaction across a diverse set of tasks to enable online generalist improvement of generalist policies?}}
That is, can a careful adaptation phase over multiple tasks enable both (i) improvement across these tasks simultaneously, and (ii) generalization beyond them, improving performance on tasks not encountered during post-training? \looseness -1

In this paper, we undertake a systematic study to \textbf{quantify generalist policy improvement} across different axes of generalization. We show that appropriate recipes, when instantiated carefully, can enable not only generalist policy improvement across a diverse set of training tasks, but also \textbf{task-generalization} -- improvement on tasks not seen during training.
Further, we show that this improvement \emph{scales} with the number and diversity of tasks trained on, as we showcase in \Cref{fig:hook}.
\chg{Our results suggest that moving beyond the single-task paradigm is not only possible but also \emph{crucial} for achieving broader, generalist improvement.}

We isolate the \textbf{essential algorithmic ingredients} that enable multi-task learning and generalization. 
\chg{Specifically, we compare three families of learning approaches:
\begin{itemize}
    \item[(a)] \textbf{\emph{Direct Policy Adaptation}}, which directly improves the policy via filtered supervised fine-tuning on successful trajectories \citep[FSFT,][]{peters2007reinforcement, zhou2024autonomous, qin2025supervised}. \item[(b)] \textbf{\emph{Critic-Guided Test-Time Refinement}}, which uses test-time optimization against a learned state-action critic to improve actions. Here, we focus on Best-of-N sampling ~\citep[BoN,][]{nakamoto2025steering, giridhar2026imitation} with respect to that critic's induced action ranking as a representative recipe. 
    \item[(c)]\textbf{\emph{Critic-Guided Policy Extraction}}, which updates a policy so as to maximize the value the critic assigns to the actions produced. Here, we adopt off-policy generative policy optimization \citep[OGPO,][]{patil2026ogpo} as a representative recipe, while also ablating alternative approaches such as  combining FSFT and BoN sampling, or using AWR-inspired policy updates \citep{peng2019advantage}.
\end{itemize}
}
\chg{Overall, while all three approaches enable some degree of generalization given a sufficient diversity in training tasks, we find that \emph{critic-guided methods enable a substantially greater degree of task generalization than direct policy adaptation alone}. In particular, we show that critic-guided methods afford generalization both across training scenes, and, perhaps more surprisingly, across object types. 
Furthermore, we find that critic-guided policy extraction (OGPO) significantly outperforms test-time refinement (BoN), while also exhibiting similar scaling trends with training task diversity.
This finding further points to \emph{critic-learning} as the key component that enables generalist robotic policy improvement as task diversity scales. 
}

\chg{In addition to investigating algorithmic choices, we also systematically evaluate the \textbf{training task distribution} necessary for achieving such generalist improvement. We show that both the number of distinct training tasks, as well as the type of diversity exhibited in these training tasks, is critical to achieving generalist improvement. In particular, we find that a high degree of diversity in the required solution strategies within the training task distribution is crucial to learning generalizable task-solving behaviors.}

\chg{We believe these results illustrate a path forward for efficient online improvement of generalist robot policies. Rather than expensive single-task improvement on each deployment task of interest, our results show that, by amortizing this cost across a diverse set of tasks, one can enable generalist improvement, achieving improved performance on future deployment tasks of interest without paying the cost of additional online fine-tuning.}

\section{Related Work}

\textbf{VLA post-training.\;}
\looseness=-1
Diffusion and flow-matching provide expressive generative policy classes that, when trained through behavior cloning at sufficient scale and diversity, can form the basis of generalist policies~\citep{janner2022planning,chi2023diffusion,reed2022a}. Recent \chg{RL fine-tuning} methods directly optimize such policies~\citep{psenka2024qsm,park2025flow,wagenmaker2025steering,patil2026ogpo,li2026q}, but scaling them to large VLAs remains challenging. Existing VLA post-training therefore often relies on simpler, readily scalable methods. One family directly updates the policy using high-quality experience, including filtered supervised fine-tuning~\citep{qin2025supervised,yang2026aloe}, conditioned behavior cloning~\citep{intelligence2025pi}, and soft variants that weight demonstrations by estimated value~\citep{peng2019advantage,su2026ig}. A complementary family leaves the policy unchanged and instead uses learned value functions for test-time action selection, such as Best-of-N~\citep{nakamoto2025steering,dodeja2026life}\chg{, or by learning residual policies to modify the output action \citep{ankile2024imitation,dong2026expostablereinforcementlearning,xiao2026self} or steer the pre-trained policy's denoising process \citep{wagenmaker2025steering}.} We likewise focus on simple, scalable techniques, demonstrating~(cf. \Cref{sec:scaling}) that they can achieve competitive performance.
\chg{While these existing approaches to VLA post-training have demonstrated performance gains~\citep{zhang2024grape,hu2024flare,ankile2024imitation,myers2024policy,li2025grrl,chen2025conrft,dong2025matters,guo2025improving,lu2025vlarl,sun2026prior}, they are limited, primarily, to improvement on  single, individual tasks~\citep{xu2026rl,pan2026sop,gr00tn1_2025,xiaomiroboticsteam2026xiaomirobotics1scalingvisionlanguageactionmodels}.}
In contrast, we study joint post-training across increasingly broad task distributions and ask whether the resulting policy also improves on tasks outside the training distribution.

\textbf{Multi-task generalization.\;}
\looseness=-1
A separate literature studies how robotic performance scales with the diversity of training tasks. Prior work considers multi-task scaling without foundation-model policies~\citep{nauman2026bigger}, while meta-reinforcement learning instead develops algorithms that \emph{learn to rapidly adapt} to unseen tasks (rather than to directly transfer a fixed policy to new tasks, the setting we consider), traditionally with relatively small policies~\citep{duan2016rl2fastreinforcementlearning,finn2017modelagnosticmetalearningfastadaptation,yu2021metaworldbenchmarkevaluationmultitask}. More recent work extends this perspective to VLA policies through behavior cloning~\citep{bianchi2026roboticpolicyadaptationweightspace}. Other works study generalization arising from scaling behavior cloning during pretraining~\citep{zheng2026egoscale,dyna2026dyna2} or post-training~\citep{lin2025data,lin2026dygro}. Here we focus on self-improvement via reinforcement learning. Perhaps closest to our setting is~\citet{liu2026can}, who compare generalization of offline \chg{SFT from expert demonstrations} with on-policy reinforcement-learning algorithms such as PPO~\citep{schulman2017proximal} and GRPO~\citep{shao2024deepseekmath}. Our study differs in two important ways. 
\chg{First, \citet{liu2026can} utilize a single fixed distribution over training environments for all experiments; in contrast, a primary contribution of our work is quantifying which training distributions are necessary to induce generalization. Second, while we consider standard online RL fine-tuning procedures commonly used for VLAs, and study which such fine-tuning procedures enable generalization, this work compares non-RL approaches (SFT on expert demonstrations) to on-policy RL. While \citet{liu2026can} show that RL fine-tuning can enable greater generalization than SFT, our work provides informative guidance as to which design decisions are key to enabling this in practice.
We note as well that multi-task generalization in post-training} has been long studied in LLM post-training. We refer to Appendix~\ref{app:extended_related_work} where we draw parallels to this work.\looseness -1

\renewcommand{\Ptrain}{P_{\mathrm{train}}}
\newcommand{\Ptest}{P_{\mathrm{test}}}

\section{Problem Setting and Experimental Methodology}
\label{sec:method}

\subsection{Problem Setting: Generalist Improvement of Generalist Policies}
\label{sec:setting}
We model our environment as a contextual Markov decision process $\mathcal{M}=(\mathcal{S}, \mathcal{A}, \mathcal{C}, P, R, \mu_0, \gamma)$, where $\mathcal{S}$ and $\mathcal{A}$ are continuous state and action spaces, $\mathcal{C}$ is a space of contexts encoding different tasks\footnote{In practice, a task will correspond to following a specific prompt in a specific manipulation environment.}, $P: \mathcal{S \times A \times C} \to \Delta(\mathcal{S})$ is a stochastic transition function, $R: \mathcal{S} \times \mathcal{A} \times \mathcal{C} \to [R_{\min}, R_{\max}]$ is a context-conditional reward function specifying each task, $\mu_0: \mathcal{C} \to \Delta(\mathcal{S})$ is an initial distribution and $\gamma \in [0, 1)$ is a discount factor. We model a generalist agent as a context-conditional policy $\pi: \mathcal{S} \times \mathcal{C} \to \Delta(\mathcal{A})$\footnote{We overload notation and condition $\pi$ on $c \in \mathcal{C}$ for simplicity; in practice, if $\pi$ is a VLA, the task conditioning is provided in the form of a language command, and the policy outputs action chunks.}, and evaluate its \chg{expected return} from a state-action pair $(s,a)$ and task $c$ according to its Q-function:
\begin{equation}
    \textstyle Q^\pi(s,a,c) = \mathbb{E}_{P,\pi,c} \left[\sum_{t \geq 0} \gamma^t R(s_t, a_t,  c) \mid s_0 = s, a_0 = a\right],  \;\; 
    \label{eq:q}
\end{equation}
where above, $\mathbb{E}_{P,\pi,c}$ denotes expectation with respect to $a_t \sim \pi(\cdot \condon s_t, c)\;\text{and}\; s_{t+1} \sim P(\cdot\condon s_t, a_t,c)$. We denote $V^\pi(s,c)=\mathbb{E}_{a \sim \pi(\cdot\condon s,c)}[Q^\pi(s,a,c)]$.

\chg{In this work, we are interested in enabling improvement on tasks from some test distribution $\rho_\mathrm{test} \in \Delta(\mathcal{C})$, which may or may not correspond to the training distribution.
We refer to this setting as \textbf{generalist improvement of generalist policies} (or \textbf{generalist improvement} for brevity).}
Given a task distribution $\rho_\mathrm{test}$, the objective of generalist improvement of generalist policies is to update $\pi$ so as to maximize:
\begin{equation}
   \textstyle  J(\pi; \rho_{\mathrm{test}}) = \mathbb{E}_{c \sim \rho_\mathrm{test}, s_0 \sim \mu_0(c)} \left[V^{\pi}(s_0, c)\right] . 
\end{equation}

\newcommand{\pigen}{\pi_{\mathrm{gen}}}
\newcommand{\rhotrain}{\rho_{\mathrm{train}}}
\newcommand{\rhotest}{\rho_{\mathrm{test}}}
\newcommand{\pift}{\pi_{\mathrm{ft}}}

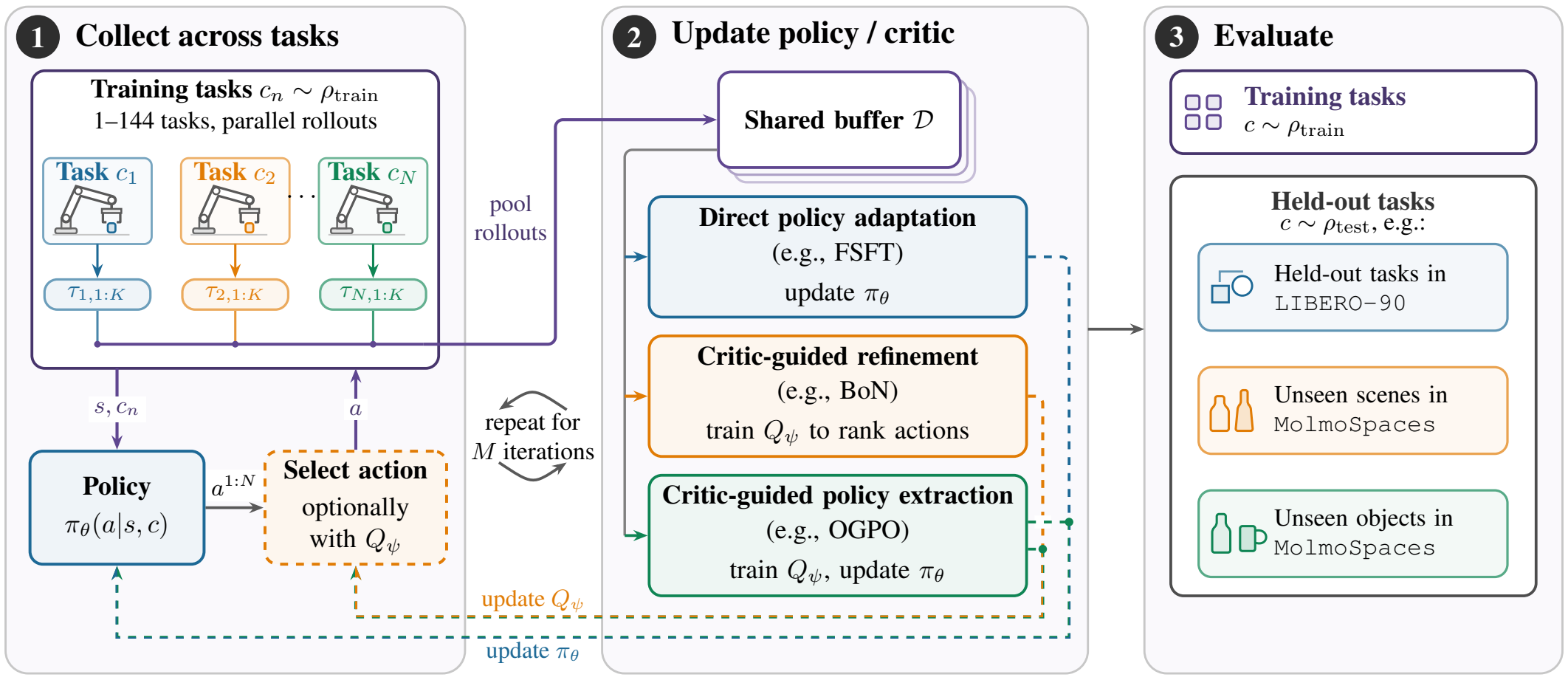
\begin{figure}[t]
\centering
\begingroup
\definecolor{mtblue}{HTML}{1D6996}
\definecolor{mtpurple}{HTML}{5F4690}
\definecolor{mtteal}{HTML}{38A6A5}
\definecolor{mtgreen}{HTML}{2CA02C}
\definecolor{mtorange}{HTML}{E17C05}
\definecolor{mtmagenta}{HTML}{E17C05}
\definecolor{mtgray}{HTML}{0F8554}
\definecolor{mtred}{HTML}{E17C05}
\definecolor{mtogpogreen}{HTML}{0F8554}
\colorlet{mtpanel}{mtpurple!3}   
\colorlet{mtpolicy}{mtblue}
\colorlet{mtcritic}{mtorange}
\resizebox{\linewidth}{!}{%
\begin{tikzpicture}[
  font=\rmfamily\fontsize{10}{12}\selectfont,
  panel/.style={draw=black!22, fill=mtpanel, rounded corners=3mm,
  line width=.8pt},
  badge/.style={circle, fill=black!82, text=white, minimum size=5.8mm,
    inner sep=0pt, font=\rmfamily\bfseries\fontsize{11}{12}\selectfont},
  heading/.style={font=\rmfamily\bfseries\fontsize{12}{14}\selectfont,
    text=black, anchor=west},
  subheading/.style={font=\rmfamily\fontsize{9.7}{11}\selectfont,
    text=black, anchor=west},
  card/.style={draw=black!55, fill=white, rounded corners=1.7mm,
    line width=.9pt, align=center, inner sep=3pt},
  pill/.style={rounded corners=2mm, inner xsep=5pt, inner ysep=3pt,
    font=\rmfamily\bfseries\fontsize{9.5}{11}\selectfont},
  flow/.style={-{Stealth[length=1.9mm,width=1.45mm]},
    draw=black!65, line width=1.05pt, rounded corners=1mm},
  interaction/.style={flow, draw=mtpurple, line width=1.15pt},
  update/.style={flow, draw=mtorange, line width=1.15pt},
  edge label/.style={fill=white, inner sep=2pt, text=black,
    font=\rmfamily\fontsize{9}{11}\selectfont},
  detail/.style={font=\rmfamily\fontsize{9.3}{11}\selectfont, text=black},
  object/.style={line width=.8pt, line join=round, draw=mtpurple!85!black,
    fill=mtpurple!8}
]
\path[use as bounding box] (-.015,-.965) rectangle (20.685,7.915);

\path[panel] (0,-.95) rectangle (6.10,7.90);
\path[panel] (7.92,-.95) rectangle (14.37,7.90);
\path[panel] (15.12,-.95) rectangle (20.67,7.90);

\node[badge] at (.43,7.50) {1};
\node[heading] at (.80,7.52) {Collect across tasks};
\node[badge] at (8.35,7.50) {2};
\node[heading] at (8.72,7.52) {Update policy / critic};
\node[badge] at (15.55,7.50) {3};
\node[heading] at (15.92,7.52) {Evaluate};
\draw[flow] (14.37,3.62) -- (15.12,3.62);

\path[card, draw=mtpurple!75!black, line width=1.05pt]
  (.35,3.10) rectangle (5.75,7.05);
\node[font=\rmfamily\bfseries\fontsize{10}{12}\selectfont]
  at (3.05,6.78) {Training tasks $c_n\sim\rho_{\mathrm{train}}$};
\node[detail] at (3.05,6.36) {1--144 tasks, parallel rollouts};

\foreach \x/\idx/\col in {1.22/1/mtblue,3.05/2/mtmagenta,4.88/N/mtgray} {
  \path[draw=\col!60, fill=\col!3, rounded corners=1mm, line width=.8pt]
    (\x-.72,4.75) rectangle (\x+.72,5.89);
  \node[font=\rmfamily\bfseries\fontsize{10}{11}\selectfont, text=\col]
    at (\x,5.69) {Task $c_{\idx}$};
  \begin{scope}[shift={(\x-.56,4.89)},scale=.68]
    \path[fill=black!12, draw=black!60, line width=.7pt, rounded corners=.25mm]
      (0,0) rectangle (.42,.10);
    \path[fill=black!8, draw=black!60, line width=.7pt]
      (.12,.10) rectangle (.30,.30);
    \draw[black!65, line width=4pt, line cap=round]
      (.21,.30) -- (.45,.83) -- (1.10,.65);
    \draw[white, line width=2.2pt, line cap=round]
      (.21,.30) -- (.45,.83) -- (1.10,.65);
    \foreach \u/\v in {.21/.30,.45/.83,1.10/.65}
      \filldraw[draw=black!65, fill=white, line width=.65pt]
        (\u,\v) circle (.067);
    \draw[black!65, line width=1pt] (1.10,.58) -- (1.10,.45);
    \path[draw=black!65, fill=black!10, line width=.7pt]
      (.91,.40) rectangle (1.29,.48);
    \draw[black!65, line width=.9pt, line join=round]
      (.93,.40) -- (.93,.22) -- (1.00,.22)
      (1.27,.40) -- (1.27,.22) -- (1.20,.22);
    \path[draw=\col, fill=\col!18, line width=.75pt, rounded corners=.3mm]
      (1.025,.02) rectangle (1.175,.20);
    \draw[black!22, line width=.6pt] (-.05,-.025) -- (1.45,-.025);
  \end{scope}
  \node[card, draw=\col!65, fill=\col!7, inner sep=1.5pt,
    text=\col, text width=13mm, minimum height=4mm,
    font=\rmfamily\fontsize{9}{10}\selectfont] (traj\idx) at (\x,4.08)
    {$\tau_{\idx,1:K}$};
  \draw[flow, draw=\col, line width=.85pt]
    (\x,4.75) -- (traj\idx.north);
  \draw[draw=\col!80, line width=.85pt]
    (traj\idx.south) -- (\x,3.42);
}
\node[font=\rmfamily\fontsize{10}{11}\selectfont, text=black]
  at (3.96,5.37) {$\cdots$};
\draw[draw=mtpurple, line width=1.1pt] (1.22,3.42) -- (5.55,3.42);
\foreach \x in {1.22,3.05,4.88}
  \fill[mtpurple] (\x,3.42) circle (1.25pt);

\node[card, draw=mtblue, left color=mtblue!7, right color=mtblue!7,
  line width=1.1pt, text width=21mm, minimum height=15mm] (policy) at (1.48,1.25)
  {\textbf{Policy}\\[2pt]$\pi_\theta(a|s,c)$\\[1pt]
   };
\node[card, draw=mtorange, fill=mtorange!7, line width=1.1pt, dashed,
  text width=22mm, minimum height=15mm] (selection) at (4.65,1.25)
  {\textbf{Select action}\\[2pt]
   {\fontsize{10}{11}\selectfont\color{black}optionally with $Q_\psi$}};
\draw[flow] (policy.east) -- (selection.west);
\node[edge label] at (3.04,1.55) {$a^{1:N}$};
\draw[interaction] (1.48,3.10) -- (policy.north);
\node[edge label, fill=mtpanel!2, text=mtpurple!75!black] at (1.48,2.55) {$s,c_n$};
\draw[interaction] (selection.north) -- (4.65,3.10);
\node[edge label, fill=mtpanel!2, text=mtpurple!75!black] at (4.65,2.55) {$a$};

\draw[flow] (7.45,2.55) .. controls (7.00,2.85) and (7.00,2.85) .. (6.55,2.55);
\draw[flow] (6.55,1.85) .. controls (7.00,1.55) and (7.00,1.55) .. (7.45,1.85);
\node[detail, align=center] at (7.00,2.20) {repeat for\\$M$ iterations};

\path[card, draw=mtpurple!35, fill=mtpurple!3]
  (9.67,5.57) rectangle (12.87,6.84);
\path[card, draw=mtpurple!50, fill=mtpurple!4]
  (9.57,5.67) rectangle (12.77,6.94);
\node[card, draw=mtpurple, fill=white, text width=30mm,
  minimum height=12.7mm, line width=1.05pt] (buffer) at (11.07,6.4)
  {\textbf{Shared buffer $\mathcal{D}$}};
\draw[interaction] (5.55,3.42) -- (7.30,3.42) -- (7.30,6.4) -- (buffer.west);
\node[edge label, fill=none, text=mtpurple!75!black, align=center]
  at (6.71,5.10) {pool\\rollouts};

\node[card, draw=mtblue, fill=mtblue!7, line width=1.05pt,
  text width=48mm, minimum height=16mm] (fsft) at (11.05,4.58)
  {\textbf{Direct policy adaptation}\\[1pt](e.g., FSFT)\\[3pt]
   {\fontsize{10}{11}\selectfont\color{black}update $\pi_\theta$}};
\node[card, draw=mtorange, fill=mtorange!7, line width=1.05pt,
  text width=48mm, minimum height=16mm] (td) at (11.05,2.73)
  {\textbf{Critic-guided refinement}\\[1pt](e.g., BoN)\\[3pt]
   {\fontsize{10}{11}\selectfont\color{black}train $Q_\psi$ to rank actions}};
\node[card, draw=mtogpogreen, fill=mtogpogreen!7, line width=1.05pt,
  text width=48mm, minimum height=16mm] (ogpo) at (11.05,.88)
  {\textbf{Critic-guided policy extraction}\\[1pt](e.g., OGPO)\\[3pt]
   {\fontsize{10}{11}\selectfont\color{black}train $Q_\psi$, update $\pi_\theta$}};
\draw[draw=black!50, line width=.95pt, rounded corners=1mm]
  ([yshift=-4mm]buffer.west) -- (8.22,6) -- (8.22,.88);
\draw[flow, draw=mtblue] (8.22,4.58) -- (fsft.west);
\draw[update] (8.22,2.73) -- (td.west);
\draw[flow, draw=mtogpogreen] (8.22,.88) -- (ogpo.west);


\draw[draw=mtpolicy, dashed, line width=1.05pt, rounded corners=1mm]
  (fsft.east) -- (14.12,4.58) -- (14.12,1.06);
\draw[draw=mtcritic, dashed, line width=1.15pt, rounded corners=1mm]
  (td.east) -- (13.77,2.73) -- (13.77,.70);
\draw[draw=mtpanel!2, line width=5pt] (13.66,1.06) -- (13.88,1.06);
\draw[draw=mtogpogreen, dashed, line width=1.05pt]
  ([yshift=1.8mm]ogpo.east) -- (14.12,1.06);
\draw[draw=mtogpogreen, dashed, line width=1.05pt]
  ([yshift=-1.8mm]ogpo.east) -- (13.77,.70);

\draw[flow, draw=mtogpogreen, dashed, line width=1.05pt, sharp corners]
  (14.12,1.06) -- (14.12,-.47) -- (1.48,-.47) -- (policy.south);
\begin{scope}
  \clip (-1,-1.5) -- (22,-1.5) -- (22,9) -- (14.12,9) -- (14.12,-.47)
    -- (1.48,-.47) -- (1.48,9) -- (-1,9) -- cycle;
  \draw[flow, draw=mtpolicy, dashed, line width=1.05pt, sharp corners]
    (14.12,1.06) -- (14.12,-.47) -- (1.48,-.47) -- (policy.south);
\end{scope}
\node[edge label, fill=none, inner sep=1.5pt, text=mtpolicy, anchor=north]
  at (7.01,-.47) {update $\pi_\theta$};

\draw[flow, draw=mtogpogreen, dashed, line width=1.15pt, sharp corners]
  (13.77,.70) -- (13.77,-.20) -- (4.65,-.20) -- (selection.south);
\begin{scope}
  \clip (4.65,-.20) rectangle (13.77,3);
  \draw[flow, draw=mtcritic, dashed, line width=1.15pt, sharp corners]
    (13.77,.70) -- (13.77,-.20) -- (4.65,-.20) -- (selection.south);
\end{scope}
\node[edge label, fill=none, inner sep=1.5pt, text=mtcritic, anchor=south]
  at (7.01,-.20) {update $Q_\psi$};

\fill[mtogpogreen] (14.12,1.06) circle (1.6pt);
\fill[mtogpogreen] (13.77,.70) circle (1.6pt);

\path[card, draw=mtpurple!75!black, line width=1.05pt]
  (15.47,5.95) rectangle (20.32,7.05);
\begin{scope}[shift={(1.73,1.05)}]
  \foreach \x/\y in {13.94/5.23,14.22/5.23,13.94/5.49,14.22/5.49}
    \path[draw=mtpurple, fill=mtpurple!12, rounded corners=.35mm,
      line width=.85pt] (\x,\y) rectangle ++(.18,.18);
\end{scope}
\node[font=\rmfamily\bfseries\fontsize{10}{12}\selectfont, text=mtpurple!75!black,
  anchor=west] at (16.32,6.68) {Training tasks};
\node[detail, anchor=west] at (16.32,6.28) {$c\sim\rho_{\mathrm{train}}$};

\path[card, draw=black!70, line width=1.05pt]
  (15.47,.08) rectangle (20.32,5.65);
\node[font=\rmfamily\bfseries\fontsize{10}{12}\selectfont, text=black!80]
  at (17.89,5.32) {Held-out tasks};
\node[detail] at (17.89,4.98) {$c\sim\rho_{\mathrm{test}}$, e.g.:};

\path[card, draw=mtblue!70, fill=mtblue!5, line width=.9pt]
  (15.84,3.60) rectangle (19.94,4.75);
\begin{scope}[shift={(2.11,.255)}]
  \path[draw=mtblue, fill=mtblue!12, line width=.8pt]
    (13.91,3.72) rectangle (14.13,3.94);
  \path[draw=mtblue, fill=white, line width=.8pt]
    (14.31,3.94) circle (.13);
  \draw[draw=mtblue!70, line width=.8pt]
    (14.01,3.99) -- (14.01,4.13) -- (14.32,4.13);
\end{scope}
\node[detail, anchor=west, text width=30mm]
  at (16.72,4.175) {Held-out tasks in \texttt{LIBERO-90}};

\path[card, draw=mtred!70, fill=mtred!5, line width=.9pt]
  (15.84,1.965) rectangle (19.94,3.115);
\begin{scope}[shift={(2.09,.28)}]
  \path[draw=mtred, fill=mtred!10, line width=.85pt,
    rounded corners=.3mm]
    (13.91,1.99) -- (13.91,2.27) -- (13.97,2.34) -- (13.97,2.45)
    -- (14.09,2.45) -- (14.09,2.34) -- (14.15,2.27) -- (14.15,1.99) -- cycle;
  \path[draw=mtred, fill=mtred!20, line width=.85pt,
    rounded corners=.3mm]
    (14.24,1.99) -- (14.27,2.31) -- (14.31,2.35) -- (14.31,2.53)
    -- (14.40,2.53) -- (14.40,2.35) -- (14.44,2.31) -- (14.47,1.99) -- cycle;
\end{scope}
\node[detail, anchor=west, text width=30mm]
  at (16.72,2.54) {Unseen scenes in \texttt{MolmoSpaces}};

\path[card, draw=mtgray!70, fill=mtgray!5, line width=.9pt]
  (15.84,.33) rectangle (19.94,1.48);
\begin{scope}[shift={(2.14,-.045)}]
  \path[draw=mtgray, fill=mtgray!12, line width=.85pt,
    rounded corners=.3mm]
    (13.88,.70) -- (13.88,1.00) -- (13.94,1.07) -- (13.94,1.21)
    -- (14.06,1.21) -- (14.06,1.07) -- (14.12,1.00) -- (14.12,.70) -- cycle;
  \path[draw=mtgray, fill=mtgray!20, line width=.85pt,
    rounded corners=.3mm] (14.25,.72) rectangle (14.45,1.05);
  \draw[draw=mtgray, line width=.85pt] (14.45,.99)
    .. controls (14.65,1.03) and (14.64,.77) .. (14.45,.83);
\end{scope}
\node[detail, anchor=west, text width=30mm]
  at (16.72,.905) {Unseen objects in \texttt{MolmoSpaces}};

\end{tikzpicture}%
}
\endgroup
\caption{
\textbf{Generalist improvement and evaluation.}
\textbf{(1)} A generalist policy collects $K$ rollouts $\tau_{n,k}$ across training tasks $c_n\sim\rho_{\mathrm{train}}$.
\textbf{(2)} Task-specific experience is pooled in a shared replay buffer and used to enact \chg{one of three families of methods}: direct policy adaptation updates the policy \chg{according to success labels}, critic-guided refinement learns critics for action ranking\chg{, and critic-guided policy extraction learns critics and updates the policy towards actions with high critic values.} This process is iterated $M$ times.
\textbf{(3)} Evaluation is carried out over training tasks ($c\sim\rho_{\mathrm{train}}$), as well as on unseen test tasks ($c\sim\rho_{\mathrm{test}}$) to test generalization.}
\label{fig:overview}
\end{figure}

\textbf{Generalist improvement protocol.\;}
\looseness = -1 In this work we assume we are given some initial generalist policy $\pigen$, achieving non-zero but imperfect zero-shot success across tasks, and \chg{that we} can interact online with a set of training tasks drawn from a distribution $\rhotrain$. \chg{As stated, our goal is to learn} a fine-tuned policy $\pift$ that maximizes $J(\pift; \rhotest)$. 
Crucially, $\rhotest$ need not coincide with $\rhotrain$\chg{, in which case the policy must generalize} to tasks it was not trained on.
At the same time, these two distributions need to be related (see Figure~\ref{fig:molmo_visualization_small}), as generalization to arbitrary tasks is in practice unrealistic (see Appendix~\ref{app:libero}, ~\citealp{yu2020gradientsurgerymultitasklearning} and ~\citet{malik2021generalizablereinforcementlearningtractable}).
For simplicity, we will assume that $\rhotrain$ is supported on a finite set of tasks $[c_1,\ldots,c_{N_{\mathrm{task}}}]$. We consider interaction with the training tasks through repeated iterations $m = 1, \ldots, M$, each involving online interaction and training.
For each iteration $m$, we first collect $K$ episodes~$\tau_{1,1:K},\dots,\tau_{N_{\mathrm{task}},1:K}$ from each training task $[c_1,\dots,c_{N_{\mathrm{task}}}]$, with the policy frozen, and then update the policy given the newly collected experience.
This protocol approaches traditional, single-task online RL by choosing $\rho_{\mathrm{train}}=\rho_{\mathrm{test}}$ and setting $K =1$, and tends towards multi-task offline RL \citep[or ``batch online RL'',][]{dong2025matters} when a single iteration is applied ($M=1$, $K \gg 1$) and $\rho_\mathrm{train}$ has larger support.
To simulate realistic fleet-level robot deployments, we choose a moderately large $K$ and $M$, discussed in~\Cref{sec:allocation}.
We visualize the protocol in Figure~\ref{fig:overview}.

\subsection{Mechanisms for Generalist Improvement}

\looseness -1 In this work, we aim to understand what algorithmic approaches enable generalist improvement. We distinguish between two complementary and independent mechanisms for policy improvement: 
\textit{direct policy adaptation}, which updates the parameters of the robot policy alone, and \textit{critic-guided refinement}, which instead learns a value function to ``grade'' robot actions.
\chg{We consider approaches that isolate each of these components (FSFT, BoN), in order to determine which mechanism drives generalist improvement, as well as representative approaches that combine both (in particular, we primarily consider OGPO).
We emphasize that the focus of this work is not primarily developing novel algorithms, but rather understanding whether existing techniques for policy learning can enable generalist improvement. We therefore focus on existing paradigmatic approaches. We defer specific implementation details of the algorithms described below to Appendix \ref{app:implementation_methods}.}

\textbf{Direct policy adaptation.}
As a representative and scalable approach to direct policy adaptation, we consider weighted supervised learning.
At each iteration, we sample a batch of state-action-task tuples from the buffer, and define a weighting function $w: \mathcal{S \times A \times C} \to \mathbb{R}$ in order to minimize a weighted negative log-likelihood under the policy:
\begin{equation}
    \mathcal{L}_\text{DPA}(\theta) = -\,\mathbb{E}_{(s,a,c)\sim\mathcal{D}} \big[ w(s,a,c)\log \pi_\theta(a|s,c)\big], \label{eq:weighted_supervised}
\end{equation}
for policy parameters $\theta$.
When an explicit likelihood is not available, a tractable lower bound 
can be adopted (e.g., the flow matching loss in the case of modern VLAs~\citep{intelligence2025pi_}).
The simplest instantiation of this mechanism is referred to as \%-BC~\citep{chen2021decision} or filtered supervised fine-tuning~\citep[FSFT,][]{zhou2024autonomous}, which \chg{sets} $w(s_t,a_t,c)=\mathbf{1}_{G(s_t,a_t,c)>G_{\min}}$ where $G(s_t,a_t,c)=\sum_{t' \geq t} R(s_{t'}, a_{t'}, c)$ is the empirical return, and $G_{\min}$ a threshold.
This mechanism implements a form of policy improvement~\citep{qin2025supervised} while isolating the policy update mechanism and conveniently remaining as close as possible to a purely supervised objective. \chg{We thus adopt FSFT as our paradigmatic approach for direct policy adaptation; we nevertheless also consider approaches that set $w(s,a,c)$ based on a learned value function in Section~\ref{sec:policy_extraction}.}

\textbf{Critic-guided \chg{test-time} refinement.}
Orthogonally to direct policy adaptation, improvement can also occur through utilizing a learned value function.
We consider training a parametric state-action critic $Q_\psi$ to estimate the state-action value function expressed in Equation~\ref{eq:q}.
While many approaches for value estimation exist (and are ablated in Section~\ref{sec:exp_value}), we find that training $Q_\psi$ together with a state value function
$V_\eta$ through the following standard temporal difference (TD) learning update performs reliably:
\begin{align}
    \label{eq:q_loss}
    \textstyle \mathcal{L}_{Q}(\psi) & = \mathbb{E}_{(s,a,r,s',c) \sim \mathcal{D}} \Big[ \big(Q_\psi(s,a,c) - (r + \gamma V_\eta(s',c))\big)^2 \Big] \quad \text{and} \quad \\ 
    \label{eq:v_loss}
    \textstyle \mathcal{L}_V(\eta) &= \mathbb{E}_{(s,a,c) \sim \mathcal{D}} \Big[ \big(Q_\psi(s,a,c) - V_\eta(s,c)\big)^2 \Big].
\end{align}
\chg{We refer to} Section~\ref{sec:exp_value} for ablations of our critic-learning objective (e.g., Monte Carlo returns and distributional critics).
As our goal is to understand the mechanisms that enable generalist improvement, we wish to isolate the contribution of policy adaptation from critic refinement. 
To achieve this, we build on a long line of works on test-time action selection~\citep{kalashnikov2018qtopt} that guide the policy via Best-of-N (BoN) sampling~\citep{nakamoto2025steering}.
In particular, we keep the policy parameters frozen and re-rank sampled actions at test-time according to their estimated values.
That is, for a task $c$ and the current state $s_t$ in a rollout, we draw $N$ action candidates $\{a_t^i\}_{i=1}^{N} \sim \pi_\theta(s_t, c)$, score them and execute action chunk $a_t^{i^\star}$ with
\begin{equation}
    \label{eq:bon}
    i^\star = \arg \max_i Q_{\mathrm{targ}}(s_t, a_t^i, c).
\end{equation}
\chg{where $Q_{\mathrm{targ}}$ is a target critic network \citep{lillicrap2019continuous} derived from $Q_{\psi}$ (see \Cref{app:implementation_methods} for details).
This allows online} experience to influence behavior through action selection while preserving the policy parameters inherited from pre-training; moreover, \chg{despite its simplicity, we found this recipe to perform reliably in practice (see Section~\ref{sec:scaling}).}

\chg{
\textbf{Critic-guided policy extraction.\;} 
Lastly, we consider a third approach to policy improvement: updating the base policy so as to maximize the value the learned critic assigns to its actions \citep{peng2019advantage,lillicrap2019continuous}. Formally, rather than simply using the critic to select actions from the base policy as in Equation \ref{eq:bon}, we \emph{train} the policy to produce actions with large critic values, for example, maximizing an objective of the form:
\begin{equation}\label{eq:cpe}
    \mathcal{L}_{\mathrm{CPE}}(\theta) = \mathbb{E}_{(s,c) \sim \mathcal{D}} \mathbb{E}_{a \sim \pi_\theta(\cdot \mid s, c)}[Q_{\mathrm{targ}}(s,a)].
\end{equation}
As a representative recipe, we focus on off-policy generative policy optimization \citep[OGPO,][]{patil2026ogpo}. Specifically, OGPO is targeted at generative control policies \citep{chi2023diffusion,pan2026much}, such as flow-policies \citep{black2024pi_0}, which produce actions via a sequence of denoising steps.
OGPO alternates between training a critic on observed transitions, using an update similar to Equation \ref{eq:q_loss}, and updating the generative control policy $\pi_\theta$ to maximize (Equation \ref{eq:cpe}), applying a PPO-like update \citep{schulman2017proximal} to the action generation process to shift its generated actions towards higher critic values (see Appendix~\ref{app:taxonomy} for a detailed description).
}

\chg{
In addition to OGPO, we ablate against two other approaches to critic-guided policy extraction:
(i)~FSFT+BoN, which adapts the policy with $\mathcal{L}_\mathrm{DPA}$ and filters action at test-time according to a critic trained with $\mathcal{L}_\mathrm{Q/V}$, and (ii)
flow-based advantage-weighted regression \citep[AWR,][]{peng2019advantage}, where we use critic-derived estimates of the advantage function to reweight the supervised learning loss following Equation \ref{eq:weighted_supervised}. We evaluate each in Section~\ref{sec:policy_extraction} and provide full details in Appendices~\ref{app:taxonomy} and~\ref{app:experimental_details}.
}

\subsection{Experimental Setup}
\label{sec:exp_setup}

Our experiments are designed to answer the following core questions:
\begin{itemize}[leftmargin=*]
    \item \textbf{Q1:} As the diversity of training tasks increases, how effective are the \chg{three} representative recipes (direct policy adaptation, critic-guided \chg{test-time} refinement \chg{and critic-guided policy extraction}) at driving self-improvement on the tasks they are exposed to?
    \item \textbf{Q2:} To what extent and at which scale does either mechanism generalize to unseen tasks?
    \item \chg{\textbf{Q3:} What is the nature of the observed generalization, and in which cases is it most effective?}
    \item \textbf{\chg{Q4:}} Which specific algorithmic design choices drive learning within and beyond the training task distribution, i.e., enable generalist improvement?
\end{itemize}

To answer these questions, we design a post-training and evaluation protocol centered on directly controlling the training and test task distributions $\rho_\mathrm{train}$ and $\rho_\mathrm{test}$ introduced in Section~\ref{sec:setting}.

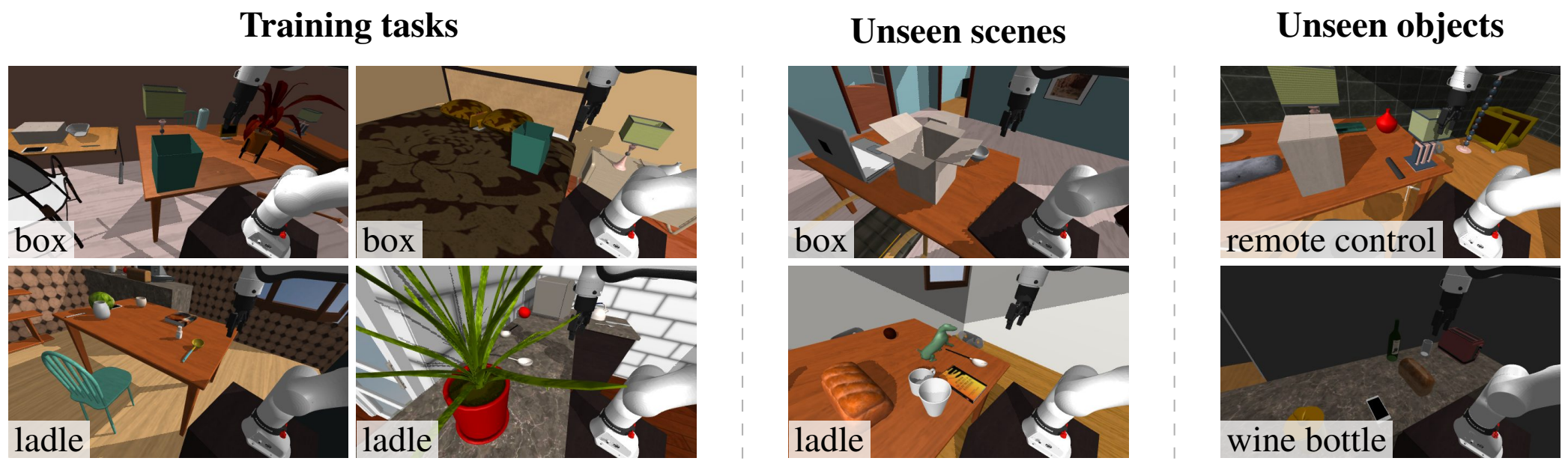
\begin{figure}[t]
\vspace{-6mm}
  \centering
  \begin{tikzpicture}[
    img/.style={inner sep=0pt, anchor=north west},
    rowlbl/.style={anchor=east, align=right, font=\footnotesize},
    head/.style={anchor=south, font=\footnotesize\bfseries},
    tag/.style={anchor=south west, inner sep=1.5pt, fill=white, fill opacity=0.75,
                text opacity=1, font=\small},
    sep/.style={gray!60, dashed}
  ]
    \def\w{3cm}   
    \def\g{0.06cm}  
    \def\dir{figs/render/molmo_pics}

    \node[img] (b1) at (0,0)                       {\includegraphics[width=\w]{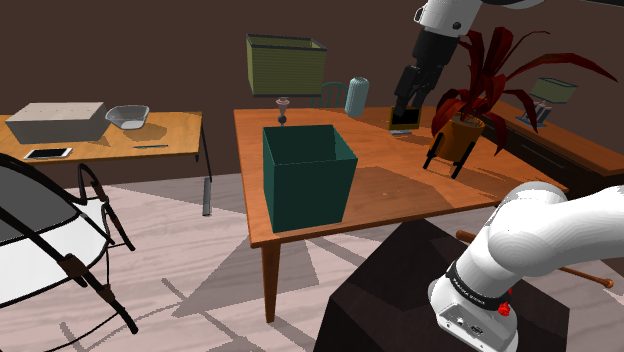}};
    \node[img] (b2) at ([xshift=\g]b1.north east)  {\includegraphics[width=\w]{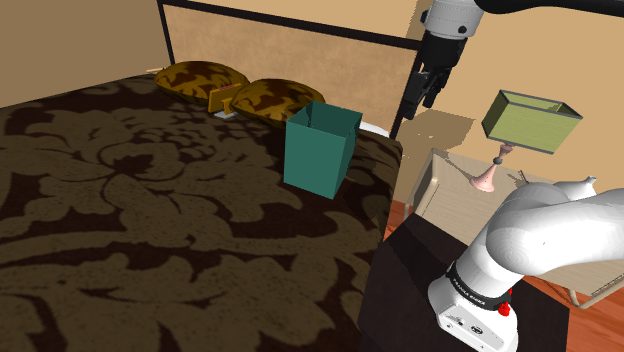}};
    \node[img] (s1) at ([yshift=-\g]b1.south west) {\includegraphics[width=\w]{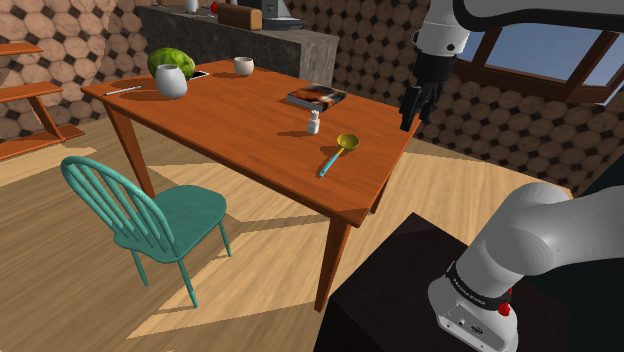}};
    \node[img] (s2) at ([xshift=\g]s1.north east)  {\includegraphics[width=\w]{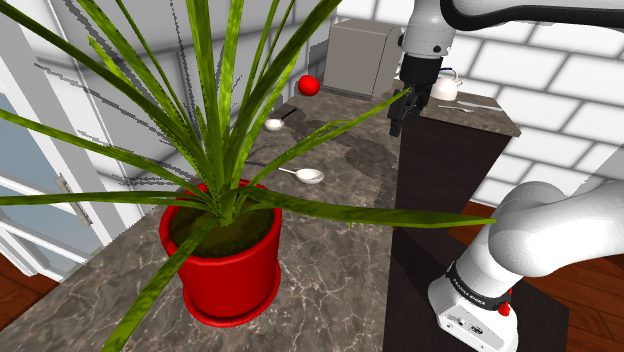}};
    \node[tag] at (b1.south west) {box};
    \node[tag] at (b2.south west) {box};
    \node[tag] at (s1.south west) {ladle};
    \node[tag] at (s2.south west) {ladle};
    \node[head] at ([yshift=2pt]b1.north east) {Training tasks};

    \draw[sep] ([xshift=0.4cm]b2.north east) -- ([xshift=0.4cm]s2.south east);

    \node[img] (u1) at ([xshift=0.8cm]b2.north east) {\includegraphics[width=\w]{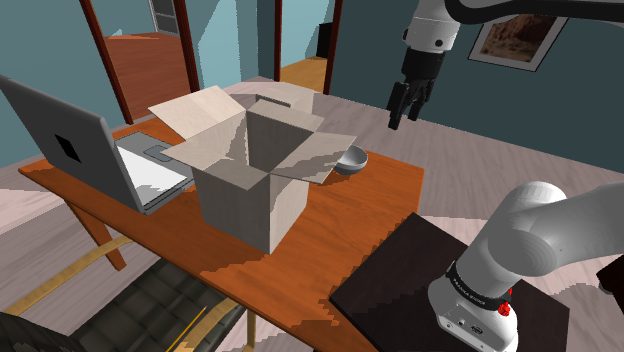}};
    \node[img] (u2) at ([xshift=0.8cm]s2.north east) {\includegraphics[width=\w]{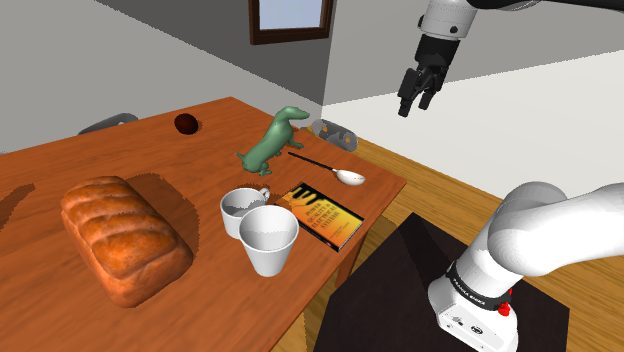}};
    \node[tag] at (u1.south west) {box};
    \node[tag] at (u2.south west) {ladle};
    \node[head] at ([yshift=2pt]u1.north) {Unseen scenes};

    \draw[sep] ([xshift=0.4cm]u1.north east) -- ([xshift=0.4cm]u2.south east);

    \node[img] (o1) at ([xshift=0.8cm]u1.north east) {\includegraphics[width=\w]{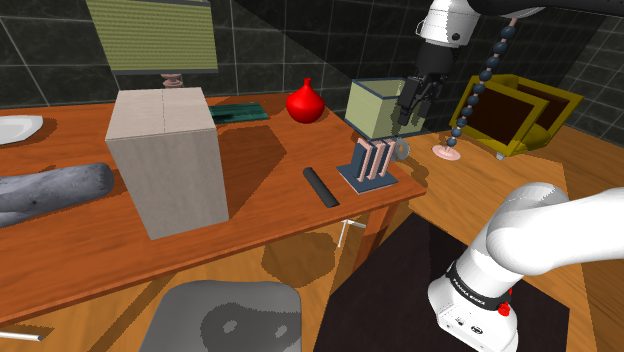}};
    \node[img] (o2) at ([xshift=0.8cm]u2.north east) {\includegraphics[width=\w]{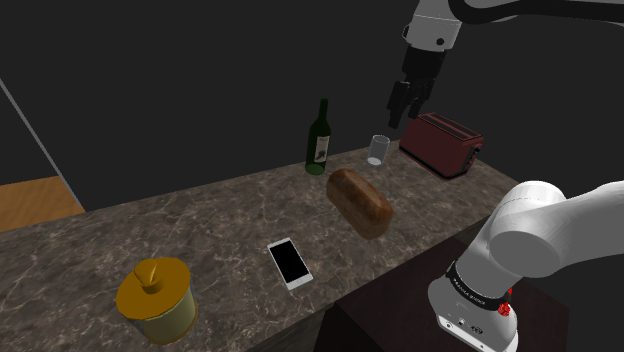}};
    \node[tag] at (o1.south west) {remote control};
    \node[tag] at (o2.south west) {wine bottle};
    \node[head] at ([yshift=2pt]o1.north) {Unseen objects};
  \end{tikzpicture}
  \caption{\textbf{Objects and scenes in \Molmo}. Left: 4 tasks covering two objects, each with two scenes. Center: unseen scenes of the training objects. Right: examples of unseen objects.}
  \label{fig:molmo_visualization_small}
\end{figure}

\textbf{Training environments.\;}
\looseness=-1
To answer Q1 and Q2, we consider training distributions that differ in both the number of distinct tasks and their diversity. 
We base our evaluation on the \texttt{MolmoSpaces}~\citep{kim2026molmospaces} benchmark, which gives  precise control over task diversity. 
We focus on the \texttt{pick} suite, which instructs the agent to grasp and lift various objects in different domestic environments. This benchmark exposes 18 task objects and 55-56 scenes per object (e.g., the ``fork" object spans 56 distinct scenes)\chg{; we visualize examples in Figure~\ref{fig:molmo_visualization_small}. This enables task variation along multiple axes: scene variation, which is primarily visual in nature, and object variation, which generally alters the grasping strategy necessary to successfully solve the task.}
In our main evaluation (Figure~\ref{fig:main}), we consider training task distributions over growing numbers of tasks, from $1$ to \chg{$144$}, spanning different objects and scenes.
Each set of tasks is procedurally selected to be a superset of smaller sets, while maintaining the same non-zero average performance of the base policy over tasks (${\approx}30\%$).
This ensures that larger sets of tasks are at least as diverse as smaller ones, as they cannot include fewer objects and scenes.
Additional evaluations, e.g., in Figure~\ref{fig:obj_vs_scenes}, ablate the diversity in the set of training tasks, which may only vary in terms of scenes, or also in terms of objects.
We additionally evaluate on the popular \texttt{LIBERO}~\citep{liu2023libero} manipulation suite, for which tasks are sampled from the \texttt{LIBERO-90} subset.

\textbf{Evaluation tasks.\;} 
\looseness=-1
To address Q1, we evaluate policy performance with respect to the training task distribution ($\rho_\mathrm{test} = \rho_\mathrm{train}$).
To address Q2, we largely leverage \texttt{MolmoSpaces}'s structured task diversity, varying task novelty across both the axes of task objects and scenes, described above and depicted in Figure~\ref{fig:molmo_visualization_small}.
Similarly, we evaluate both training and unseen tasks in \texttt{LIBERO} to assess generalization capability when task diversity cannot be easily controlled.
For a full description of task sets in \texttt{MolmoSpaces} we refer to Appendix~\ref{app:task_splits}.
In all experiments, we report the final mean success rate over three training seeds and 95\% bootstrap CIs; see Appendix~\ref{app:training_curves} for training curves.

\textbf{Base model, critic architecture, and policy improvement protocol.\;} Across experiments, we use the~$\pi_{0.5}$ VLA~\citep{intelligence2025pi_} as the base \chg{generalist policy $\pigen$}, in its \texttt{droid} and \texttt{libero} checkpoints for \texttt{MolmoSpaces} and \texttt{LIBERO}, respectively.
Method-specific hyperparameters are optimized individually, and main results are stated for the best-performing hyperparameters for each approach.
We briefly recap them here, and present detailed allocation studies and ablations justifying these decisions in Section~\ref{sec:analysis} and Appendix~\ref{app:additional_exp}.
We implement our critic networks as BroNet~\citep{nauman2024bigger} residual MLPs whose input is the concatenation of the~$\pi_{0.5}$ prefix representation, mean-pooled over the prefix token sequence, the proprioceptive state, and the flattened action chunk.
We use an ensemble of ~$Q$ and~$V$-heads and aggregate them by taking the minimum~\citep{fujimoto2018addressingfunctionapproximationerror}, both when forming the TD target in~\Cref{eq:q_loss} and when scoring candidate actions at inference time via~\Cref{eq:bon}.
In each experiment, we apply $M = 10$ iterations of policy improvement, collecting $K = 20$ episodes per task per epoch\chg{; we find that task performance does not improve much for larger $K$ and $M$, as we report in Appendix~\ref{sec:allocation}.}
\chg{For direct policy adaptation,} we fine-tune policies for $100$ gradient steps in single-task settings, and $500$ gradient steps in multi-task settings due to greater data scale and variability.
\chg{In the case of critic-guided test-time refinement, }the critic MLPs, being both smaller and trained from scratch, are trained with $100\times$ gradient steps (reaching convergence, and roughly matching the training cost of policy adaptation).
\chg{We found methods relying on critic-guided policy extraction to prefer a different allocation of gradient steps (fewer critic updates with off-policy targets, and more policy updates, as detailed in Appendix~\ref{app:hyperparameters}).}

\chg{We ensure reproducibility by open-sourcing our full pipeline at this  \href{https://github.com/lasgroup/generalist-improvement}{repository}.}

\section{Generalist Improvement Scales With Task Diversity}
\label{sec:scaling}

\begin{figure}[t]
    \centering
    \includegraphics[width=\linewidth]{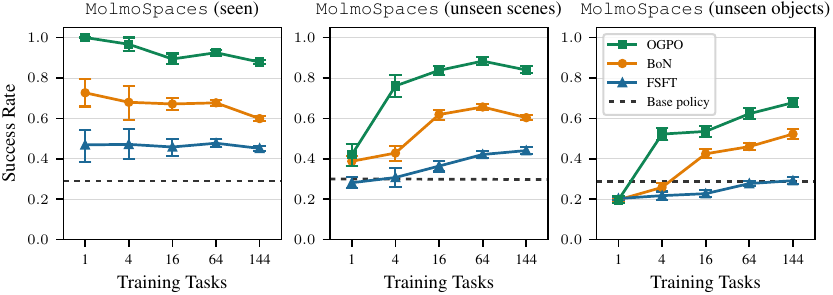}
    \caption{\textbf{Value-based generalist improvement generalizes beyond the training tasks.} Final success rate on training tasks (left) and held-out tasks (center and right, unseen scenes and objects) in \texttt{MolmoSpaces} as the number of distinct training tasks increases (x-axis). BoN \chg{and OGPO} display stronger generalization. We report the mean and 95\% stratified bootstrap CIs across three seeds.\looseness -1}
    \label{fig:main}
\end{figure}

We seek to answer our \chg{four} core questions about generalist policy improvement. Our analysis centers on Figure~\ref{fig:main}, in which each point
corresponds to a training split of $k$ objects with $k$ scenes each, evaluated
on its own training tasks (left)\chg{, on eight unseen scenes of each of its $k$
training objects (center), and on a fixed set of six unseen objects with eight
scenes each (right).} Appendix~\ref{app:task_splits} lists
every split in full.

\subsection{Q1: Can we enable generalist improvement on training tasks?}
\label{sec:q1}

\begin{wrapfigure}[22]{r}{0.39\linewidth}
  \vspace{-4mm}
  \includegraphics[width=\linewidth]{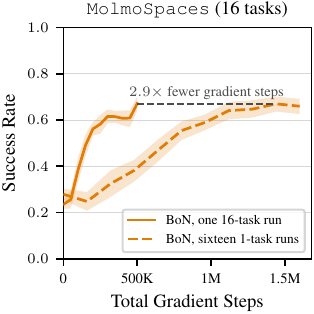}
  \caption{\chg{\textbf{Compute efficiency of generalist improvement.} Post-training a policy on multiple tasks at once achieves the performance of post-trained single-task experts in significantly fewer gradient steps.}}
  \label{fig:compute_efficiency}
\end{wrapfigure}

The left panel of Figure~\ref{fig:main} evaluates to what extent post-training can improve performance on the tasks sampled from the training task distribution.
For \chg{FSFT, BoN and OGPO}, we observe that success rate increases with respect to the base policy, and generalist improvement succeeds.
We find \chg{OGPO and BoN} to be generally more effective than FSFT in the single-task setting, where \chg{they lead by $+53$\,pp and $+26$\,pp, respectively}.
As the number of training tasks increases,
we find that in the multi-task setting \chg{OGPO and BoN} outperform FSFT by
\chg{$+43$--$50$\,pp and $+15$--$21$\,pp, respectively.}
Neither of the recipes displays fully convincing cross-task expertise: while potentially overfitting to a specific task, the average performance of single-task experts \chg{(represented as the leftmost point on these curves)} \chg{slightly} exceeds that of multi-task policies.
We observe similar trends on training tasks for \texttt{LIBERO}, in which the performance gap with respect to single-task experts is perhaps more marked -- see Appendix \ref{app:libero}.
\chg{We note, however, that generalist improvement nevertheless holds a computational edge over post-training single-task experts, as we display in Figure \ref{fig:compute_efficiency}: with a matched data collection budget, a single BoN post-training run over 16 tasks requires $2.9\times$ fewer gradient steps than 16 single-task runs to achieve similar multi-task performance improvements. Thus, even on in-distribution training tasks, generalist improvement enables more efficient improvement than single-task training.}

\subsection{Q2: Do we see generalization beyond training tasks?}
\label{sec:q2}

\looseness=-1A different pattern emerges when, instead of measuring performance on the tasks we train on, we turn to evaluating success rate on unseen tasks in \texttt{MolmoSpaces}.
In this setting, we find, to the best of our knowledge, \chg{one of} the first examples of generalization in VLA RL post-training at this scale.
As shown in~\Cref{fig:main} (center and right), single-task post-training 
\chg{leads to forgetting on unseen objects, and yields at most modest gains on unseen scenes --} while we can improve on the task we train on, performance \chg{often} degrades on others.
\begin{wrapfigure}[22]{r}{0.66\linewidth}
\centering
\includegraphics[width=\linewidth]{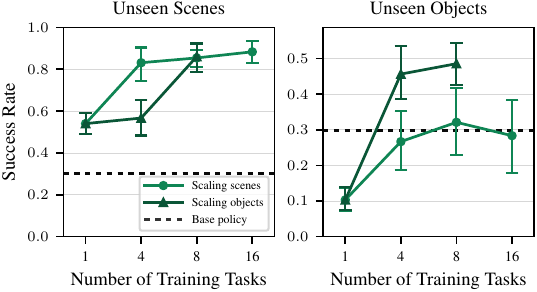}
  \caption{\textbf{Diversification transfers better along its own axis.} \chg{OGPO}
    success rate when scaling scenes of a single object, or objects with one scene
    each, evaluated on eight unseen scenes (\textit{left}) and eight unseen objects
    (\textit{right}). Each axis transfers best along itself, though only scaling
    objects surpasses the base policy on both. Mean and 95\% stratified bootstrap
    CIs over three seeds.}
  \label{fig:obj_vs_scenes}
\end{wrapfigure}
However, as we increase the number of training tasks, \chg{critic-based recipes (OGPO and BoN) exhibit} a transition from forgetting to generalization. Specifically, we observe performance gains over the base policy of up to 
\chg{$+58$\,pp} for tasks where the object remains consistent but the scene changes from the training task, and 
\chg{$+39$\,pp}
for tasks where the object also changes.
This indicates that \chg{OGPO and }BoN can even handle different objects altogether (e.g., picking up a knife after training only on other items, such as ladles, pots, and cups\chg{, as we further analyze in the next section).}
More generally, these results highlight that post-training VLAs on sufficiently diverse tasks can yield improvements \textit{beyond} the training distribution. \chg{Indeed, this generalization \emph{scales with the number of training tasks} -- the more tasks we train on, the greater the generalization benefit -- as long as the correct algorithmic approach is applied.}
In particular, this improvement is much more pronounced for \chg{OGPO and} BoN than FSFT, demonstrating that critic-guided \chg{policy extraction or} refinement is critical for enabling generalization. \chg{Together, this suggests that an effective recipe for generalist improvement is to  (i) apply critic-based policy improvement while (ii) scaling the number of tasks we RL post-train on.}
We additionally evaluate generalization in \texttt{LIBERO} in Appendix~\ref{app:libero}, in which we find that this benchmark presents a much greater challenge in terms of generalization, as inter-task variability is extreme: in this case, similar patterns occur, but only in terms of forgetting.

\subsection{\chg{Q3: What Is the Nature of the Observed Generalization?} }
\label{sec:qtodo_generalization_analysis}

\chg{
We next carefully analyze which training task splits are most critical for enabling generalization, and which types of generalization we might expect a given set of training tasks to induce.
}

\chg{\textbf{Scaling training objects is critical for generalist improvement.}}
We first
isolate the two axes -- scene and object -- of task diversity. In Figure~\ref{fig:main} they grow
together: a split with $k^2$ tasks contains $k$ objects with $k$ scenes each.
In this experiment, we instead vary the number of scenes of a single object, 
and separately the number of objects with one scene each. Both are evaluated on eight unseen objects, and on eight unseen scenes of the object whose scenes the first sweep scales (details and results for all recipes in Appendix~\ref{app:scaling_extended}). As shown in Figure~\ref{fig:obj_vs_scenes}, we observe \chg{earlier} generalization to new scenes by training on more diverse scenes, and greater generalization to new objects by training on more diverse objects. 
The two are not
symmetric: scaling objects enables the post-trained policy to surpass the base policy on both unseen objects and scenes,
whereas scaling scenes \chg{hardly} exceeds it on unseen objects, even with 16 scenes. This \chg{suggests} that, to achieve generalization to novel manipulation strategies (e.g., the grasp mechanics \chg{required to handle} a new object), it is necessary to train on tasks \chg{involving diverse} manipulation strategies, \chg{while robustness over the scene layout and appearance is easier to achieve.}

\begin{figure}[t]
    \centering
    \includegraphics[width=1.0\linewidth]{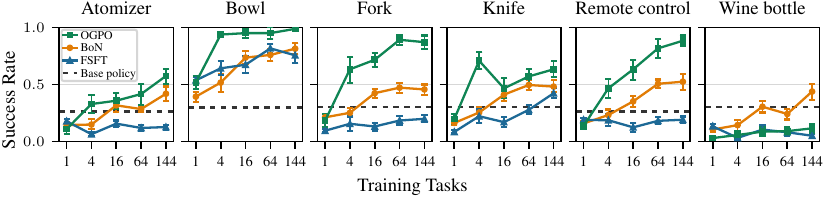}
    \caption{\texttt{MolmoSpaces}. Final success
      rate on each of the 6 unseen objects, none of which appears in any training split, as a function of the number of training tasks. Points show the mean over three seeds with 95\% stratified bootstrap CIs.}
    \label{fig:generalizaton_per_objects}
\end{figure}

\begin{figure}[t]
  \centering
  \begin{minipage}[t]{0.33\linewidth}
    \vspace{5mm}
    \centering
    \begingroup
    \definecolor{mtorange}{HTML}{E17C05}
    \resizebox{\linewidth}{!}{%
    \begin{tikzpicture}[
      x=1pt, y=1pt,
      font=\rmfamily\small,
      cell/.style={inner sep=0pt, anchor=base},
      lbl/.style={cell, anchor=base west},
      head/.style={cell, text=mtorange},
      rule/.style={draw=mtorange, line width=0.9pt},
      midrule/.style={draw=mtorange!60, line width=0.5pt}
    ]
      \def\W{145}

      \fill[mtorange!5] (0,-18) rectangle (\W,0);
      \draw[rule]    (0,0)   -- (\W,0);
      \draw[midrule] (0,-18) -- (\W,-18);

      \node[head] at (14,-12) {Tasks};
      \node[head, anchor=base west] at (34,-12) {Objects};

      \node[cell] at (14,-29) {1};
      \node[lbl]  at (34,-29) {\emph{box} or \emph{spatula}};
      \node[cell] at (14,-40.5) {4};
      \node[lbl]  at (34,-40.5) {\emph{box}, \emph{spatula}};
      \node[cell] at (14,-52) {16};
      \node[lbl]  at (34,-52) {$+$\,\emph{cup}, \emph{tissuepaper}};
      \node[cell] at (14,-63.5) {64};
      \node[lbl]  at (34,-63.5) {$+$\,\emph{boiler}, \emph{ladle},};
      \node[lbl]  at (34,-73.5) {\phantom{$+$\,}\emph{pot}, \emph{spoon}};
      \node[cell] at (14,-88) {144};
      \node[lbl]  at (34,-88) {$+$\,\emph{salt/pepper},};
      \node[lbl]  at (34,-98) {\phantom{$+$\,}\emph{soapdispenser},};
      \node[lbl]  at (34,-108) {\phantom{$+$\,}\emph{mug}, \emph{fruit/veg.}};

      \draw[rule] (0,-114) -- (\W,-114);
    \end{tikzpicture}%
    }
    \endgroup
  \end{minipage}\hfill
  \begin{minipage}[t]{0.64\linewidth}
    \vspace{0pt}
    \centering
    \includegraphics[width=\linewidth]{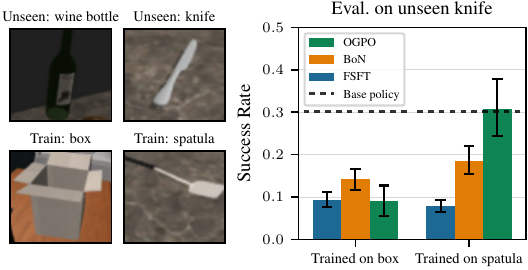}
  \end{minipage}

  \caption{\textbf{Generalization across objects.}
  \textbf{Left:} Training objects per \texttt{MolmoSpaces} split. Each split contains all objects of the previous one; $+$ marks objects added. The 1-task split is trained separately on a single \emph{box} or \emph{spatula} task. Full task details are in Table~\ref{tab:molmo_splits}.
  \textbf{Right:} We train each method on tasks involving a single object (box or spatula) and evaluate on the unseen knife. For methods that use a critic (BoN, OGPO), training on a similar object transfers better, whereas FSFT shows no such pattern. We report the mean and 95\% stratified bootstrap CIs across three seeds.}
  \label{fig:generalization_analysis}
\end{figure}

\chg{
\textbf{Critic learning enables generalization on unseen objects that resemble training objects.}
Figure~\ref{fig:generalization_analysis} illustrates this directly: for BoN, training on a similarly shaped object transfers better (e.g., boxes to bowls, spatulas to knives), whereas FSFT shows no such pattern. 
The per-object results (Figure~\ref{fig:generalizaton_per_objects}) show the same trend at scale. BoN improves on atomizers and wine bottles mainly at 144 tasks (atomizer $0.29 {\rightarrow} 0.42$, wine bottle $0.24 {\rightarrow} 0.44$), the split that adds bottle-like objects such as soap dispensers and salt/pepper shakers (Figure~\ref{fig:generalization_analysis}, left). 
Forks and knives show substantial improvement with OGPO when we move from 1 to 4 tasks; at 4 tasks each training split contains spatulas, which require similar grasping motion to forks and knives; we believe this enables this generalization. The performance on forks and knives plateaus at 64 tasks, however. We hypothesize this is because the objects added between 64 and 144 tasks do not exhibit significant similarity to forks and knives.
OGPO shows a similar pattern on atomizers ($0.42 {\rightarrow} 0.58$) but not on wine bottles ($0.09 {\rightarrow} 0.12$), while FSFT improves on neither. Since BoN keeps the policy frozen, the critic alone suffices to transfer to similar objects, while policy updates alone do not. Across all objects, we see that critic learning is essential to generalization---with the exception of bowls, we observe substantial generalization only when applying critic learning.
}

\chg{
\textbf{Policy updates help on some objects but degrade performance on others.} FSFT and OGPO both improve on some objects (e.g., bowls) but degrade on wine bottles, where both fall sharply below the base policy while BoN recovers to about 0.44 at 144 tasks. 
We find that policy updates can enable more effective generalization to objects similar to those in the training distribution (e.g., OGPO on forks), while leading to poor generalization and even forgetting on objects further from the training distribution (e.g. wine bottles). In contrast, critic-only updates can enable effective generalization even on objects far from the training distribution (e.g., BoN on wine bottle). 
}

\subsection{\chg{Q4:} What enables generalist policy improvement?}
\label{sec:analysis}

Having evaluated our recipes at scale, we now validate key design decisions for multi-task VLA post-training, and investigate several \chg{alternative techniques for critic-guided policy extraction}; see Appendix~\ref{app:additional_exp} for additional ablations including data/compute scaling and resource allocation.

\begin{figure}[t]
    \centering
    \includegraphics[width=1.0\linewidth]{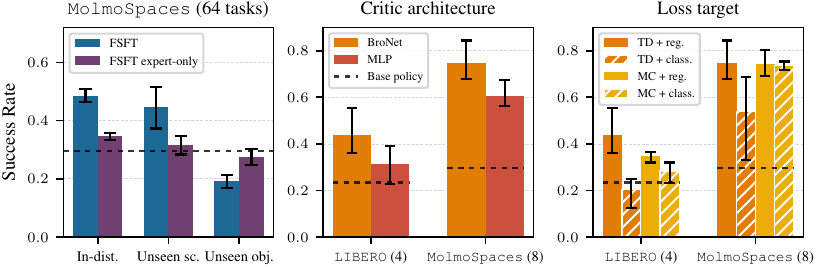}
    \caption{\textbf{Recipe ablations.} Left: full VLA vs.\ expert-only tuning for FSFT on \texttt{MolmoSpaces} (64 tasks). Full tuning is essential in-distribution and on unseen scenes, but degrades most on unseen objects. Center: architecture for BoN. Right: TD vs MC, classification vs regression for BoN. \chg{Center and right are
    evaluated on \texttt{LIBERO} (4 tasks; see Table~\ref{tab:libero_splits}) and the object-diverse \texttt{MolmoSpaces} (8 objects, 1 scene each; see Table~\ref{tab:diversity_splits}).}}
    \label{fig:ablation}
\end{figure}

\subsubsection{What Makes Direct Policy Adaptation Effective?}

An important fine-tuning decision concerns which parameters to update: the entire VLA or only the action expert, i.e., the flow model mapping tokens from the backbone to continuous actions. We compare both settings for FSFT on the \texttt{MolmoSpaces} (64 tasks) split, evaluating in-distribution, on unseen scenes, and on strongly OOD unseen objects. The left panel of Figure~\ref{fig:ablation} shows that \emph{\textbf{full VLA fine-tuning is beneficial}}: it outperforms expert-only tuning by more than $+12$\,pp both in-distribution and on unseen scenes, whereas expert-only tuning improves only marginally over the base policy. This comes at a cost: on unseen objects, full VLA tuning degrades performance by $10.6$\,pp relative to the base policy, against $-2.2$\,pp for expert-only tuning. This suggests that full fine-tuning increases overfitting risks. The two settings therefore trade in-distribution gains against OOD performance.
\begin{wrapfigure}[20]{r}{0.44\linewidth}
  \vspace{7mm}
  \centering
  \includegraphics[width=0.88\linewidth]{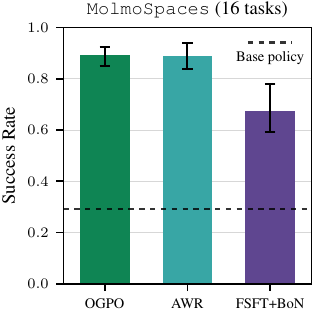}
      \caption{\chg{\textbf{Policy extraction schemes.} We compare OGPO~\citep{patil2026ogpo} to AWR~\citep{peng2019advantage} and a direct combination of FSFT and BoN. We find that natively actor-critic algorithms perform best.
      }} 
  \label{fig:policy_extraction}
\end{wrapfigure}
\subsubsection{What Makes a Good Value Function in VLA Post-Training?}
\label{sec:exp_value}
Value function learning admits many possible implementations. To find a scalable recipe for multi-task VLA post-training, we conduct controlled ablations along multiple design axes, focusing on training-tasks improvement.
\chg{We will use BoN as a representative algorithm, arguably being the simplest recipe involving a critic.}
\Cref{fig:ablation} (center, right) compares two network architectures, regression and classification objectives, as well as TD and MC targets.
Compared to standard MLP critics, adding \textbf{residual connections}~\citep[BroNet;][]{nauman2024bigger} increases performance by more than $+10$\,pp.
Regarding the training objective, combining \textbf{TD targets with a simple regression loss} yields the largest and most consistent improvements in our experiments.
Notably, we find distributional critic formulations to scale poorly to multi-task settings; we hypothesize this is due to return distributions of different tasks varying significantly. Finally, \chg{BoN} introduces one further hyperparameter, the number of
sampled candidates $N$. We ablate its impact on performance and inference time
in Appendix~\ref{app:bon_n_ablation}, and use $N=32$ throughout.

\subsubsection{\chg{Recipes for Policy Extraction}}
\label{sec:policy_extraction} 

\looseness=-1
\chg{
Through our experiments, we find that OGPO~\citep{patil2026ogpo} largely retains the generalization properties of BoN while enabling expressive policy extraction, and therefore reaches significantly higher performance.
}

\chg{
As such, OGPO, and critic-guided policy extraction in general, can be interpreted as the combination of the other two mechanisms, namely direct policy adaptation and critic-guided test-time refinement, where the latter is applied at training time instead.
OGPO is, however, one of many ways in which this combination can be instantiated.
We thus ablate two alternatives
in Figure~\ref{fig:policy_extraction}.
As a straightforward instantiation, we consider FSFT+BoN, which simultaneously fine-tunes the policy on successful rollouts and trains a critic for inference-time action selection.
}
\chg{
We also evaluate advantage-weighted regression ~\citep[AWR;][]{peng2019advantage}, a popular technique which weights the objective in Equation~\ref{eq:weighted_supervised} by exponentiated advantages, thus implementing a form of regularized policy improvement.
We find that all three schemes are effective. Yet, OGPO and AWR generally outperform a straightforward combination of FSFT and BoN; we hypothesize this is due to a finer implementation of policy improvement.
}

\section{Discussion}

\looseness=-1This work studies the adaptation regimes of generalist policies as the diversity of their set of fine-tuning tasks increases.
We isolate \chg{three} techniques for self-improvement, centered respectively on updating the policy parameters, on training value estimates \chg{and refining actions at test-time, and on extracting a better policy according to those value estimates}.
While \chg{all} mechanisms \chg{improve performance on} training tasks, we find that, at a sufficient scale, value-based RL is instrumental in inducing a degree of generalization beyond observed tasks, and in achieving generalist improvement of generalist policies. We believe these results show a promising way forward \chg{for efficiently improving} generalist robot policies, moving beyond expensive single-task improvement to enable improvement through generalization on diverse training tasks.

\looseness=-1 \textbf{Limitations. \;}
Our contributions are empirical in nature.
As such, while we dedicated significant effort and compute to broadening the scope of our evaluation, it remains limited to manipulation environments, which currently form a major domain of application for VLAs.
Further studies on loco-manipulation and whole-body control \citep{gr00tn1_2025} remain important directions for future work.
Moreover, the evaluation was designed with the constraint that the base policy (in our case, $\pi_{0.5}$) would make non-trivial progress on training tasks out of the box: for this reason, including tasks of arbitrary complexity would be less informative.
Finally, given the large scale of our evaluation, and the number of diverse tasks required, it remains constrained to simulated environments.

The main goal of this work is thus to contribute to the understanding of mechanisms and trends of online self-improvement in dynamical systems.
We believe that our results represent a promising starting point for future works, which can bootstrap off the simple and scalable RL recipes we derived, and study the benefits of multi-task post-training at scale on real hardware.

\clearpage

\subsection*{Reproducibility statement}

We ensure reproducibility by releasing code for our full training pipeline at the URL listed at the end of Section~\ref{sec:exp_setup}. All data it uses is collected online, and no offline dataset is necessary to reproduce results.
Moreover, we thoroughly report training details, hyperparameters and task splits in Appendix~\ref{app:experimental_details}.

\subsection*{Acknowledgements}
This work was supported under project IDs 143, 037 and 0220 as part of the Swiss AI Initiative, through a grant from the ETH Domain and computational resources provided by the Swiss National Supercomputing Centre (CSCS) under the Alps infrastructure.
M.B. and L.D.B. were supported by the Max Planck ETH Center for Learning Systems.
R.R. was supported by the German Research Foundation (DFG) within the RTG project ConVeY funded by grant GRK 2428.
Y.A. received funding from grant no. 21039 of the Hasler Foundation and the ETH AI Center.
M.C. was supported by the National Science Centre under Grant No. 2023/50/E/ST6/00665.
D.M. was supported by the Knut and Alice Wallenberg Foundation.
This work was supported by the German Federal Ministry of Research, Technology and Space (BMFTR) under the Robotics Institute Germany (RIG) funded by BMFTR grant 16ME0997K.

\clearpage
\bibliography{refs}
\bibliographystyle{iclr2027_conference}

\appendix

\newpage

\section{Extended Related Work}
\label{app:extended_related_work}

\textbf{Connections to LLM post-training.\;}
\looseness=-1
A closely related pattern to the ones we observe appears in LLM post-training. For instance, increasing the diversity of instruction-finetuning tasks improves generalization to held-out tasks~\citep{wei2021finetuned,iyer2022opt,longpre2023flan,chung2024scaling}, with evidence of approximately log-linear scaling with task coverage~\citep{longpre2023flan} and stronger gains for larger models~\citep{chung2024scaling}. These works primarily study supervised fine-tuning on fixed datasets, whereas we consider \emph{online self-improvement} through reinforcement learning and study how its benefits scale with task diversity---a question that, to our knowledge, has not been systematically investigated in robotics. 
More directly, recent work comparing RL and supervised fine-tuning finds that RL can generalize better to unseen task variants~\citep{chu2025sftmemorizesrlgeneralizes} and better preserve previously acquired capabilities during post-training~\citep{shenfeld2026rl}. These observations relate to our results in \Cref{fig:main}, where FSFT similarly exhibits weaker generalization and degraded performance on the training tasks compared to BoN.

\newpage
\section{A Taxonomy of Post-Training Algorithms}
\label{app:taxonomy}

\begin{wrapfigure}[20]{r}{0.40\linewidth}
  \vspace{-6mm}
\begingroup
  \definecolor{mtblue}{HTML}{1D6996}
  \definecolor{mtorange}{HTML}{E17C05}
  \resizebox{\linewidth}{!}{%
  \begin{tikzpicture}[
    font=\rmfamily\fontsize{10}{12}\selectfont,
    lobe/.style={line width=1.6pt},
    title/.style={font=\rmfamily\bfseries\fontsize{11}{12.5}\selectfont,
      align=center},
    entry/.style={inner sep=1.2pt},
    ours/.style={entry, font=\rmfamily\itshape\fontsize{10}{12}\selectfont},
    split/.style={densely dashed, draw=black!55, line width=0.9pt},
    splitlbl/.style={font=\rmfamily\fontsize{8}{9.5}\selectfont,
      text=black!65, inner sep=1.5pt, align=right}
  ]
\path[use as bounding box] (-3.78,-2.82) rectangle (3.78,4.02);
  \def\splity{0.00}
  \begin{scope}
    \clip (-1.0,0) circle (2.6);
    \draw[split] (-3.70,\splity) -- (1.70,\splity);
  \end{scope}

  \draw[lobe, draw=mtblue]   (-1.0,0) circle (2.6);
  \draw[lobe, draw=mtorange] ( 1.0,0) circle (2.6);

  \node[title, text=mtblue]   at (-2.00,3.32) {Direct Policy\\Adaptation};
  \node[title, text=mtorange] at ( 2.00,3.32) {Critic-Guided\\Refinement};

  \node[ours]  at (-2.35, 0.90) {FSFT};
  \node[entry] at ( 0.00, 1.25) {FSFT+BoN};
  \node[entry] at (-0.62, 0.55) {RECAP};
  \node[entry] at ( 0.62, 0.55) {PARL};

  \node[entry] at (-2.35,-0.90) {DPPO};
  \node[entry] at ( 0.00,-1.25) {EXPO};
  \node[entry] at (-0.62,-0.55) {AWR};
  \node[entry] at ( 0.62,-0.55) {\textit{OGPO}};

  \node[ours] at (2.35,0.00) {BoN};
  \end{tikzpicture}%
  }
\endgroup
\vspace{-6mm}
  \caption{\textbf{A taxonomy of a few VLA post-training approaches.} Approaches above the dashed line adopt a purely supervised objective for policy extraction. We italicize the three approaches at the center of the analysis in Sections ~\ref{sec:q1} and~\ref{sec:q2}.}
  \label{fig:taxonomy}
\end{wrapfigure}
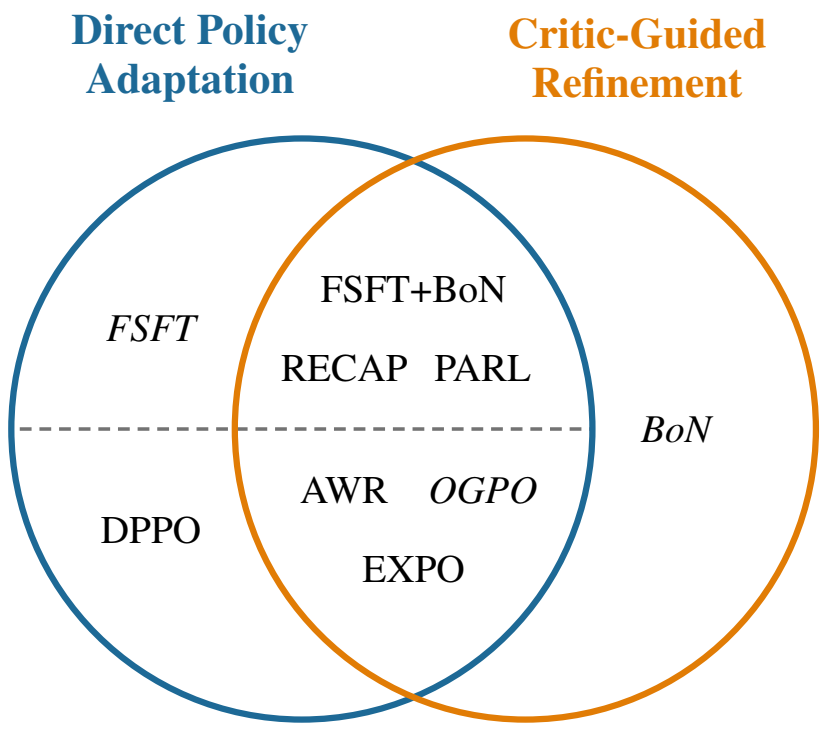

Many algorithms have been proposed for finetuning pretrained generative control policies, such as Vision-Language-Action (VLA) models.
We review a few representative approaches and connect them to the two policy improvement mechanisms identified in \Cref{sec:method}: direct policy adaptation and critic-guided refinement.
As displayed in Figure~\ref{fig:taxonomy}, we distinguish whether these methods use value estimates and perform policy updates, and whether they use an unweighted supervised loss for policy extraction.
As FSFT, BoN and FSFT+BoN are already introduced in Section~\ref{sec:method}, we focus on the remaining methods listed in Figure~\ref{fig:taxonomy}.
This list is not meant to be exhaustive, but rather a representative comparison of existing methods. For simplicity, we will absorb the task conditioning in the state.

\paragraph{DPPO.}
\label{par:dppo}
Diffusion Policy Policy Optimization \citep{ren2025diffusion} is designed for generative policies involving a denoising process $a_{K:0} = (a_{K},\dots,a_{0})$: starting from $a_{K}\sim\mathcal{N}(0,\mathbf{I})$, the generative model samples $a_{k-1}\sim \pi_\theta(\cdot \mid s, a_{k},k,c) \quad \forall k=K,\dots, 1$, until it arrives at the generated chunk $a_0$.
DPPO applies on-policy PPO to a two-layer MDP, embedding the inner denoising MDP within the outer environment MDP.
Policy gradients use tractable Gaussian likelihoods at each denoising step, with environment-step advantages estimated from on-policy rollouts and a state-value baseline. This allows policy extraction through simple PPO updates. For $k=1,\dots,K$,
\begin{equation}
    \omega_{k}(\theta)=\frac{\bar\pi_\theta(a_{k-1}\mid s,a_{k},k)}{\bar\pi_{\theta_{\mathrm{EMA}}}(a_{k-1}\mid s,a_{k},k)},
    \quad 
    \widehat{A}_{k}=\gamma_{\mathrm{den}}^{k-1}
    \widehat{A}^{\mathrm{env}},
\end{equation}

\begin{equation}
\mathcal L_{\mathrm{DPPO}}(\theta)= -\mathbb E \left[\sum_{k=1}^{K} C_\epsilon\!\left(\omega_{k}(\theta),\widehat A_{k}\right)\right],
\end{equation}

where $\theta_{\mathrm{EMA}}$ is an exponential moving average of the policy parameters,  $\widehat A^{\mathrm{env}}$ is estimated using environment rewards and a learned value baseline, $\gamma_{\mathrm{den}}$ discounts credit to earlier denoising transitions and the PPO surrogate is $C_\epsilon(\omega,A)=\min\{\omega A,\operatorname{clip}(\omega, 1-\epsilon, 1+\epsilon)A\}$.
We do not classify this method as leveraging critic-based guidance, as the learned value function is not action-conditional, and is mostly used for variance reduction rather than for evaluating actions.
Nevertheless, it updates the policy parameters, and thus performs direct policy adaptation.

\paragraph{RECAP.}
RL with Experience and Corrections via Advantage-conditioned
Policies \citep{intelligence2025pi} learns a state-only distributional critic from Monte Carlo (MC) returns and converts estimated advantages into binary conditioning labels.
The policy is then \textit{conditioned} on these labels, and trained with a supervised flow matching loss on both successful and unsuccessful trajectories.
With the original method's undiscounted $n$-step estimation, the advantage and the advantage labels for some state $s_0$ are expressed as
\begin{equation}
    \widehat{A}=\sum_{j=0}^{n-1}r_{t+j} + V_\eta(s_{n}) - V_\eta(s_t), \quad I_t=\mathbf{1}\{\widehat{A}>\varepsilon\},
\end{equation}
for some threshold $\varepsilon$.
As such, this method combines a form of critic-based guidance (through advantage conditioning) with policy adaptation through conditional behavior cloning. 

\paragraph{AWR.} Advantage Weighted Regression~\citep{peng2019advantage} is an actor-critic algorithm which avoids differentiation through the critic, and updates the actor through a weighted supervised objective. Its early instantiations rely on a critic trained through Monte Carlo return estimation, but AWR has since been applied widely, e.g., with temporal-difference critics in offline RL~\citep{park2025ogbenchbenchmarkingofflinegoalconditioned}.
With the notation introduced in Section~\ref{sec:method}, the advantage-weighted policy extraction objective is
\begin{equation}
    \mathcal{L}_\text{AWR}(\theta) = -\,\mathbb{E}_{(s,a)\sim\mathcal{D}} \big[ e^{\frac{\widehat A(s,a)}{\beta}}\log \pi_\theta(a|s)\big], \label{eq:weighted_supervised_app}
\end{equation}
where $\widehat A$ is the advantage, often computed from $Q_\psi(s,a)$, and $\beta$ is a temperature parameter that recovers pure behavior cloning as $\beta \to \infty$.

\paragraph{PARL.} Policy Agnostic RL~\citep{mark2024policy} trains a state-action critic $Q_\psi(s,a)$, originally with either IQL~\citep{kostrikov2021offline} or Cal-QL~\citep{nakamoto2024calqlcalibratedofflinerl}, while policy extraction relies on two steps: action optimization, and policy updates.
The former consists of sampling action candidates from the current policy, and leveraging first- and zeroth-order optimization to approximately find $a^\star = \arg \max_{a} Q_\psi (s,a)$ for each state and task in the training batch.
The latter simply encourages the policy to regress the best candidate through unweighted behavior cloning (see Equation~\ref{eq:weighted_supervised}).
PARL is thus a close neighbor of FSFT+BoN; PARL optimizes over actions at training-time instead of test-time, and trains its policy on both successful and unsuccessful trajectories.

\paragraph{EXPO.}
Expressive Policy Optimization ~\citep{dong2026expostablereinforcementlearning} combines a critic trained through TD learning with a policy improvement scheme that integrates an expressive base policy with a lightweight edit policy. The edit policy is trained to maximize Q-values with a SAC-style objective~\citep{haarnoja2018softactorcriticoffpolicymaximum}, while the base policy is updated according to a simple behavior cloning loss.
Crucially, EXPO integrates a BoN-like mechanism, both for selecting actions during evaluation, and for computing TD targets.
Thus, EXPO is also closely connected to FSFT+BoN, mostly differing in the introduction of an edit policy, and in the slightly different recipe used for value learning. \looseness -1

\paragraph{OGPO.} Off-Policy Generative Policy Optimization (OGPO) \citep{patil2026ogpo} decouples the bi-level MDP formulation proposed in \citet{ren2025diffusion} and uses off-policy Q functions with TD-learning for the environment MDP, while using on-policy PPO-style policy extraction~\citep{schulman2017proximal} for the denoising MDP of the generative control policy. For each replay state from the replay buffer, OGPO rolls out a group of $G$ stochastic actions to obtain group-relative advantages. This mitigates the need for first-order optimization through the Q-functions and allows zeroth-order updates by conditioning the likelihoods via the PPO objective keeping the critics frozen during policy extraction:
\chg{
\begin{equation}\label{eq:ogpo_group_advantage}
    \widehat{A_i}=Q_{\mathrm{targ}}(s,a^{i,(0)}) - \frac{1}{G}\sum_{j=1}^{G}Q_{\mathrm{targ}}(s,a^{j, (0)}), \quad \omega_i(\theta) = \prod_{k=1}^K\frac{\pi_\theta(a^{i,(k-1)}\mid s,a^{i,(k)},k)}{\pi_{\theta_{\mathrm{slow}}}(a^{i,(k-1)}\mid s,a^{i,(k)},k)}
\end{equation}
}and
\begin{equation}\label{eq:ogpo_policy_update}
    \mathcal{L}_{\mathrm{OGPO}}(\theta)= -\mathbb{E}_{\mathcal{D},\theta_{\mathrm{ref}}} \left[\frac{1}{G}\sum_{i=1}^{G} C_\epsilon\!\left(\omega_i(\theta),\widehat{A_i}\right)\right].
\end{equation}
Here, $\widehat{A_i}$ is the group relative advantage, $\omega_i(\theta)$ is the annealed importance sampling ratio, $Q_\mathrm{targ}$ is the EMA version of the online $Q$ function, and similarly environment rollouts are conducted via $\pi_{\textsc{ema}}$.
\chg{
Moreover, $(a^{i,(K)},\dots,a^{i, (0)})_{i = 1}^G \sim \pi_{\theta}(\cdot \mid s)$ is a group of denoising sequences, and $\pi_{\theta}(a^{(k-1)} \mid s,a^{(k)})$ denotes denoising-step likelihoods.
}
OGPO proposes an ODE-to-SDE conversion of pretrained flow policies that allows the log probability updates on pretrained VLAs.
\chg{
Via these likelihoods,  $\omega_i(\theta)$ captures the  importance  ratio between the denoising trajectories under current policy $\pi_{\theta}$, and a slower-updating policy $\pi_{\theta_{\mathrm{slow}}}$ computed via a parameter exponential-moving average. Finally, $C_\epsilon$ is a  clipping function   $C_\epsilon(\omega,A) := \min\{\omega A,\operatorname{clip}(\omega, 1-\epsilon, 1+\epsilon)A\}$ \citep{schulman2017proximal}. Informally, this update up-weights denoising sequences which lead to positive $\hat A$, down-weights those which lead to negative $\hat A$, and incorporates clipping $C_\epsilon$ to mitigate aggressive updates where the policy is  uncertain.
}
OGPO also proposes two key stabilization techniques: (1) a success buffer for FSFT on the base objective, and (2) conservative advantages that stabilize the gradient updates through underfit Q-functions in the beginning of RL-finetuning. Finally to improve critic accuracy, OGPO incorporates Q-variance reduction which averages the TD targets over $N_{vr}$ actions sampled from the actor. 
\begin{equation}\label{eq:ogpo_td_qvr}
    \mathcal{L}_{\mathrm{critic}} = \mathbb{E}\left[\left(Q(s_t , a_t ) -  \left(r_t + \gamma 
    \frac{1}{N_{vr}}\sum_{i=1}^{N_{vr}} Q(s_{t+1}, a^{(i)}_{t+1})\right)\right)^2 \right],
a^{(i)}_{t+1} \sim \pi_\theta(\cdot\mid s_{t+1}).
\end{equation}

\newpage
\section{Additional Empirical Results}
\label{app:additional_exp}

\subsection{Generalist improvement on \texttt{LIBERO}: does generalization occur for completely novel tasks?}
\label{app:libero}

\begin{wrapfigure}[21]{h}{0.66\linewidth}
\vspace{-5mm}
  \includegraphics[width=\linewidth]{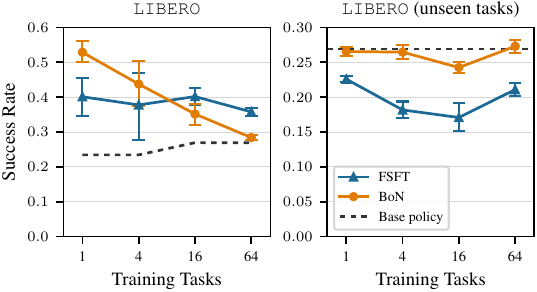}
  \vspace{-4mm}
  \caption{\textbf{Value-based generalist improvement is limited when tasks are loosely related.} Performance of FSFT and BoN on training tasks (left) and held-out tasks (right) in \texttt{LIBERO} as the training tasks grow in number. The single-task lead of BoN reverses in larger multi-task experiments, but FSFT exhibits substantial forgetting on held-out tasks, unlike BoN.} 
  \label{fig:main_libero}
\end{wrapfigure}

We extend the evaluation reported in Figure~\ref{fig:main} to \texttt{LIBERO}, in terms of performance on seen and unseen tasks.
\chg{For simplicity, we only consider the two most fundamental mechanisms out of the three introduced in Section~\ref{sec:method}.}
When evaluated on seen tasks (Figure~\ref{fig:main_libero}, left), both direct policy adaptation (FSFT) and critic-guided refinement (BoN) are generally able to improve success rates with respect to the base policy. As observed in Figure~\ref{fig:main}, performance of multi-task models does not reach that of single-task experts, particularly for BoN, which barely improves over the base policy when trained over $64$ tasks.

The evaluation on unseen tasks (Figure~\ref{fig:main_libero}, right) is instead distinct from the one in Figure~\ref{fig:main}, and highlights the limits of generalization to unseen tasks.
Here, the degree of diversity of tasks is harder to control, and the variability across tasks is higher.
Unseen tasks often involve completely different setups, motion patterns and goals compared to the training tasks.
At the scale we train on~($N_{\mathrm{task}}\leq 64$), generalization does not emerge.
Nevertheless, BoN remains preferable to FSFT, as it only displays minimal forgetting. 

\subsection{Extended Scaling Task Diversity Ablation }
\label{app:scaling_extended}
\begin{figure}[h]
    \centering
    \includegraphics[width=\linewidth]{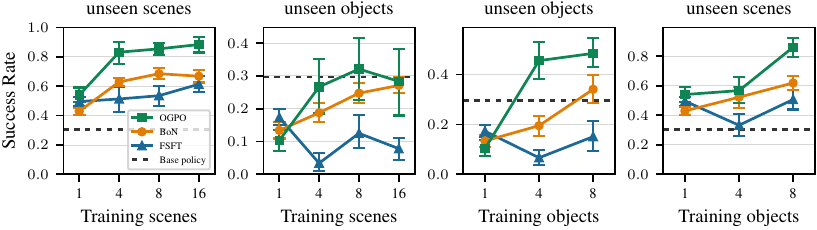}
    \caption{\textbf{Diversify along the axis of desired generalization.} Scaling
    scenes of a single object (\textit{left two}) or objects with one scene each
    (\textit{right two}), evaluated on unseen scenes and unseen objects. Each axis
    transfers best along itself, and BoN scales better than FSFT, which falls below
    the base policy on unseen objects. Mean over three seeds, 95\% stratified
    bootstrap CIs.}
    \label{fig:scaling_ablation}
\end{figure}

We extend Figure~\ref{fig:obj_vs_scenes} \chg{by reporting trends for BoN and} FSFT, and specify the task splits
in full. We scale the number of scenes of a single object, $\{1, 4, 8, 16\}$,
and separately the number of objects, $\{1, 4, 8\}$, with one scene each,
evaluating both on eight held-out scenes of that object and on eight unseen
objects. \texttt{MolmoSpaces} has 18 objects, of which we reserve 8 for
evaluation, so at most 10 remain for training, explaining why the object sweep has
no 16-object split. Table~\ref{tab:diversity_splits} lists the splits. 

As shown in Figure~\ref{fig:scaling_ablation}, scaling scenes improves
performance on unseen scenes by $+36.5$\,pp for BoN and $+31$\,pp for FSFT, but
does not transfer across objects: both fall below the base policy there,
although BoN recovers with more scenes while FSFT does not. Scaling objects
eventually transfers: BoN falls below the base policy at $1$ and $4$ objects,
but overtakes it at $8$. At a matched budget of eight tasks each axis transfers
best along itself, by $+6.6$\,pp on unseen scenes and $+9.4$\,pp on unseen
objects. It is therefore beneficial to diversify along the axis of desired
generalization, though object diversity is more effective overall.

\begin{table}[h]
  \centering
  \caption{\textbf{Task splits for the diversity ablation.} Task IDs are given
  without the \texttt{molmo\_} prefix and grouped by object;
  \emph{Init.\ SR} is the mean success rate of the base policy over a split.
  Scaling scenes varies the number of scenes of a single object; scaling
  objects varies the number of objects, with one scene each. Both regimes are
  evaluated on the same two held-out splits: eight unseen scenes of box, the
  one object common to every training split, and one scene of each of eight
  objects never seen during training. We note that the \texttt{MolmoSpaces} (8 tasks) split is used in Figures~\ref{fig:ablation} and \ref{fig:lr_ablation}.}
  \label{tab:diversity_splits}
  \vspace{2mm}
\begingroup
  \definecolor{mtorange}{HTML}{E17C05}
  \begin{tikzpicture}[
    x=1pt, y=1pt,
    font=\rmfamily,
    cell/.style={inner sep=0pt, anchor=base},
    lbl/.style={cell, anchor=base west},
    grp/.style={cell, anchor=base west, font=\rmfamily\itshape, text=black!55},
    sub/.style={cell, anchor=base west, text width=226pt, align=left},
    head/.style={cell, text=mtorange},
    rule/.style={draw=mtorange, line width=0.9pt},
    midrule/.style={draw=mtorange!60, line width=0.5pt},
    sep/.style={draw=mtorange!25, line width=0.4pt}
  ]
  \def\W{397}

  \fill[mtorange!5] (0,-22) rectangle (\W,0);
  \draw[rule]    (0,0)   -- (\W,0);
  \draw[midrule] (0,-22) -- (\W,-22);

  \node[head, anchor=base west] at (3,-14.5) {Split};
  \node[head] at (258,-14.5) {Tasks};
  \node[head] at (312,-14.5) {Objects};
  \node[head] at (370,-14.5) {Init. SR (\%)};

  \node[grp] at (3,-34.5) {Scaling scenes};
  \node[lbl]  at (14,-49.5) {1 task};
  \node[cell] at (258,-49.5) {4};
  \node[cell] at (312,-49.5) {1};
  \node[cell] at (370,-49.5) {29.2};
  \node[sub] at (25,-62.5) {\emph{box}: 267, 366, 138, 143};
  \node[lbl]  at (14,-78.5) {4 tasks};
  \node[cell] at (258,-78.5) {4};
  \node[cell] at (312,-78.5) {1};
  \node[cell] at (370,-78.5) {29.2};
  \node[sub] at (25,-91.5) {\emph{box}: 267, 366, 138, 143};
  \node[lbl]  at (14,-107.5) {8 tasks};
  \node[cell] at (258,-107.5) {8};
  \node[cell] at (312,-107.5) {1};
  \node[cell] at (370,-107.5) {29.2};
  \node[sub] at (25,-120.5) {\emph{box}: 164, 86, 70, 267, 366, 138, 143, 212};
  \node[lbl]  at (14,-136.5) {16 tasks};
  \node[cell] at (258,-136.5) {16};
  \node[cell] at (312,-136.5) {1};
  \node[cell] at (370,-136.5) {33.9};
  \node[sub] at (25,-149.5) {\emph{box}: 207, 393, 146, 55, 164, 86, 70, 267, 366, 138, 143, 212, 351, 333, 289, 122};
  \node[grp] at (3,-177.5) {Scaling objects};
  \node[lbl]  at (14,-192.5) {1 task};
  \node[cell] at (258,-192.5) {4};
  \node[cell] at (312,-192.5) {1};
  \node[cell] at (370,-192.5) {29.2};
  \node[sub] at (25,-205.5) {\emph{box}: 267, 366, 138, 143};
  \node[lbl]  at (14,-221.5) {4 tasks};
  \node[cell] at (258,-221.5) {4};
  \node[cell] at (312,-221.5) {4};
  \node[cell] at (370,-221.5) {29.2};
  \node[sub] at (25,-234.5) {\emph{box}: 138, \emph{cup}: 24, \emph{tissuepaper}: 221, \emph{spatula}: 373};
  \node[lbl]  at (14,-250.5) {8 tasks — \texttt{MolmoSpaces} (8 tasks)};
  \node[cell] at (258,-250.5) {8};
  \node[cell] at (312,-250.5) {8};
  \node[cell] at (370,-250.5) {29.7};
  \node[sub] at (25,-263.5) {\emph{boiler}: 369, \emph{box}: 138, \emph{cup}: 24, \emph{ladle}: 187, \emph{pot}: 325, \emph{tissuepaper}: 221, \emph{spatula}: 373, \emph{spoon}: 148};
  \node[grp] at (3,-291.5) {Evaluation};
  \node[lbl]  at (14,-306.5) {Unseen scenes};
  \node[cell] at (258,-306.5) {8};
  \node[cell] at (312,-306.5) {1};
  \node[cell] at (370,-306.5) {30.3};
  \node[sub] at (25,-319.5) {\emph{box}: 666, 602, 543, 273, 916, 505, 769, 675};
  \node[lbl]  at (14,-335.5) {Unseen objects};
  \node[cell] at (258,-335.5) {8};
  \node[cell] at (312,-335.5) {8};
  \node[cell] at (370,-335.5) {29.7};
  \node[sub] at (25,-348.5) {\emph{atomizer}: 140, \emph{bowl}: 356, \emph{fork}: 345, \emph{knife}: 155, \emph{remotecontrol}: 357, \emph{salt/pepper shaker}: 163, \emph{soapdispenser}: 208, \emph{winebottle}: 226};

  \draw[sep] (0,-168.0) -- (\W,-168.0);
  \draw[sep] (0,-282.0) -- (\W,-282.0);
  \draw[rule] (0,-380.5) -- (\W,-380.5);
  \end{tikzpicture}
\endgroup
\end{table}

\subsection{Allocating resources: data and compute}
\label{sec:allocation}

\begin{figure} [h]
    \centering
    \includegraphics[width=\linewidth]{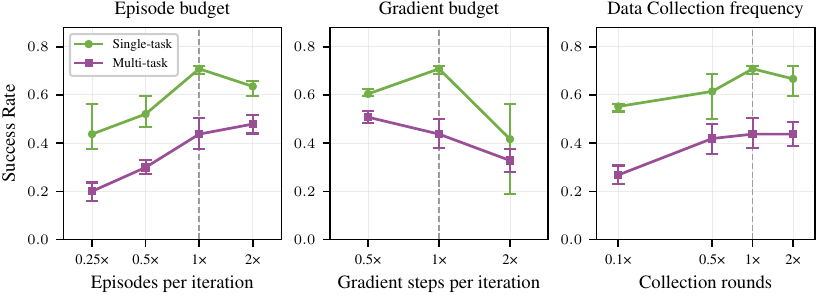}
    \caption{\textbf{Post-training benefits from more data and more frequent iterations.} Final success rates for critic-guided refinement (BoN) on \texttt{LIBERO}. Each subplot changes one hyperparameter: the number of episodes collected per iteration (left), the number of gradient steps per iteration (middle) and the number of iterations (right) over which a data and compute budget is distributed. The other hyperparameters are fixed per plot. $1\times$ marks the values used across the standard experimental protocol described in Section~\ref{sec:exp_setup}. Collecting more data is beneficial with diminishing gains, while more updates may be harmful; frequent iterations beat a single update phase after data collection.}
    \label{fig:allocation}
\end{figure}

Online post-training involves three important design axes when it comes to resource allocation: (i) how much data should be collected, (ii) how many gradient steps should be applied and (iii) how data collection and gradient updates should be interleaved.
We thus consider these three axes, and validate the main experimental protocol outlined in Section~\ref{sec:exp_setup} \chg{for BoN}: post-training for $10$ iterations, each alternating data collection ($20$ episodes per task) and model updates (100-500 gradient steps for the policy, $100\times$ for the critic).
While we present this ablation for BoN in \texttt{LIBERO}, we observed similar trends across algorithms and environments.

As shown in the first column of ~\Cref{fig:allocation}, we first study how performance changes with the amount of data collected per iteration, while keeping the number of iterations and of gradient steps fixed.
We observe that in both single- and multi-task settings (adopting the $4$-task split in this case), performance increases at a decreasing rate. We thus adopt the highest budget which brings significant benefits --- $20$ episodes per task.
The second column keeps the number of episodes and iterations fixed instead, and ablates the amount of compute that should be allocated to training.
In this case, we observe that more training is not always helpful, suggesting that the underlying training dynamics of online RL are not entirely stable. For these reasons, we settle on an intermediate budget.
Finally, in the third column, we keep the amount of data and compute fixed, and we purely validate the number of iterations that should be applied.
The leftmost point of the plot collects all the data at once, and follows this single data collection phase by a lengthy update phase, exhausting the entire gradient budget.
The rightmost point instead distributes data and compute budgets across a large number of shorter iterations, each collecting $1/20$th of the data and applying $1/20$th of the gradient steps.
Overall, we observe a gentle scaling trend which plateaus at high iteration frequencies, suggesting that frequent iterated online fine-tuning is beneficial.
Accordingly, we post-train our base model over $10$ iterations.

\newpage
\subsection{Does a multi-task pre-trained critic transfer to new tasks?}

\begin{wrapfigure}[19]{r}{0.66\linewidth}
\vspace{-5mm}
  \includegraphics[width=\linewidth]{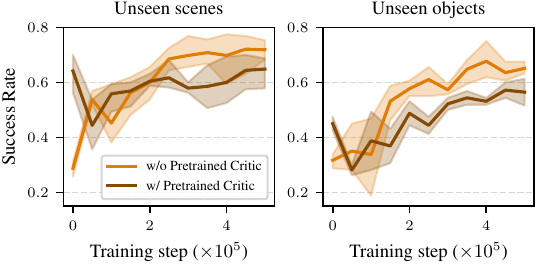}
  \caption{\textbf{Pretrained critic for accelerating value learning.} A critic
pretrained on \texttt{MolmoSpaces} (64 tasks) starts higher than one trained
from scratch, but the gap closes within one iteration and then reverses, on both
unseen scenes (\textit{left}) and unseen objects (\textit{right}). Mean and 95\%
stratified bootstrap CIs across three seeds.} 
  \label{fig:pre-train}
\end{wrapfigure}

A natural question in our setting is whether a critic pretrained on many tasks can accelerate value learning on unseen ones. We compare BoN with a critic pretrained on the \texttt{MolmoSpaces} (64 tasks) split against one trained from scratch, evaluating on unseen scenes of the pretrained critic's training objects and on unseen objects. Figure~\ref{fig:pre-train} shows that pretraining helps only initially: the pretrained critic starts at a substantially higher success rate, but the gap closes in the next iteration, after which it trails the randomly initialized critic on both splits. These results indicate that pretrained critics offer little benefit for accelerating value learning and can even be detrimental. One possible explanation is a loss of plasticity in the pretrained critic, though we leave the identification of the precise mechanism to future work.

\subsection{Stability in direct policy adaptation}

\begin{figure}[h]
    \centering
    \vspace{-2mm}
    \includegraphics[width=\linewidth]{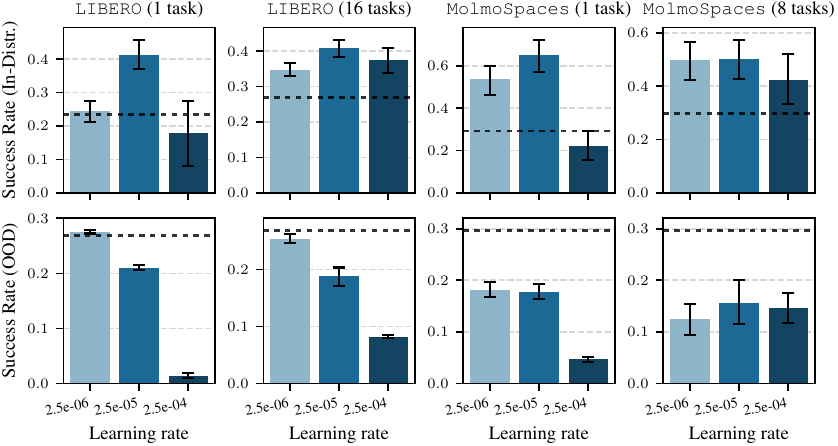}
    \vspace{-7mm}
    \caption{\textbf{Learning rate ablation for FSFT.} We sweep three learning rates across four
    task splits: \texttt{LIBERO} (1 task), \texttt{LIBERO} (16 tasks),
    \texttt{MolmoSpaces} (1 task), and \texttt{MolmoSpaces} (8 tasks). We evaluate
    performance on the training tasks (In-Distr.) and on the validation tasks (OOD).
    Bars show the mean over three seeds, with error bars denoting 95\% stratified bootstrap confidence. 
    }
    \label{fig:lr_ablation}
\end{figure}

FSFT has few hyperparameters, of which the learning rate is among the most
impactful. To assess its importance, we measure its influence on in-distribution
and out-of-distribution (OOD) performance across four single- and multi-task
splits: \texttt{LIBERO} (1 task), \texttt{LIBERO} (16 tasks),
\texttt{MolmoSpaces} (1 task), and \texttt{MolmoSpaces} (8 tasks). We sweep
three learning rates, $\{2.5 \cdot 10^{-6},\, 2.5 \cdot 10^{-5},\,
2.5 \cdot 10^{-4}\}$, over three seeds each, leaving all other hyperparameters
unchanged. Figure~\ref{fig:lr_ablation} presents the results. The smallest
value, $2.5 \cdot 10^{-6}$, degrades in-distribution performance, while the
largest, $2.5 \cdot 10^{-4}$, is detrimental OOD. These results suggest that the learning rate mediates a trade-off between
in-distribution and OOD performance. The middle value, $2.5 \cdot 10^{-5}$,
balances the two, and we adopt it as the default in our FSFT experiments.

\subsection{Number of action candidates $N$ in Best-of-$N$}
\label{app:bon_n_ablation}

\begin{figure}[h]
    \centering
    \includegraphics[width=\linewidth]{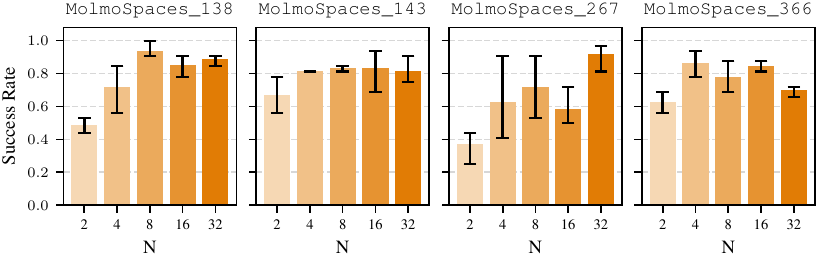}
    \caption{\textbf{Impact of the number of candidates $N$ in Best-of-$N$ on performance.} Success rate of BoN
on each of the four single-task \texttt{MolmoSpaces} runs as $N$ grows. On average, performance grows with $N$ and peaks at $N=32$, though the gains over $N \in \{8, 16\}$ are marginal. We report mean and 95\% stratified bootstrap CIs over three seeds.}
    \label{fig:bon_n_ablation}
\end{figure}

\begin{wraptable}[15]{r}{0.42\linewidth}
  \vspace{-4mm}
  \centering
\begingroup
  \definecolor{mtorange}{HTML}{E17C05}
  \begin{tikzpicture}[
    x=1pt, y=1pt,
    font=\rmfamily,
    cell/.style={inner sep=0pt, anchor=base},
    head/.style={cell, text=mtorange},
    rule/.style={draw=mtorange, line width=0.9pt},
    midrule/.style={draw=mtorange!60, line width=0.5pt}
  ]
  \def\cN{14} \def\cT{80} \def\cR{140} \def\W{165}

  \fill[mtorange!5] (0,-22) rectangle (\W,0);
  \draw[rule]    (0,0)   -- (\W,0);
  \draw[midrule] (0,-22) -- (\W,-22);

  \node[head] at (\cN,-14.5) {$N$};
  \node[head] at (\cT,-14.5) {Time (ms)};
  \node[head] at (\cR,-14.5) {Rel.};

  \foreach \n/\t/\ci/\r [count=\i] in {
      1/146.3/2.1/1.00,
      2/159.5/2.0/1.09,
      4/183.9/1.5/1.26,
      8/232.9/2.6/1.59,
      16/325.2/2.2/2.22,
      32/510.7/2.8/3.49} {
    \pgfmathsetmacro{\y}{-22-12.5-(\i-1)*14}
    \node[cell] at (\cN,\y) {\n};
    \node[cell] at (\cT,\y) {\t\,$\pm$\,\ci};
    \node[cell] at (\cR,\y) {\r$\times$};
  }

  \draw[rule] (0,-111) -- (\W,-111);
  \end{tikzpicture}
\endgroup
  \vspace{-2mm}
  \caption{\textbf{BoN inference cost.} Wall-clock time of one Best-of-$N$ step
  on a single GH200 GPU, relative to $N{=}1$.}
  \label{tab:bofn_compute_cost}
\end{wraptable}

Beyond critic learning, Best-of-$N$ introduces a single hyperparameter: the
number of sampled actions $N$. At each step it draws $N$ candidates
$\{a_t^i\}_{i=1}^N \sim \pi_\theta(s_t, c)$, scores them, and executes
$a_t^\star = \arg\max_i Q_\psi(s_t, a_t^i, c)$. Larger $N$ gives the critic more
candidates to choose from, at the cost of slower inference. To assess the impact of $N$ on performance, we ablate $N \in \{2, 4, 8, 16, 32\}$ on four single-task \texttt{MolmoSpaces}
runs. As shown in Figure~\ref{fig:bon_n_ablation}, performance roughly grows with $N$ on
all tasks and peaks on average at $N = 32$, which we adopt as our default. The
gains are diminishing, however: $N = 32$ improves only marginally over
$N \in \{8, 16\}$.

\Cref{tab:bofn_compute_cost} reports the wall-clock time of one BoN step as a function of $N$. Our default $N=32$ costs $3.49\times$ the time of $N=1$, against $1.59\times$ for $N=8$, which offers a better balance between performance and inference cost. Since our experiments run in simulation, we
keep $N=32$ to prioritize performance; on a real robot, $N=8$ would likely be preferable.

\newpage
\section{Experimental Details}
\label{app:experimental_details}
\vspace{-3pt}

\subsection{Environments and Task Splits}
\label{app:task_splits}

This section describes the environments we use and how we construct task 
splits. In all of our experiments, we utilize two benchmarks: \texttt{LIBERO-90} and \texttt{MolmoSpaces}. 
\texttt{LIBERO} is a widely used manipulation benchmark, whereas
\texttt{MolmoSpaces} offers precise control over task diversity, which makes it
better suited to studying generalization. We use its \texttt{pick} suite, which
exposes 18 objects with 55--56 scenes each, allowing us to vary scene and object
diversity independently.

\begin{table}[h]
  \centering
  \vspace{-10pt}
\caption{\textbf{\texttt{MolmoSpaces} task splits.} For each training split, we
  list the task IDs it contains, grouped by object and given without the
  \texttt{molmo\_} prefix. \emph{Scenes/obj.} is the number of scenes per
  object; \emph{Init.\ SR} the mean success rate of the base policy. Each
  training split is evaluated only on unseen scenes of the objects it was trained
  on (\emph{Eval.\ tasks}: 8 held-out scenes per training object), and every split
  is evaluated on the same 8 scenes of each of 6 objects never seen during
  training. $^\dagger$Four independent single-task runs (one per listed task),
  each evaluated on the 8 unseen scenes of its own object. $^\ddagger$Pool of
  unseen-scene tasks; each split uses the rows of its own training objects.}
  \label{tab:molmo_splits}
  \vspace{2mm}
\begingroup
  \def\molmoscale{0.89}
  \definecolor{mtorange}{HTML}{E17C05}
  \scalebox{\molmoscale}{%
  \begin{tikzpicture}[
    x=1pt, y=1pt,
    font=\rmfamily,
    cell/.style={inner sep=0pt, anchor=base},
    lbl/.style={cell, anchor=base west},
    sub/.style={cell, anchor=base west, font=\rmfamily\footnotesize},
    head/.style={cell, text=mtorange, font=\rmfamily\small},
    rule/.style={draw=mtorange, line width=0.9pt},
    midrule/.style={draw=mtorange!60, line width=0.5pt},
    sep/.style={draw=mtorange!25, line width=0.4pt}
  ]
  \def\W{446}

  \fill[mtorange!5] (0,-26) rectangle (\W,0);
  \draw[rule]    (0,0)   -- (\W,0);
  \draw[midrule] (0,-26) -- (\W,-26);

  \node[head] at (315.9,-9) {Training};
  \node[head] at (409.5,-9) {Unseen-scene eval.};
  \draw[draw=mtorange!60, line width=0.4pt] (264,-12) -- (367.7,-12);
  \draw[draw=mtorange!60, line width=0.4pt] (375,-12) -- (444,-12);
  \node[head, anchor=base west] at (3,-21) {Split};
  \node[head] at (274.6,-21) {Tasks};
  \node[head] at (313,-21) {Scenes/obj.};
  \node[head] at (354.3,-21) {SR (\%)};
  \node[head] at (385.5,-21) {Tasks};
  \node[head] at (430.6,-21) {SR (\%)};

  \node[lbl]  at (3,-34.5) {\texttt{MolmoSpaces} (1 task)$^\dagger$};
  \node[cell] at (274.6,-34.5) {1\,$\times$\,4};
  \node[cell] at (313,-34.5) {1};
  \node[cell] at (354.3,-34.5) {30.2};
  \node[cell] at (385.5,-34.5) {8\,$\times$\,4};
  \node[cell] at (430.6,-34.5) {30.0};
  \node[sub] at (14,-46.5) {\emph{box}: 366, 138; \emph{spatula}: 373, 93};
  \node[lbl]  at (3,-61.0) {\texttt{MolmoSpaces} (4 tasks)};
  \node[cell] at (274.6,-61.0) {4};
  \node[cell] at (313,-61.0) {2};
  \node[cell] at (354.3,-61.0) {30.2};
  \node[cell] at (385.5,-61.0) {16};
  \node[cell] at (430.6,-61.0) {30.0};
  \node[sub] at (14,-73.0) {\emph{box}: 366, 138; \emph{spatula}: 373, 93};
  \node[lbl]  at (3,-87.5) {\texttt{MolmoSpaces} (16 tasks)};
  \node[cell] at (274.6,-87.5) {16};
  \node[cell] at (313,-87.5) {4};
  \node[cell] at (354.3,-87.5) {29.2};
  \node[cell] at (385.5,-87.5) {32};
  \node[cell] at (430.6,-87.5) {29.9};
  \node[sub] at (14,-99.5) {\emph{box}: 267, 366, 138, 143; \emph{spatula}: 176, 373, 93, 394};
  \node[sub] at (14,-110.0) {\emph{cup}: 48, 185, 24, 62; \emph{tissuepaper}: 4, 260, 221, 348};
  \node[lbl]  at (3,-124.5) {\texttt{MolmoSpaces} (64 tasks)};
  \node[cell] at (274.6,-124.5) {64};
  \node[cell] at (313,-124.5) {8};
  \node[cell] at (354.3,-124.5) {29.5};
  \node[cell] at (385.5,-124.5) {64};
  \node[cell] at (430.6,-124.5) {29.9};
  \node[sub] at (14,-136.5) {\emph{boiler}: 332, 33, 242, 369, 126, 383, 295, 81};
  \node[sub] at (14,-147.0) {\emph{box}: 164, 86, 70, 267, 366, 138, 143, 212};
  \node[sub] at (14,-157.5) {\emph{cup}: 18, 292, 48, 185, 24, 62, 377, 78};
  \node[sub] at (14,-168.0) {\emph{ladle}: 28, 311, 50, 56, 399, 187, 121, 27};
  \node[sub] at (14,-178.5) {\emph{pot}: 309, 227, 265, 325, 277, 127, 254, 166};
  \node[sub] at (14,-189.0) {\emph{tissuepaper}: 11, 361, 103, 4, 260, 221, 348, 305};
  \node[sub] at (14,-199.5) {\emph{spatula}: 22, 210, 176, 373, 93, 394, 6, 47};
  \node[sub] at (14,-210.0) {\emph{spoon}: 19, 296, 82, 39, 162, 148, 150, 7};
  \node[lbl]  at (3,-224.5) {\texttt{MolmoSpaces} (144 tasks)};
  \node[cell] at (274.6,-224.5) {144};
  \node[cell] at (313,-224.5) {12};
  \node[cell] at (354.3,-224.5) {29.0};
  \node[cell] at (385.5,-224.5) {96};
  \node[cell] at (430.6,-224.5) {29.7};
  \node[sub] at (14,-236.5) {\emph{boiler}: 332, 33, 242, 369, 126, 383, 295, 81, 407, 410, 642, 786};
  \node[sub] at (14,-247.0) {\emph{box}: 164, 86, 70, 267, 366, 138, 143, 212, 523, 597, 669, 729};
  \node[sub] at (14,-257.5) {\emph{cup}: 18, 292, 48, 185, 24, 62, 377, 78, 196, 237, 32, 381};
  \node[sub] at (14,-268.0) {\emph{ladle}: 28, 311, 50, 56, 399, 187, 121, 27, 100, 15, 198, 71};
  \node[sub] at (14,-278.5) {\emph{pot}: 309, 227, 265, 325, 277, 127, 254, 166, 204, 233, 328, 68};
  \node[sub] at (14,-289.0) {\emph{tissuepaper}: 11, 361, 103, 4, 260, 221, 348, 305, 104, 334, 376, 382};
  \node[sub] at (14,-299.5) {\emph{spatula}: 22, 210, 176, 373, 93, 394, 6, 47, 415, 595, 618, 985};
  \node[sub] at (14,-310.0) {\emph{spoon}: 19, 296, 82, 39, 162, 148, 150, 7, 430, 695, 727, 909};
  \node[sub] at (14,-320.5) {\emph{salt/pepper shaker}: 124, 141, 149, 211, 251, 275, 312, 34, 358, 368, 387, 88};
  \node[sub] at (14,-331.0) {\emph{soapdispenser}: 12, 208, 224, 229, 236, 25, 324, 327, 355, 389, 80, 91};
  \node[sub] at (14,-341.5) {\emph{mug}: 160, 991, 900, 973, 915, 953, 504, 896, 957, 362, 650, 980};
  \node[sub] at (14,-352.0) {\emph{fruit/veg}: 241, 315, 450, 544, 603, 643, 721, 746, 777, 831, 85, 854};
  \node[lbl]  at (3,-366.5) {Unseen scenes (eval.)$^\ddagger$};
  \node[cell] at (313,-366.5) {8};
  \node[cell] at (385.5,-366.5) {96};
  \node[cell] at (430.6,-366.5) {29.7};
  \node[sub] at (14,-378.5) {\emph{boiler}: 754, 809, 883, 8, 961, 611, 498, 673};
  \node[sub] at (14,-389.0) {\emph{box}: 55, 885, 905, 273, 999, 965, 480, 995};
  \node[sub] at (14,-399.5) {\emph{cup}: 219, 398, 749, 222, 424, 799, 429, 134};
  \node[sub] at (14,-410.0) {\emph{ladle}: 925, 952, 807, 73, 692, 651, 438, 645};
  \node[sub] at (14,-420.5) {\emph{pot}: 425, 271, 346, 972, 735, 634, 797, 590};
  \node[sub] at (14,-431.0) {\emph{tissuepaper}: 766, 771, 958, 49, 637, 940, 775, 560};
  \node[sub] at (14,-441.5) {\emph{spatula}: 461, 711, 739, 748, 21, 773, 968, 510};
  \node[sub] at (14,-452.0) {\emph{spoon}: 725, 898, 471, 762, 513, 379, 570, 285};
  \node[sub] at (14,-462.5) {\emph{salt/pepper shaker}: 559, 741, 678, 163, 908, 481, 703, 653};
  \node[sub] at (14,-473.0) {\emph{soapdispenser}: 850, 531, 841, 781, 154, 607, 417, 705};
  \node[sub] at (14,-483.5) {\emph{mug}: 459, 684, 828, 444, 759, 123, 728, 997};
  \node[sub] at (14,-494.0) {\emph{fruit/veg}: 418, 588, 43, 879, 962, 477, 770, 842};
  \node[lbl]  at (3,-508.5) {Unseen objects (eval., all splits)};
  \node[cell] at (313,-508.5) {8};
  \node[cell] at (385.5,-508.5) {48};
  \node[cell] at (430.6,-508.5) {28.9};
  \node[sub] at (14,-520.5) {\emph{atomizer}: 209, 397, 192, 99, 140, 57, 344, 400};
  \node[sub] at (14,-531.0) {\emph{bowl}: 665, 118, 214, 356, 101, 136, 230, 718};
  \node[sub] at (14,-541.5) {\emph{fork}: 281, 181, 372, 345, 167, 290, 518, 31};
  \node[sub] at (14,-552.0) {\emph{knife}: 419, 433, 629, 934, 155, 223, 512, 878};
  \node[sub] at (14,-562.5) {\emph{remotecontrol}: 465, 152, 349, 353, 390, 357, 359, 135};
  \node[sub] at (14,-573.0) {\emph{winebottle}: 679, 206, 371, 448, 226, 340, 248, 798};

  \draw[sep] (0,-52.5) -- (\W,-52.5);
  \draw[sep] (0,-79.0) -- (\W,-79.0);
  \draw[sep] (0,-116.0) -- (\W,-116.0);
  \draw[sep] (0,-216.0) -- (\W,-216.0);
  \draw[sep] (0,-358.0) -- (\W,-358.0);
  \draw[sep] (0,-500.0) -- (\W,-500.0);
  \draw[rule] (0,-579.0) -- (\W,-579.0);
  \end{tikzpicture}%
  }
\endgroup
\end{table}

\begin{figure}[p]
  \centering
  \setlength{\tabcolsep}{12pt}
  \renewcommand{\arraystretch}{0.5}
  \newcommand{\molmopic}{0.1\textwidth}
  \newcommand{\molmotile}[1]{\tikz[baseline=(img.south)]{%
    \node[inner sep=0pt, anchor=south west] (img) at (0,0) {\includegraphics[width=\molmopic]{figs/render/molmo_pics/molmo_#1.jpg}};%
    \node[anchor=south west, inner sep=0.8pt, fill=white, fill opacity=0.75, text opacity=1, font=\fontsize{5}{4}\selectfont] at (img.south west) {#1};}}
  \newcommand{\molmohead}[1]{\makebox[\molmopic][c]{\footnotesize{#1}}\rule[-2pt]{0pt}{0pt}}

  \begin{tabular}{@{}*{6}{c}@{}}
    \molmohead{boiler} & \molmohead{box} & \molmohead{cup} & \molmohead{ladle} & \molmohead{pot} & \molmohead{tissuepaper} \\
    [1pt]
    \molmotile{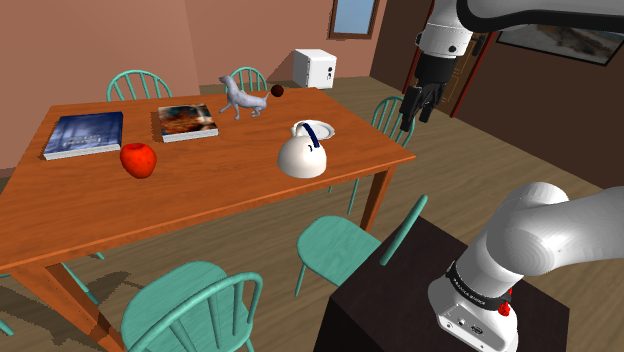} & \molmotile{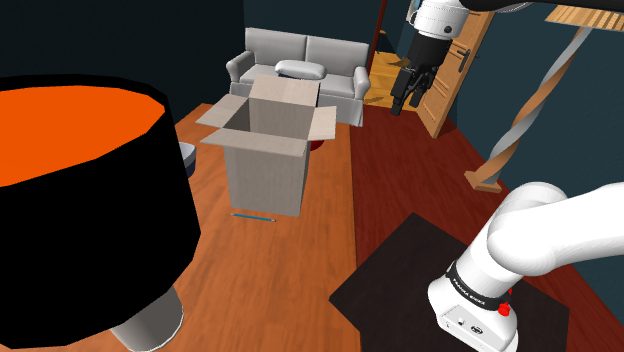} & \molmotile{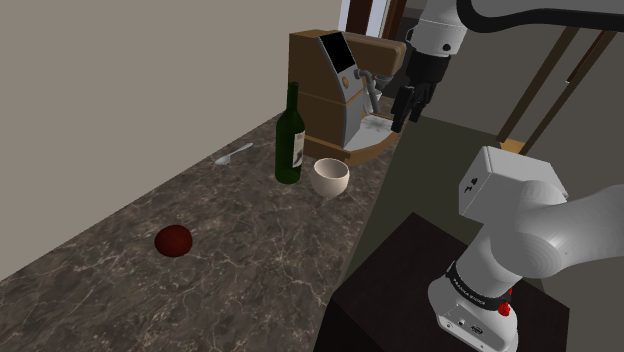} & \molmotile{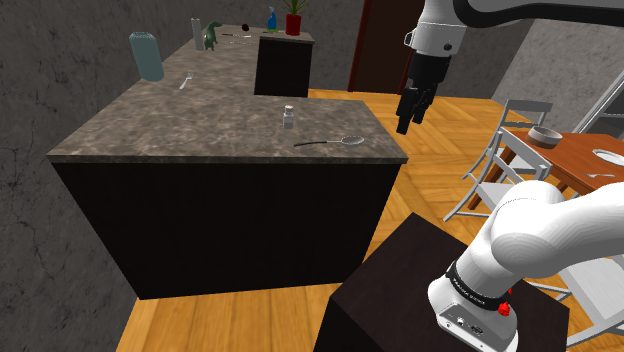} & \molmotile{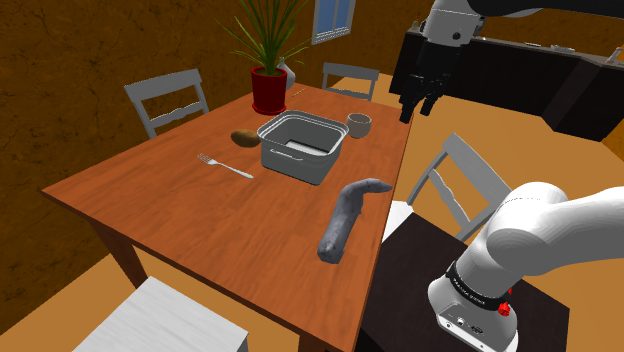} & \molmotile{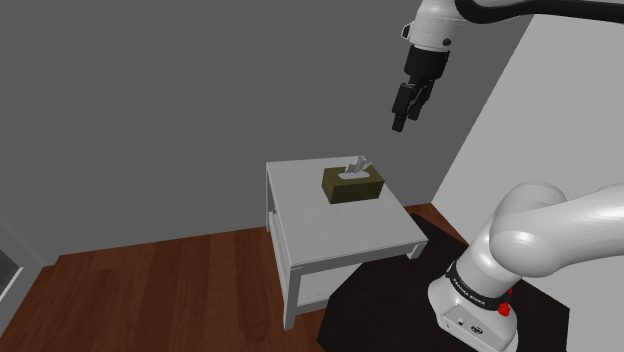} \\
    \molmotile{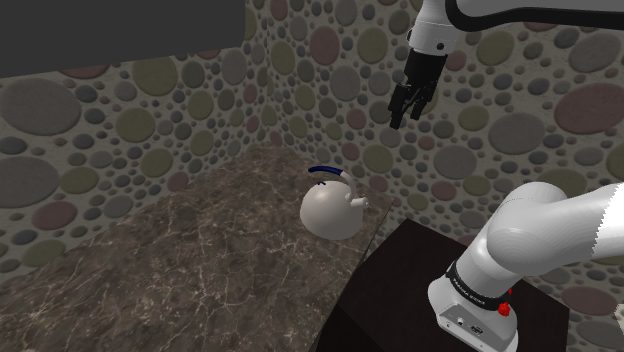} & \molmotile{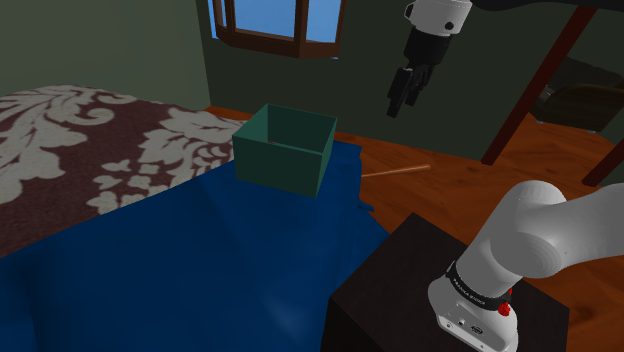} & \molmotile{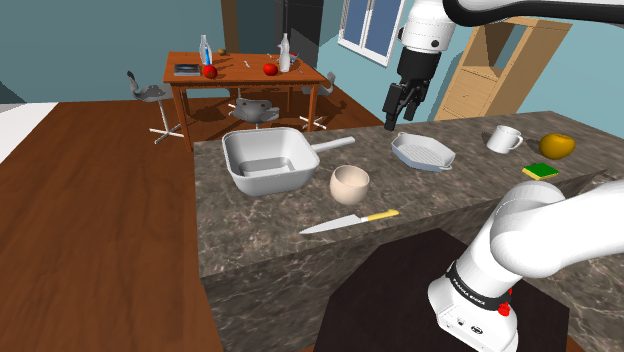} & \molmotile{311} & \molmotile{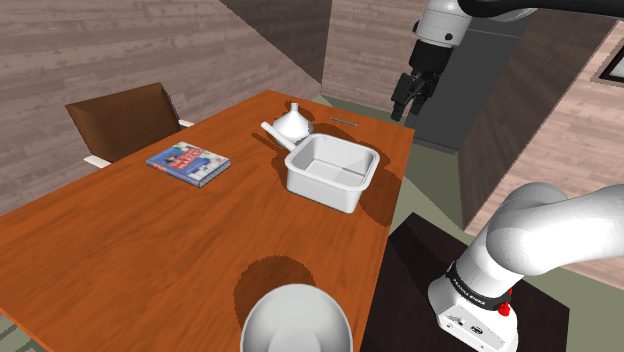} & \molmotile{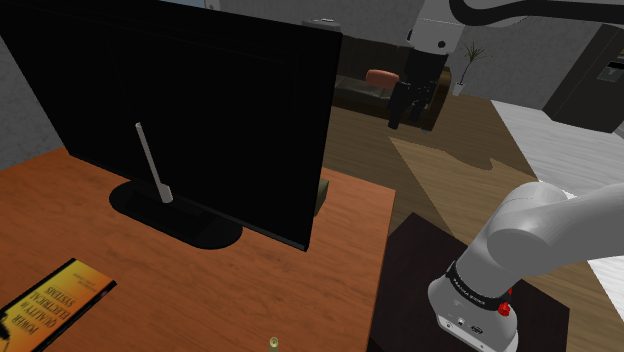} \\
    \molmotile{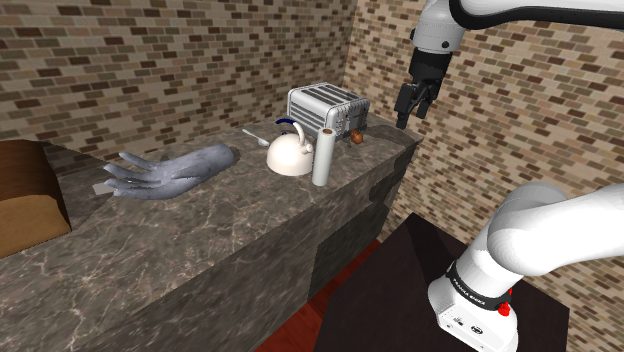} & \molmotile{70} & \molmotile{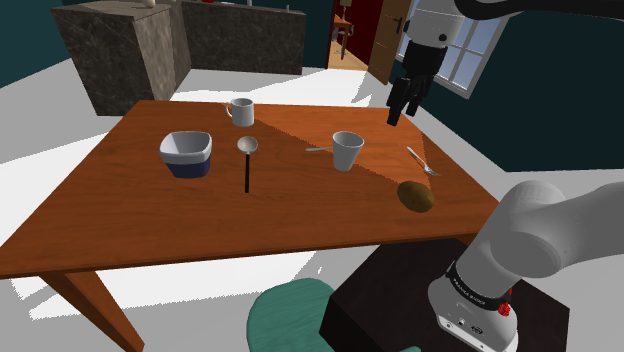} & \molmotile{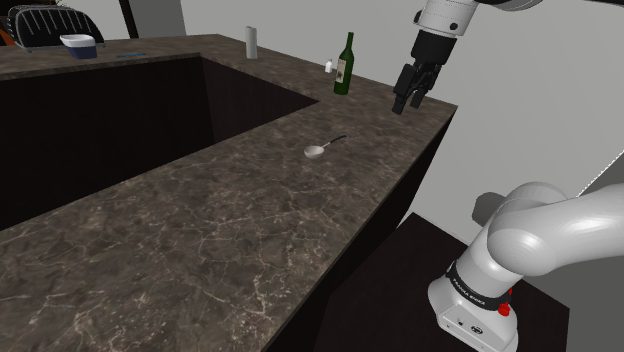} & \molmotile{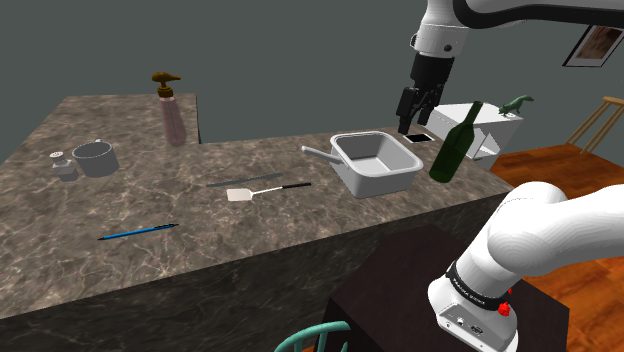} & \molmotile{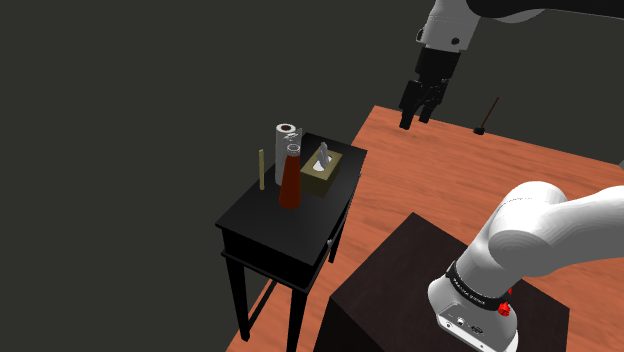} \\
    \molmotile{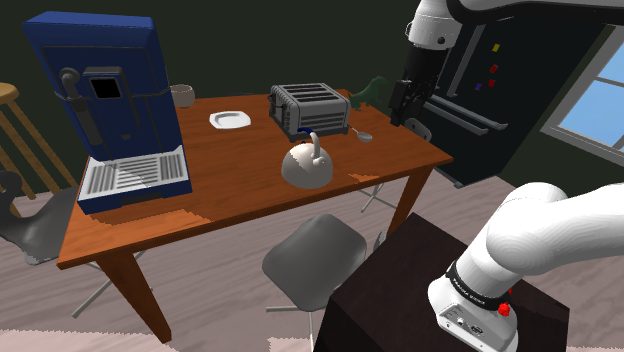} & \molmotile{267} & \molmotile{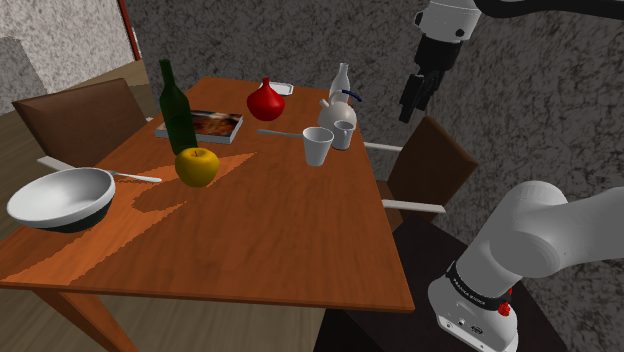} & \molmotile{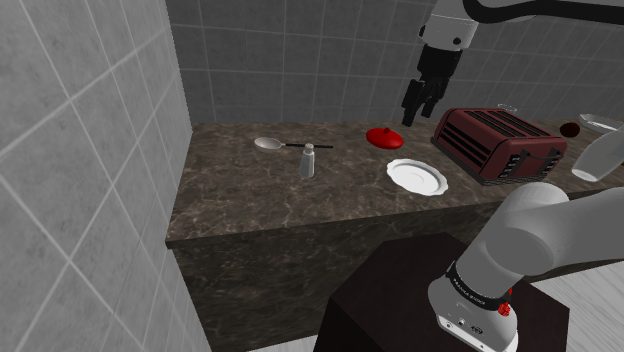} & \molmotile{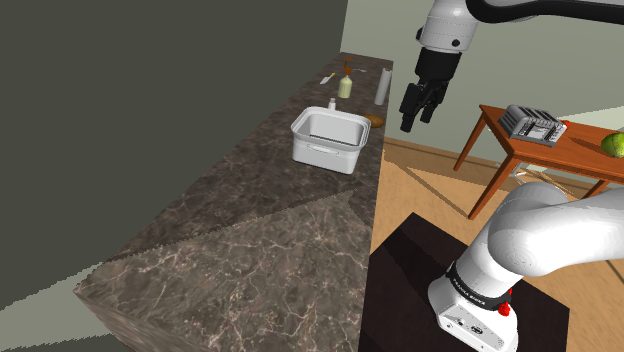} & \molmotile{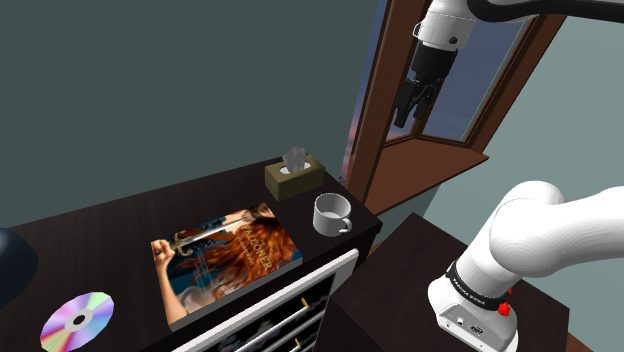} \\
    \molmotile{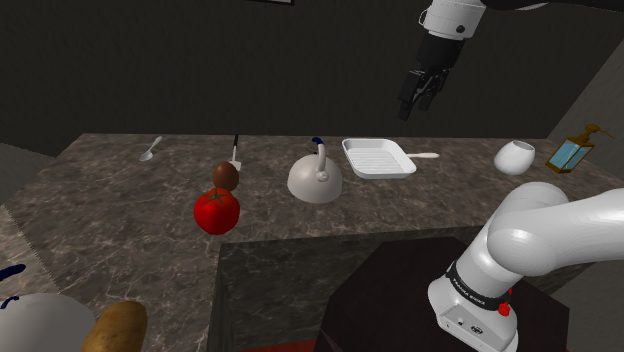} & \molmotile{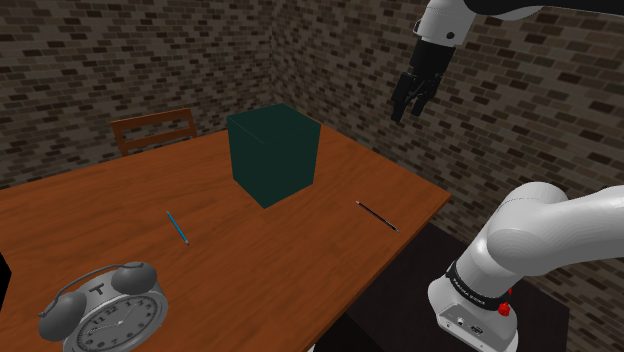} & \molmotile{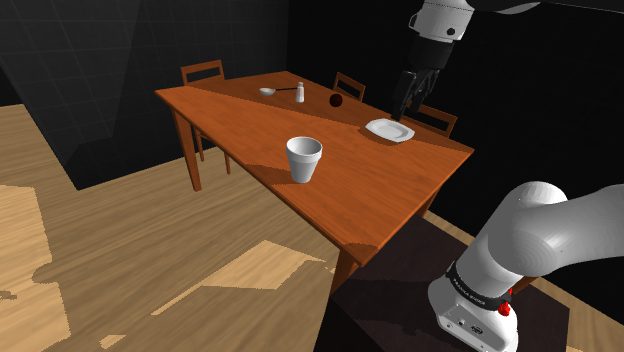} & \molmotile{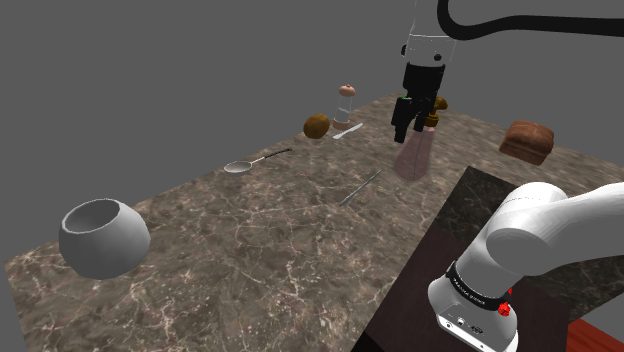} & \molmotile{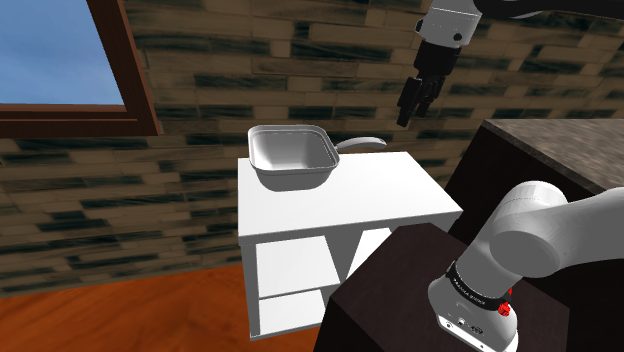} & \molmotile{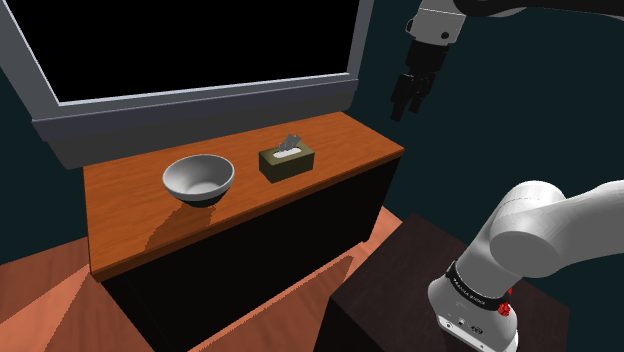} \\
    \molmotile{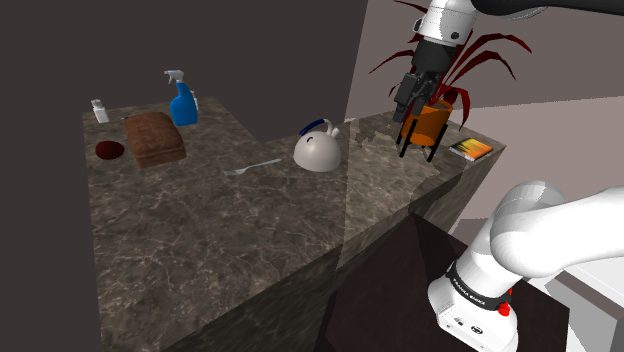} & \molmotile{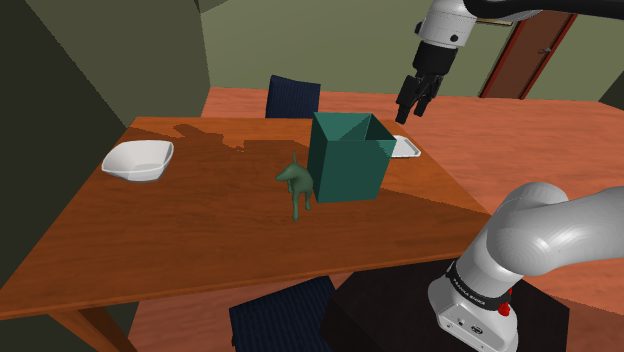} & \molmotile{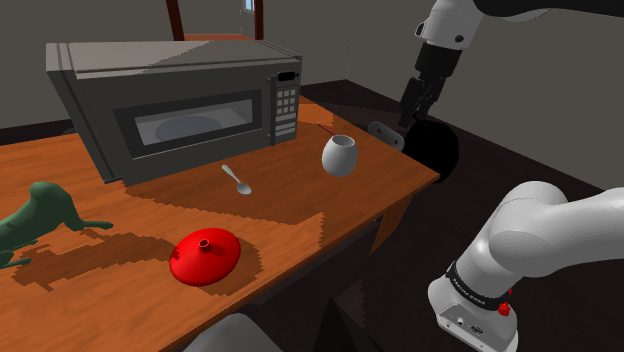} & \molmotile{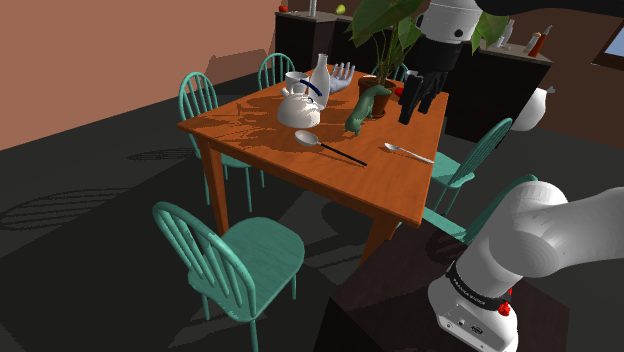} & \molmotile{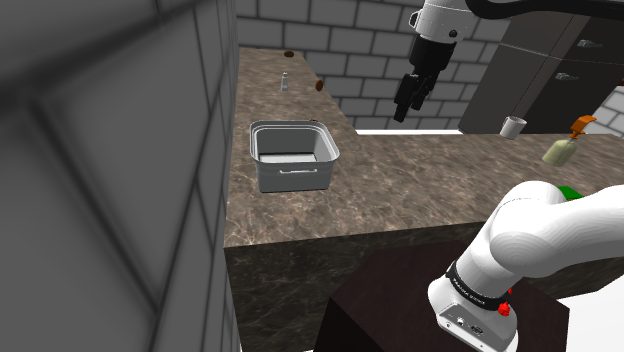} & \molmotile{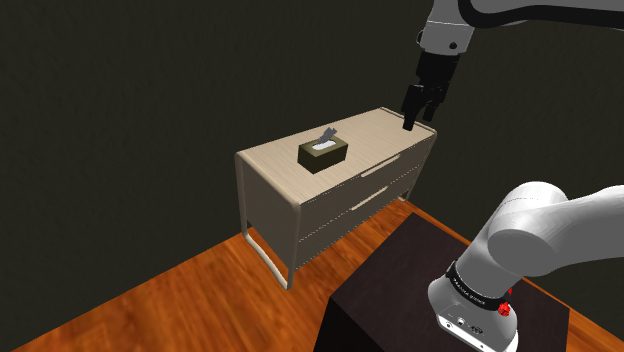} \\
    \molmotile{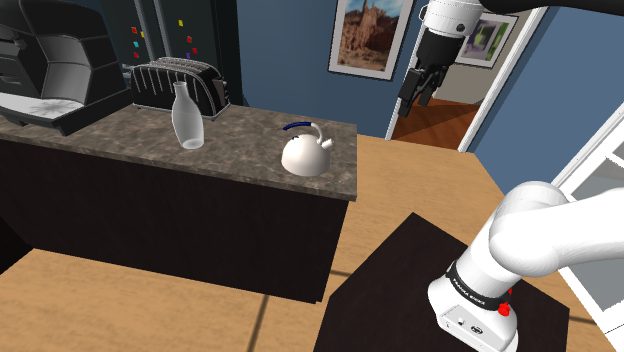} & \molmotile{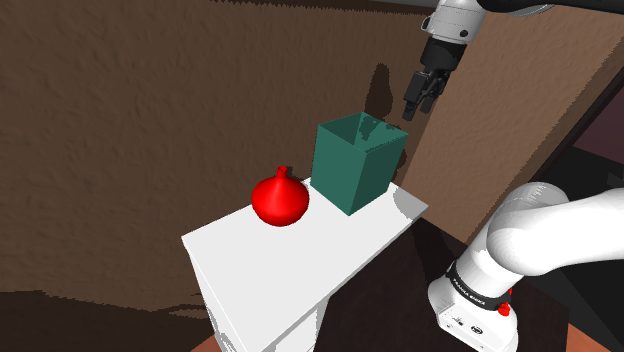} & \molmotile{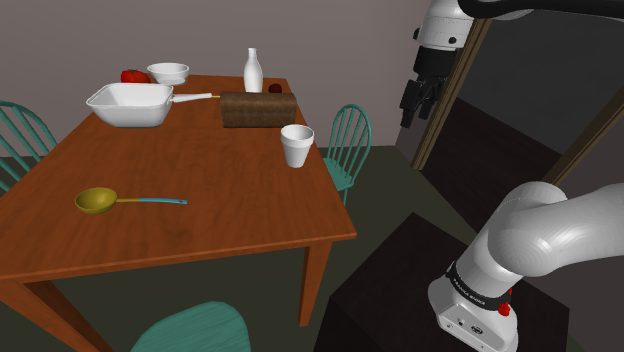} & \molmotile{121} & \molmotile{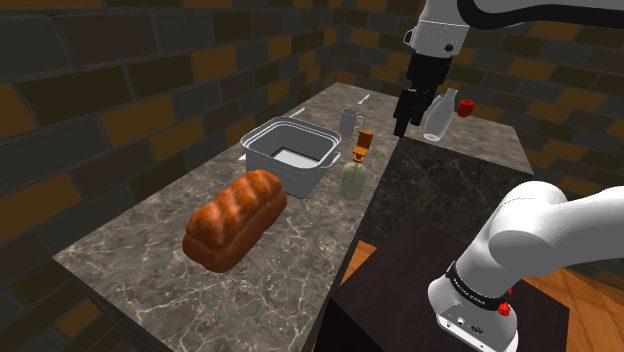} & \molmotile{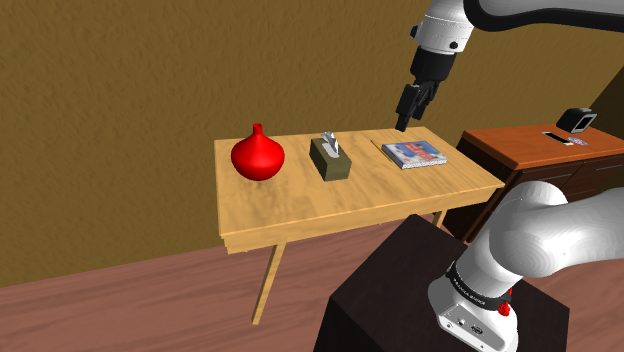} \\
    \molmotile{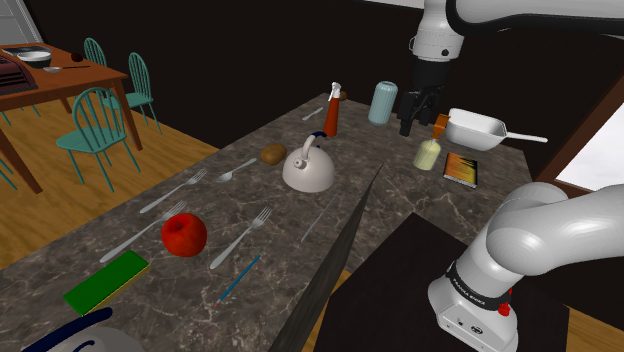} & \molmotile{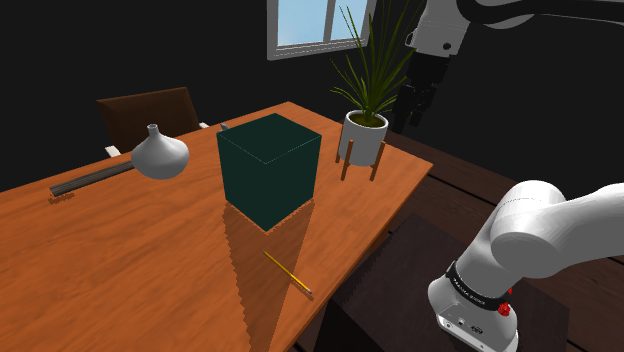} & \molmotile{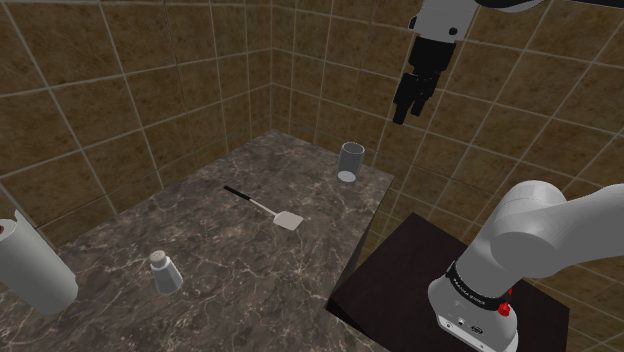} & \molmotile{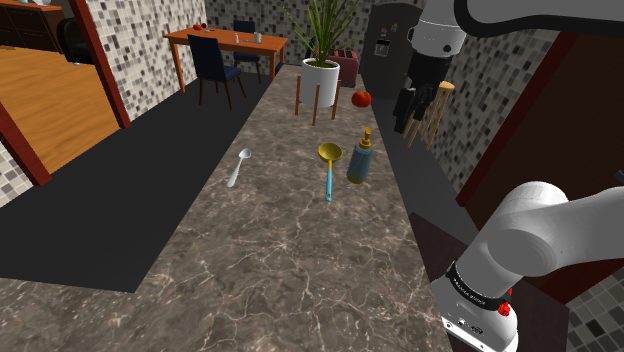} & \molmotile{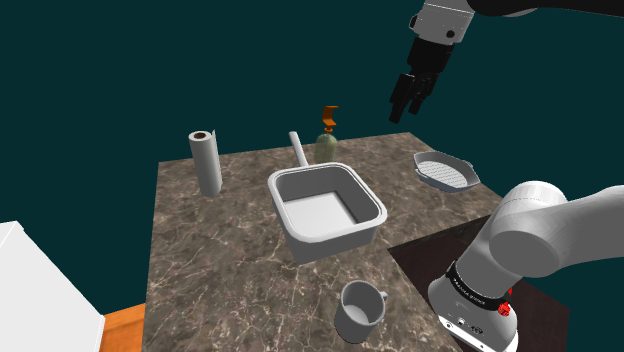} & \molmotile{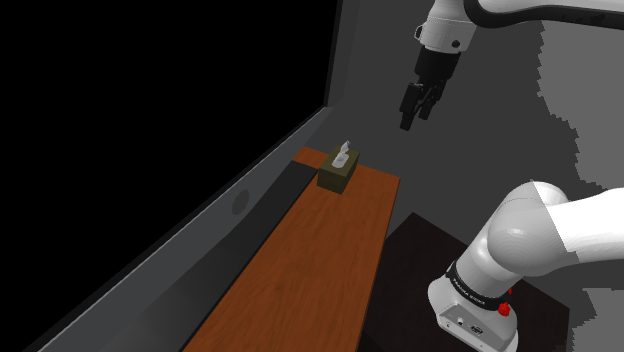} \\
    \molmotile{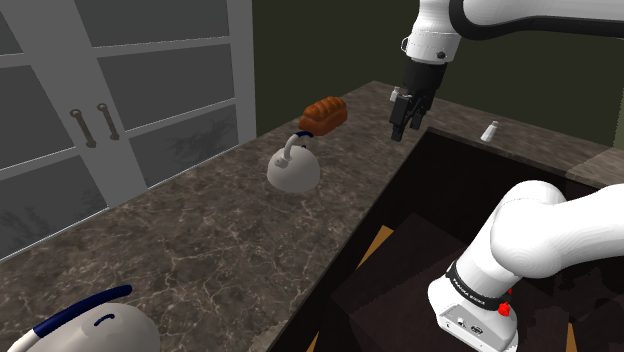} & \molmotile{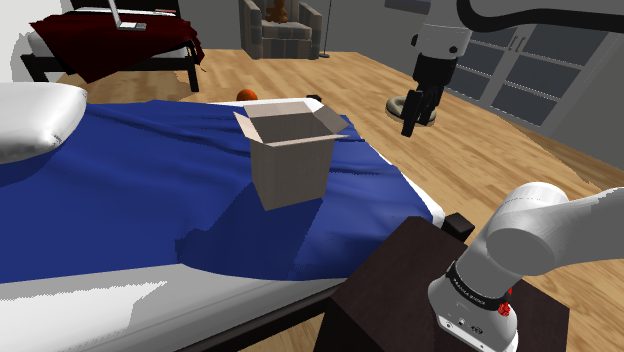} & \molmotile{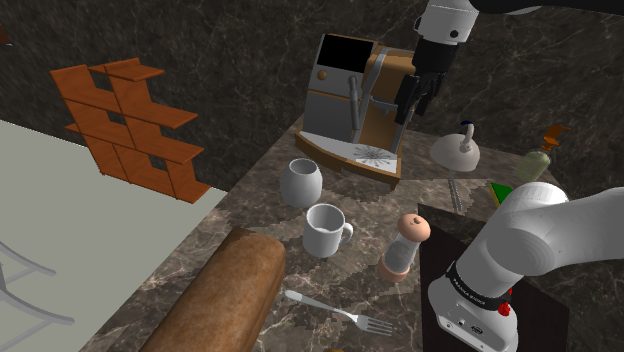} & \molmotile{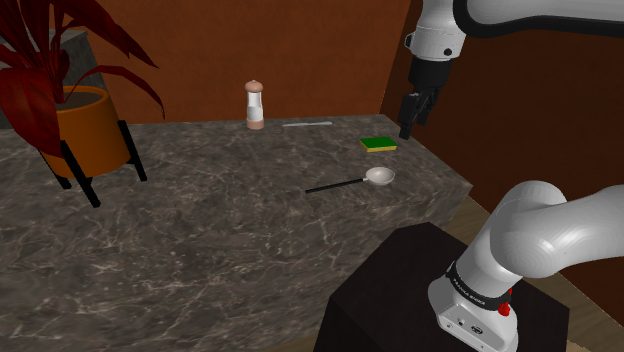} & \molmotile{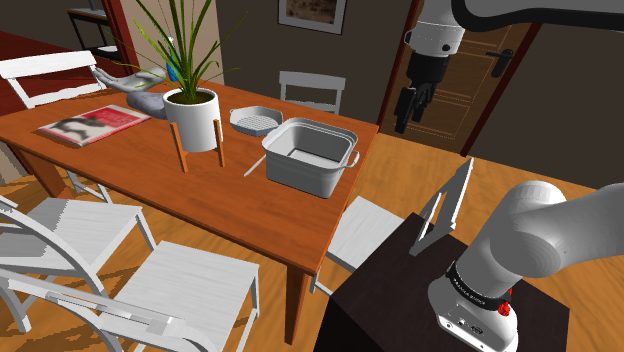} & \molmotile{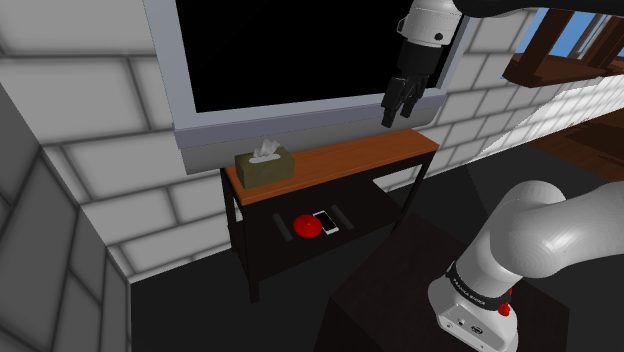} \\
    \molmotile{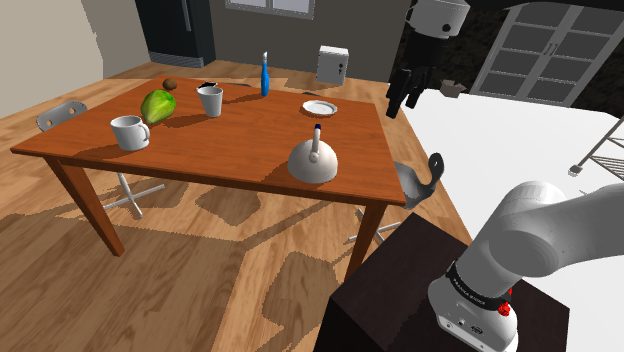} & \molmotile{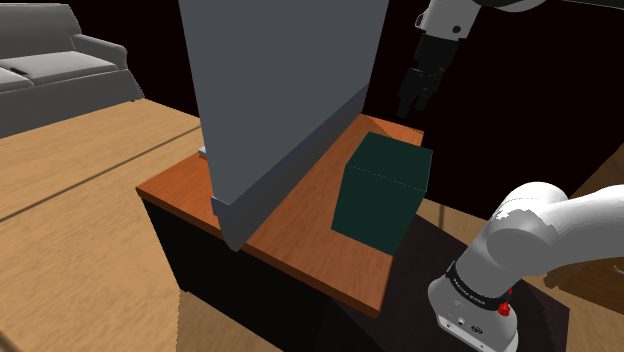} & \molmotile{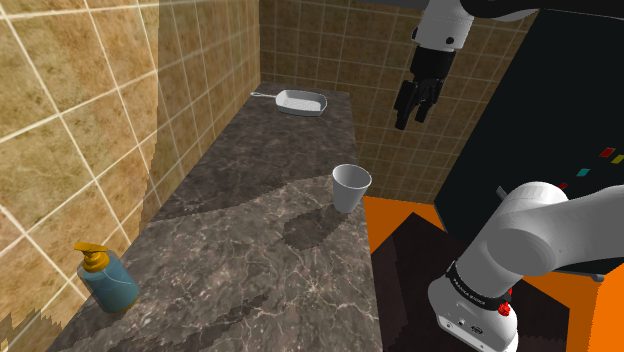} & \molmotile{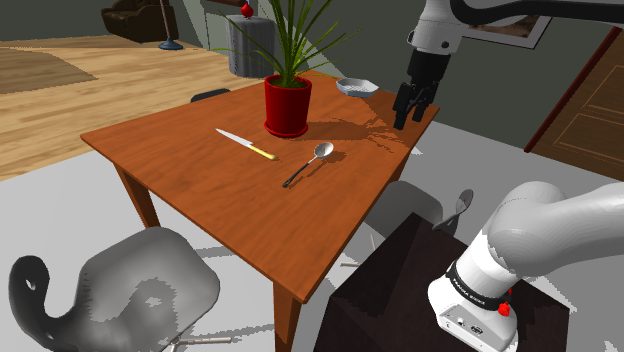} & \molmotile{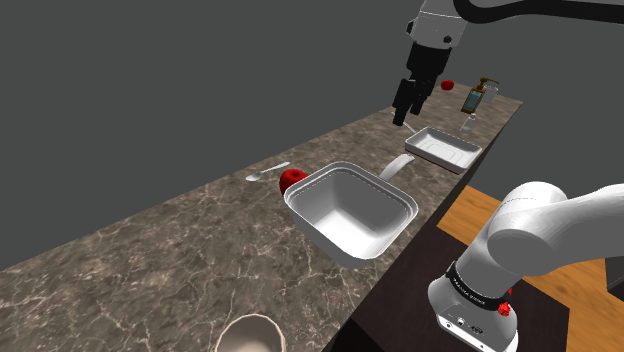} & \molmotile{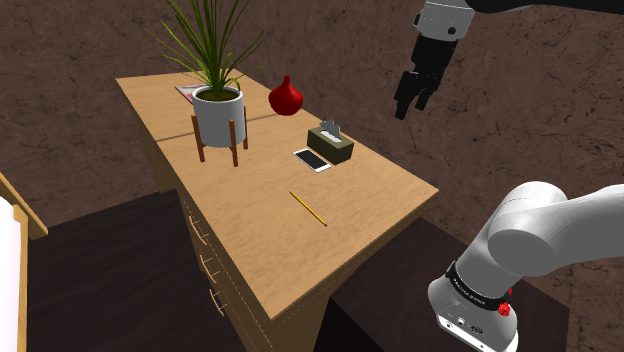} \\
    \molmotile{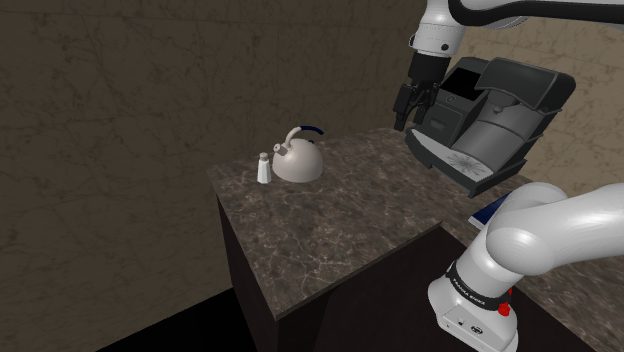} & \molmotile{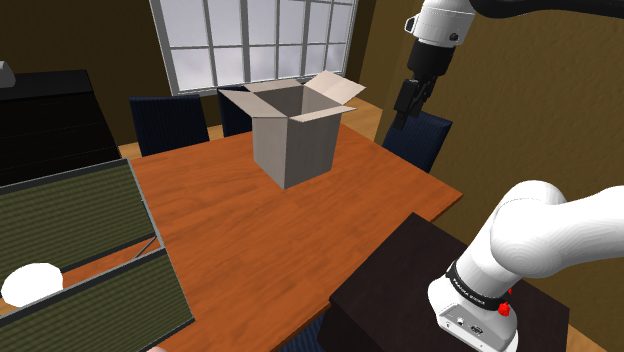} & \molmotile{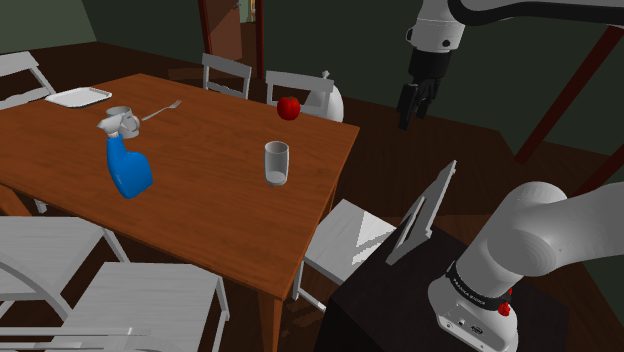} & \molmotile{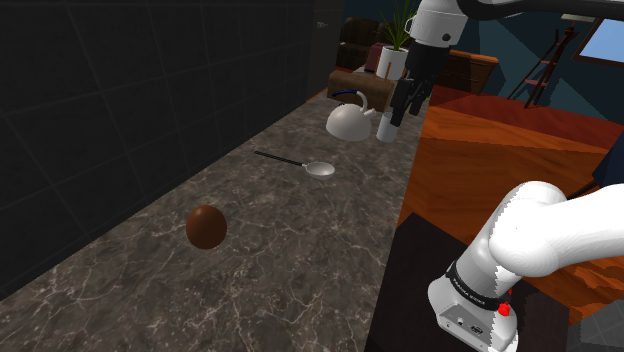} & \molmotile{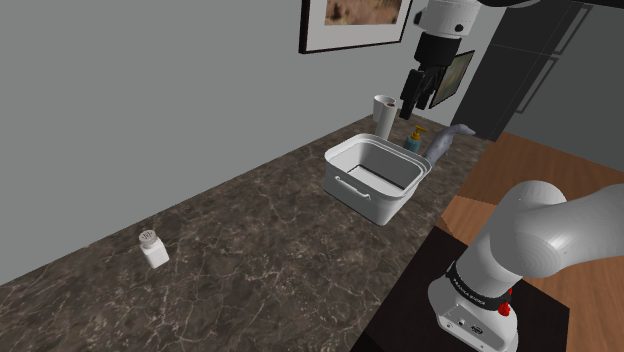} & \molmotile{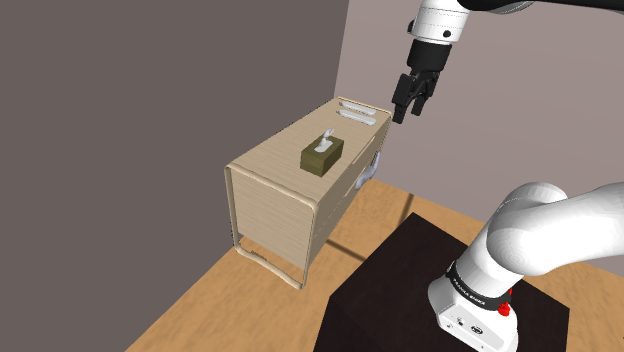} \\
    \molmotile{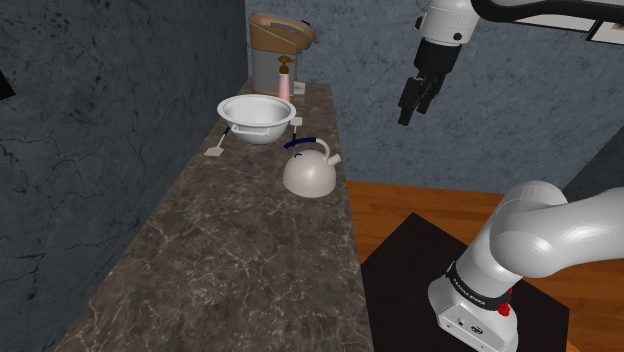} & \molmotile{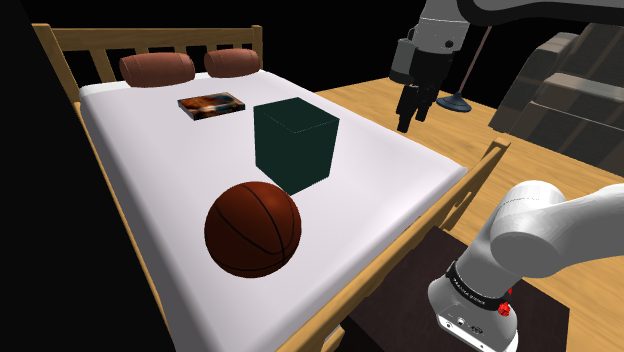} & \molmotile{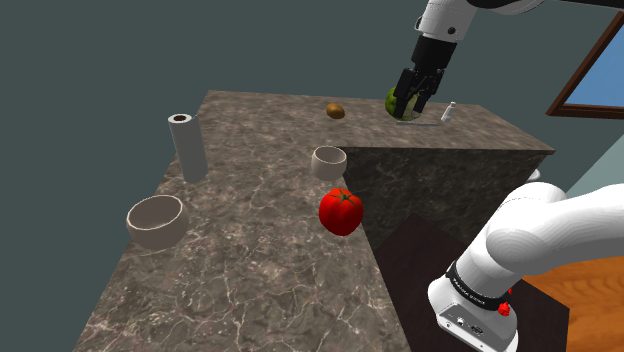} & \molmotile{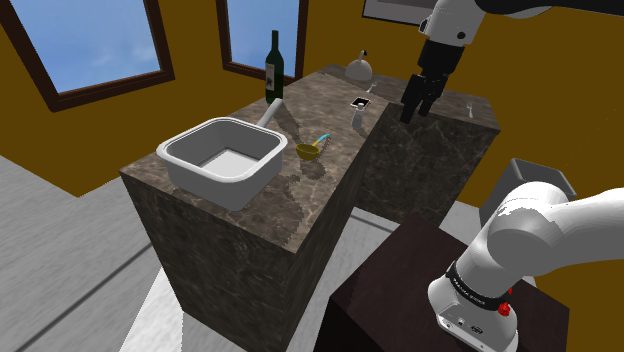} & \molmotile{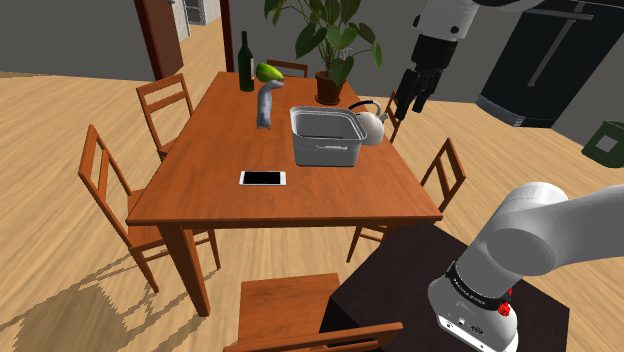} & \molmotile{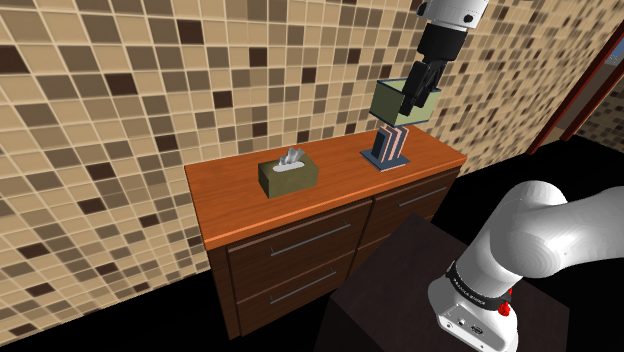} \\
  \end{tabular}
  \\
  \vspace{2mm}
  \begin{tabular}{@{}*{6}{c}@{}}
    \molmohead{spatula} & \molmohead{spoon} & \molmohead{salt/pepper shaker} & \molmohead{soapdispenser} & \molmohead{mug} & \molmohead{fruit/veg} \\
    [1pt]
    \molmotile{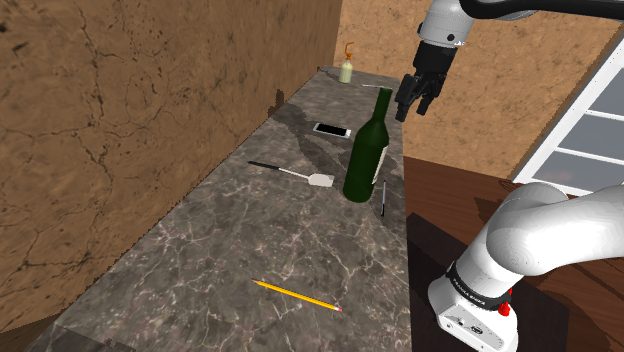} & \molmotile{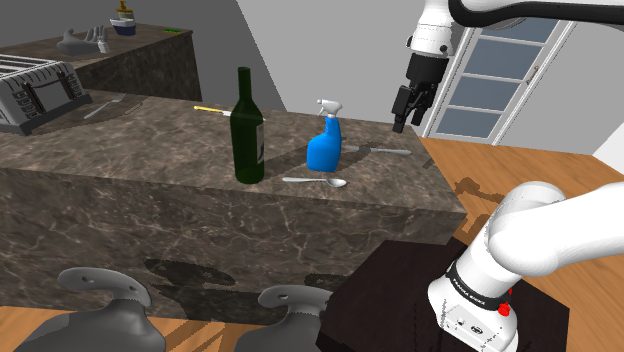} & \molmotile{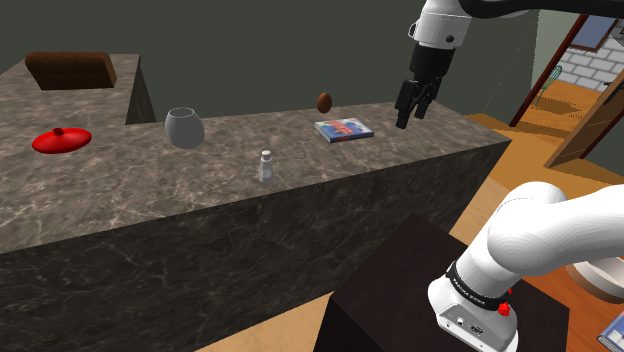} & \molmotile{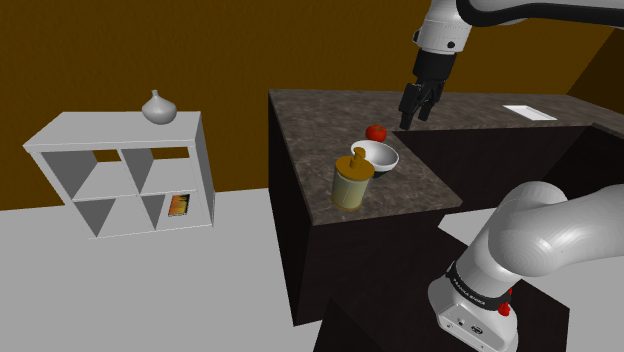} & \molmotile{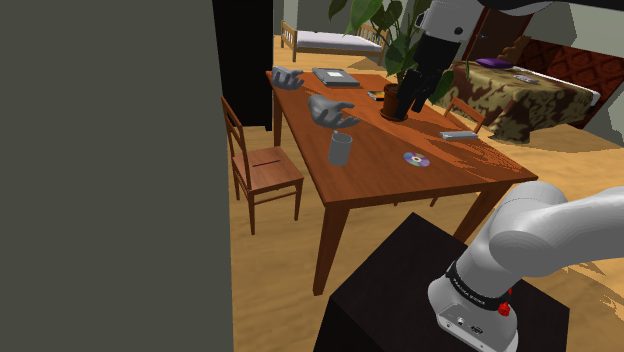} & \molmotile{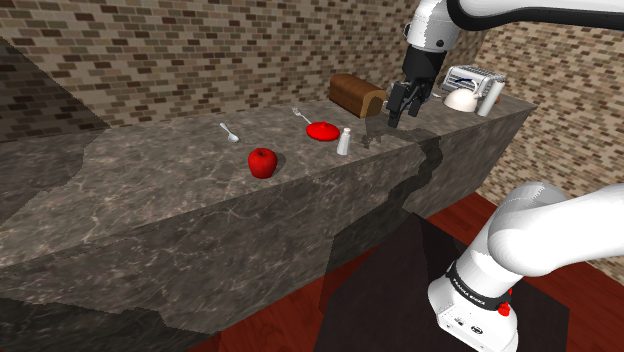} \\
    \molmotile{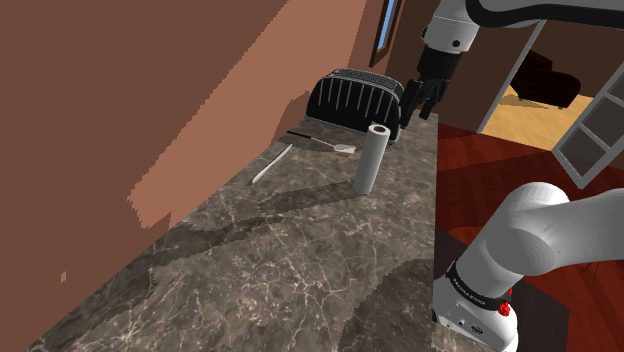} & \molmotile{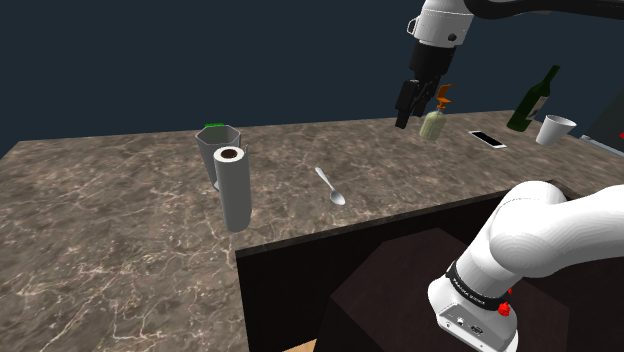} & \molmotile{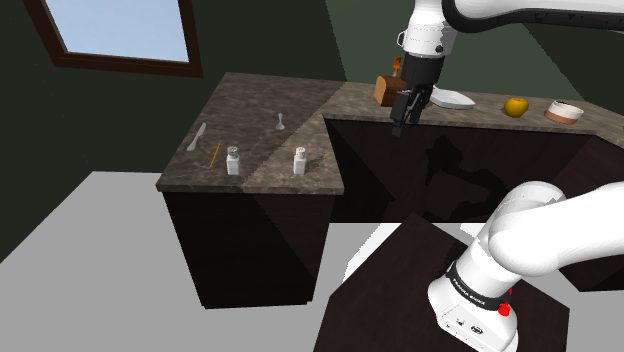} & \molmotile{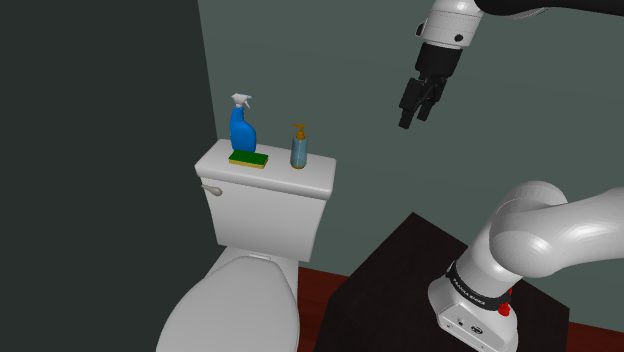} & \molmotile{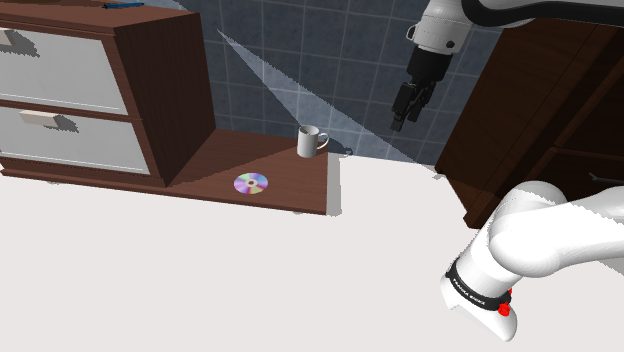} & \molmotile{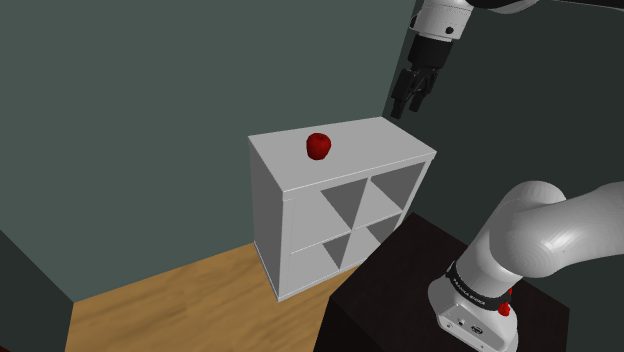} \\
    \molmotile{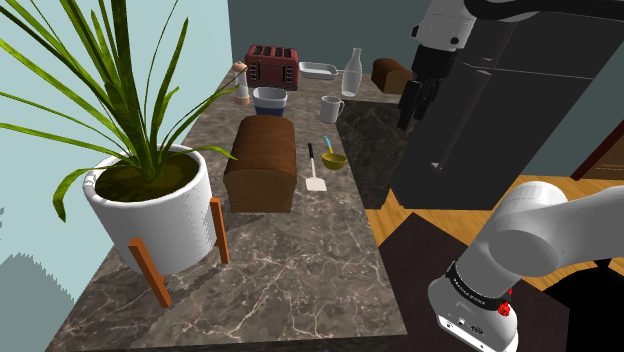} & \molmotile{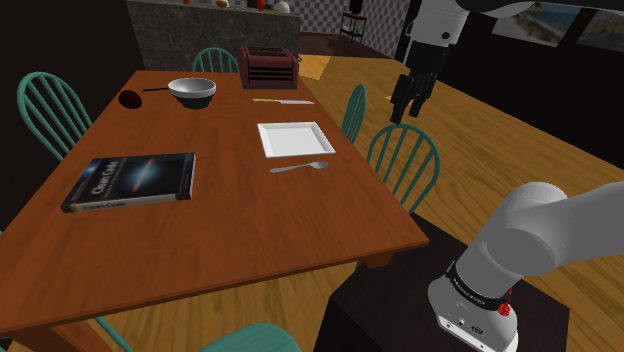} & \molmotile{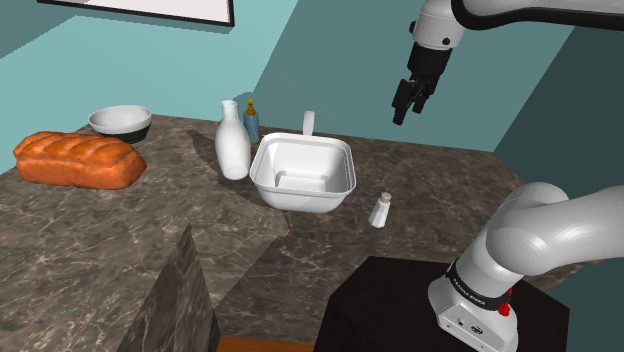} & \molmotile{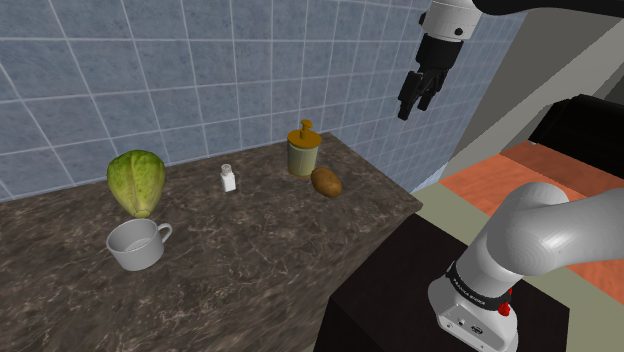} & \molmotile{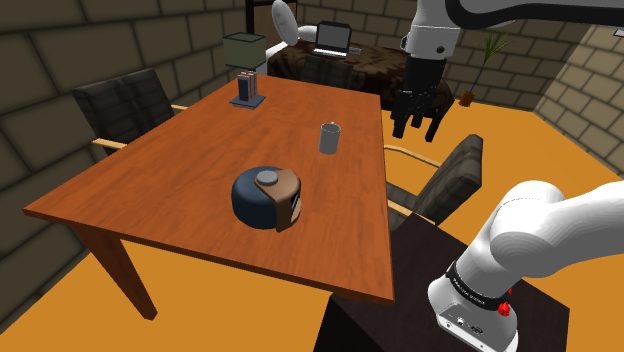} & \molmotile{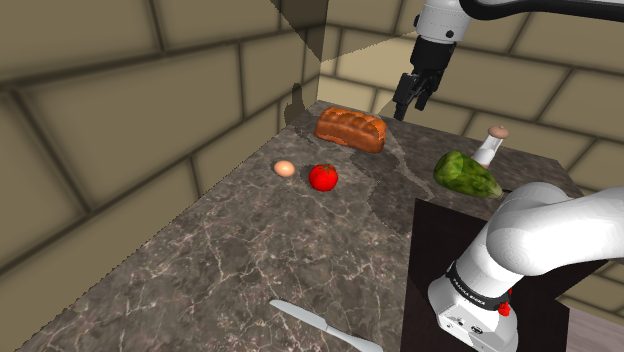} \\
    \molmotile{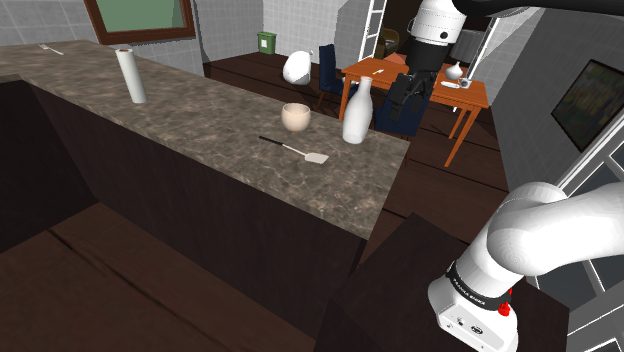} & \molmotile{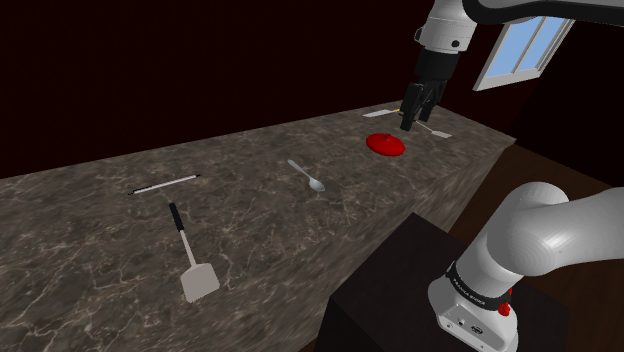} & \molmotile{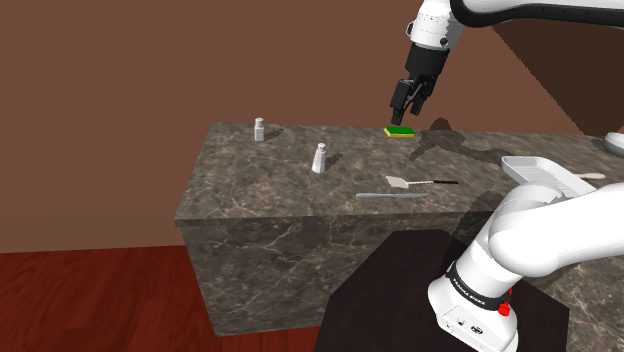} & \molmotile{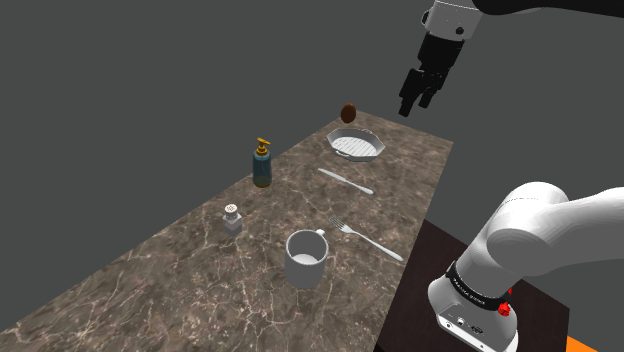} & \molmotile{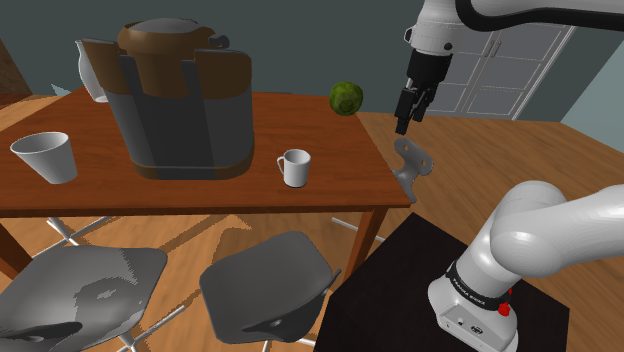} & \molmotile{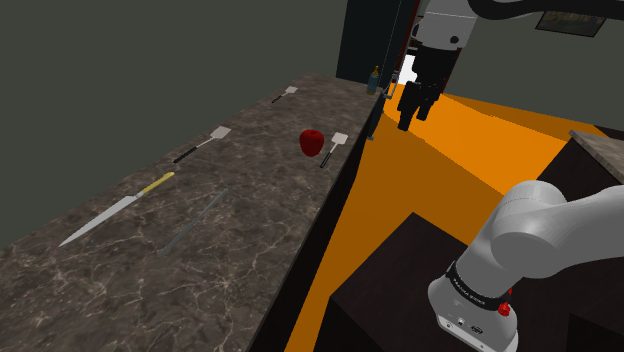} \\
    \molmotile{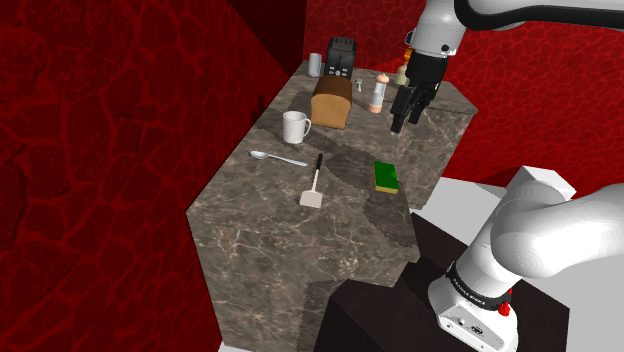} & \molmotile{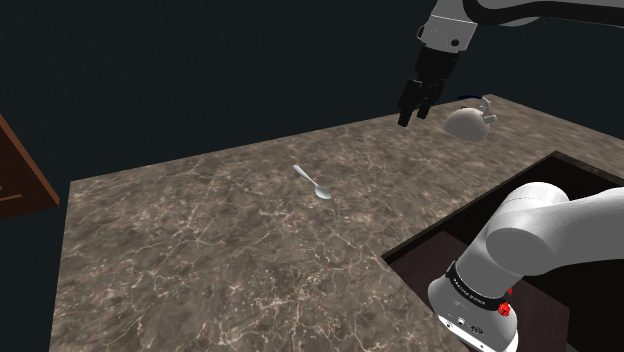} & \molmotile{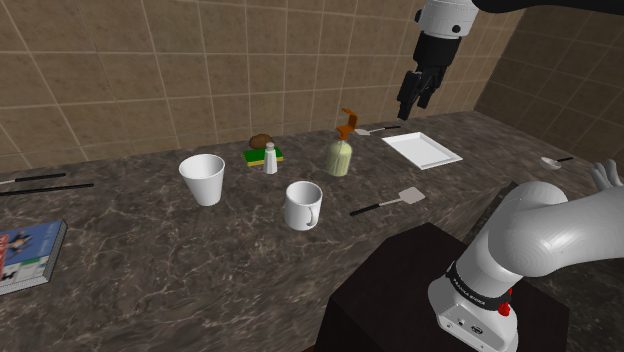} & \molmotile{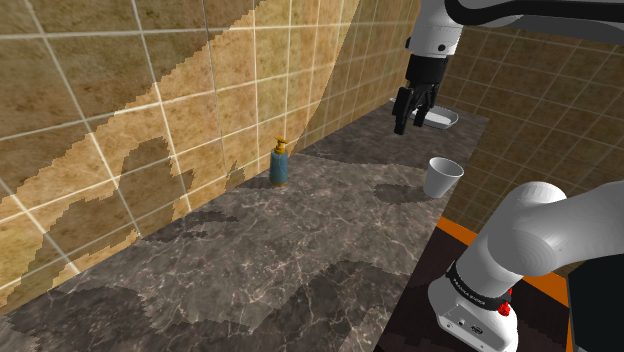} & \molmotile{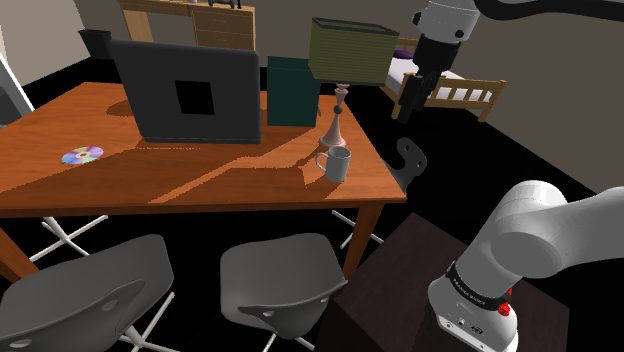} & \molmotile{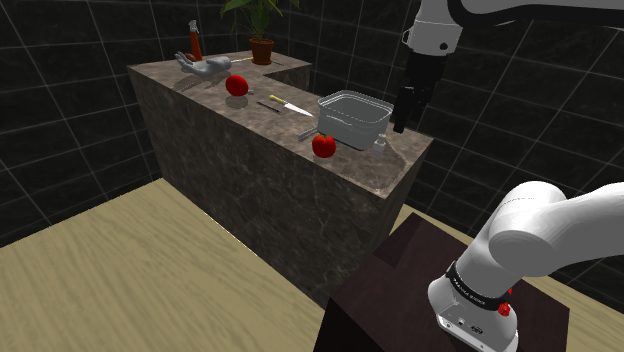} \\
    \molmotile{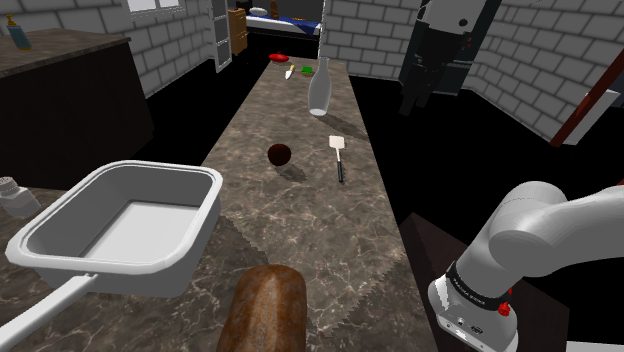} & \molmotile{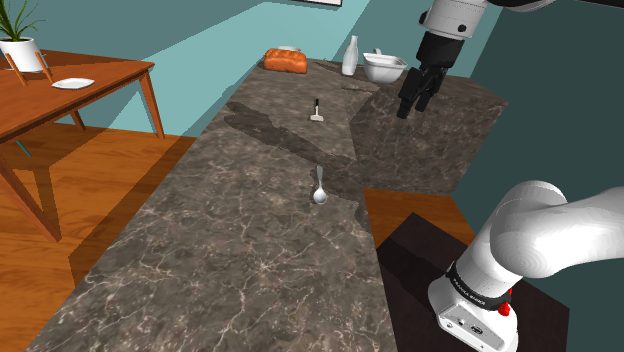} & \molmotile{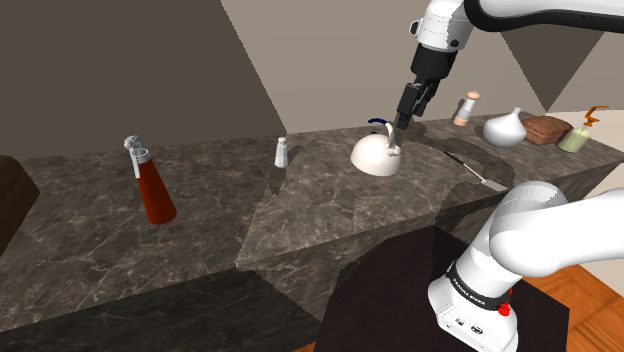} & \molmotile{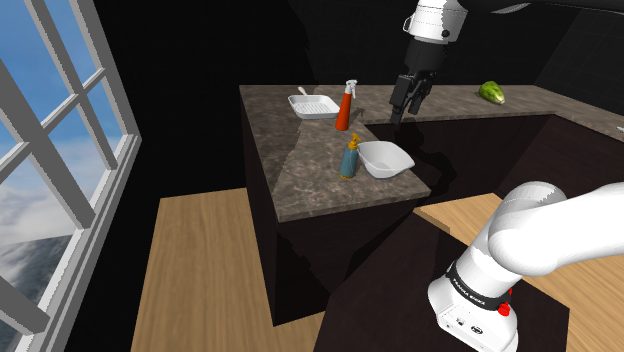} & \molmotile{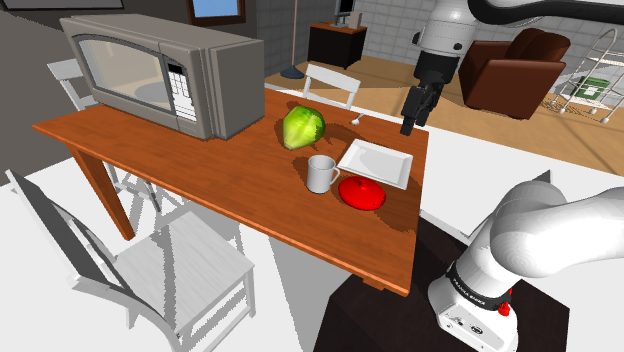} & \molmotile{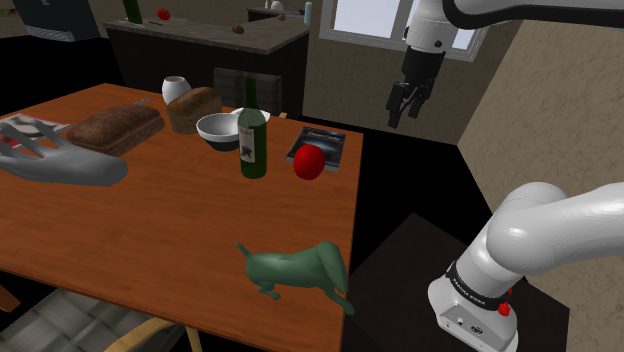} \\
    \molmotile{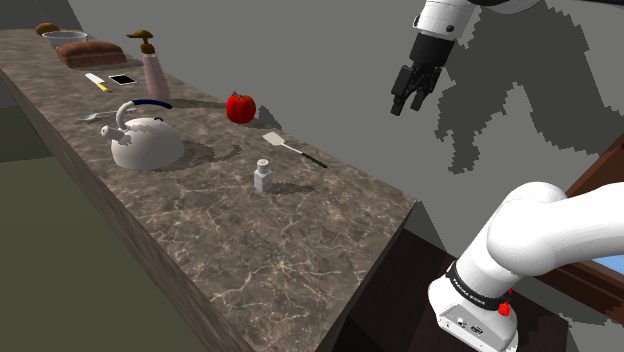} & \molmotile{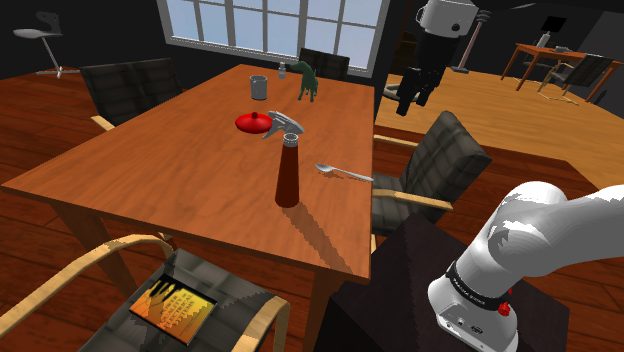} & \molmotile{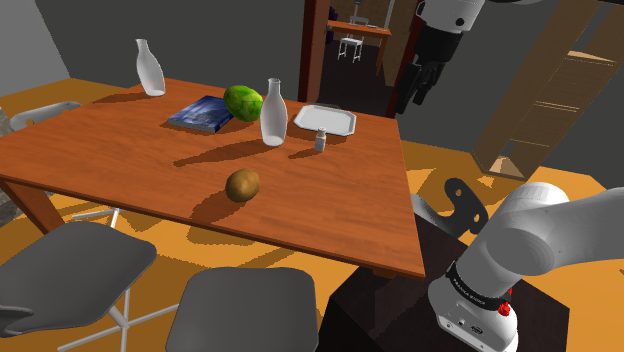} & \molmotile{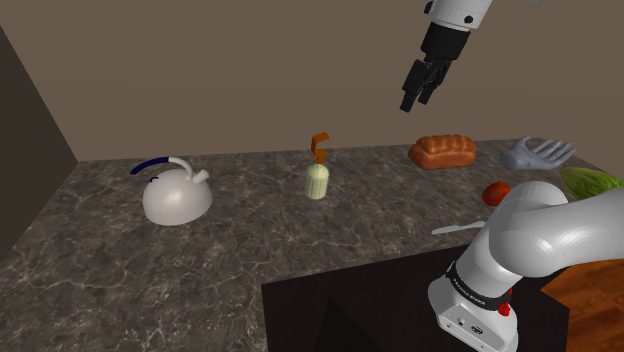} & \molmotile{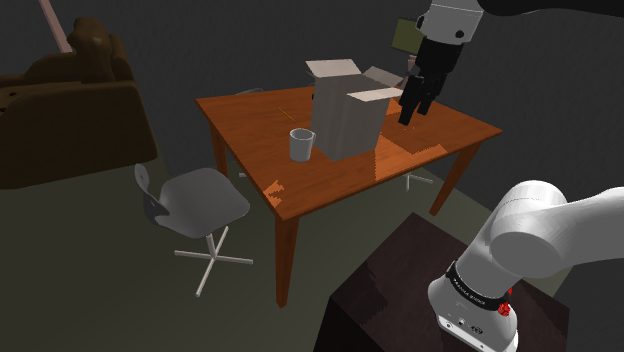} & \molmotile{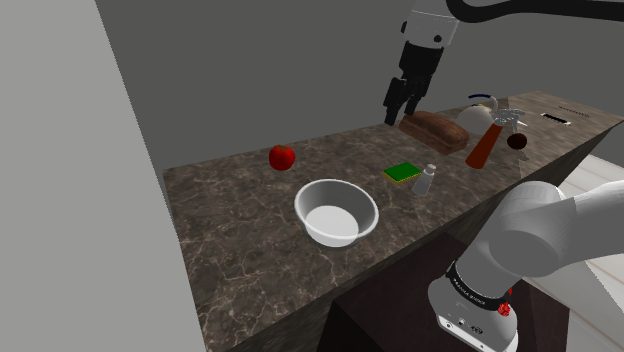} \\
    \molmotile{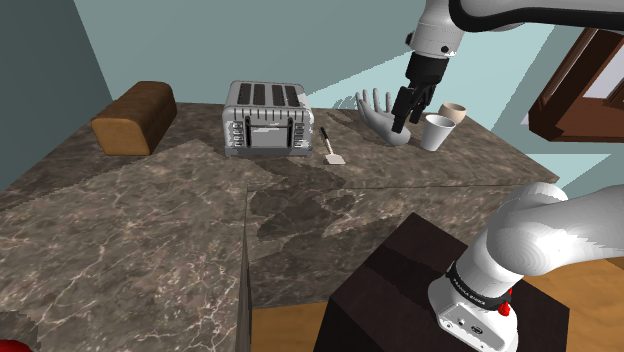} & \molmotile{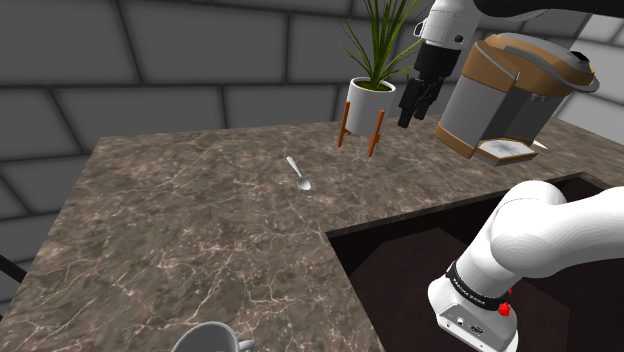} & \molmotile{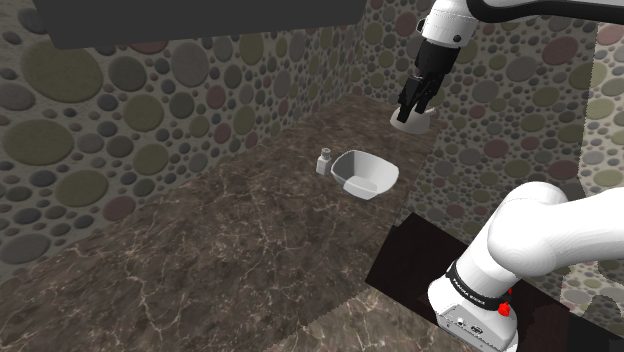} & \molmotile{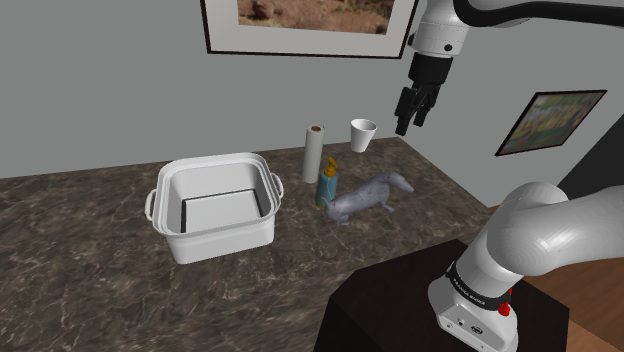} & \molmotile{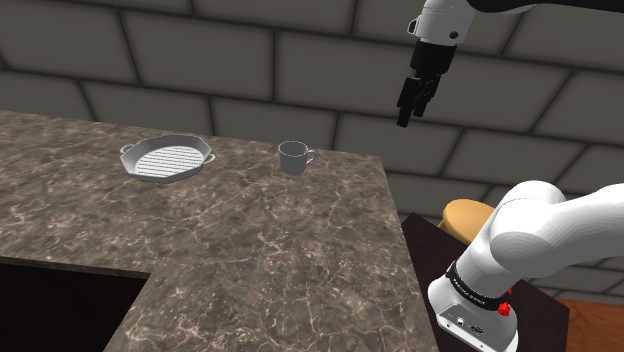} & \molmotile{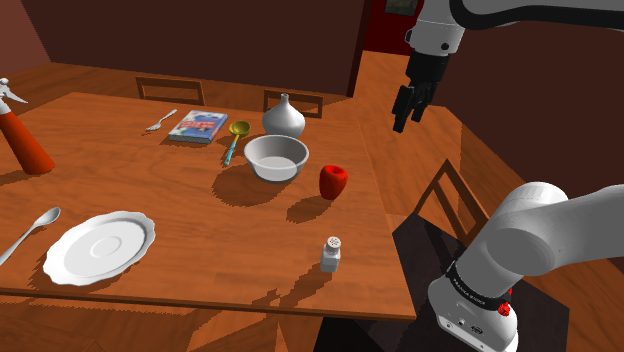} \\
    \molmotile{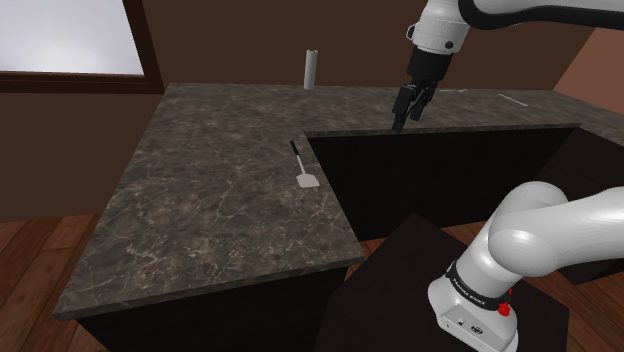} & \molmotile{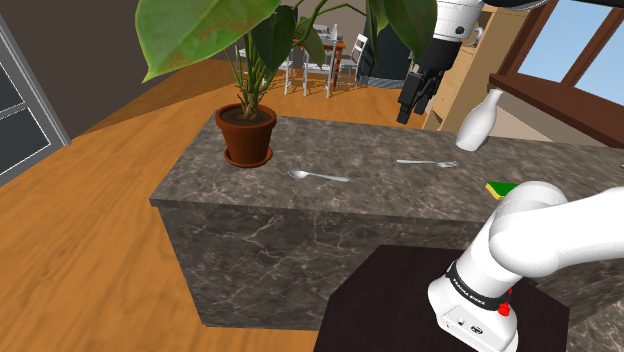} & \molmotile{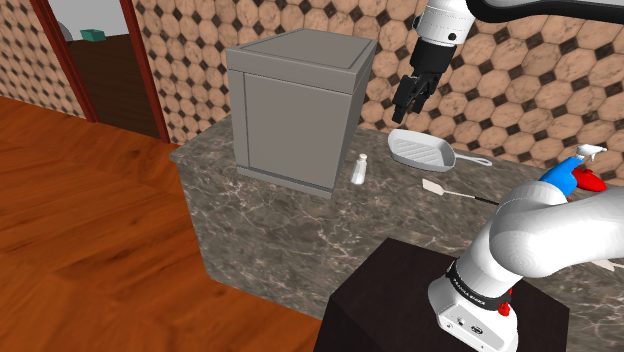} & \molmotile{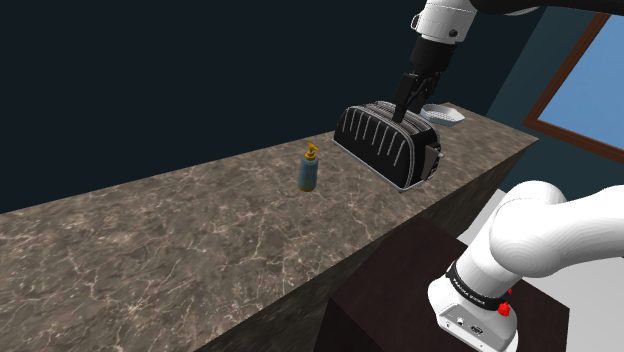} & \molmotile{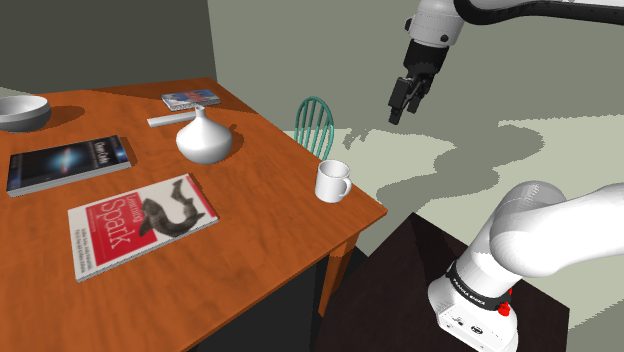} & \molmotile{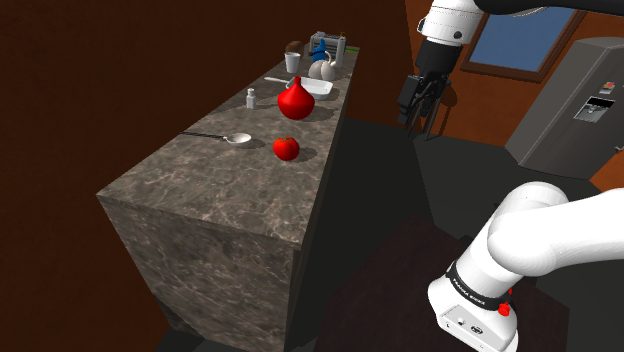} \\
    \molmotile{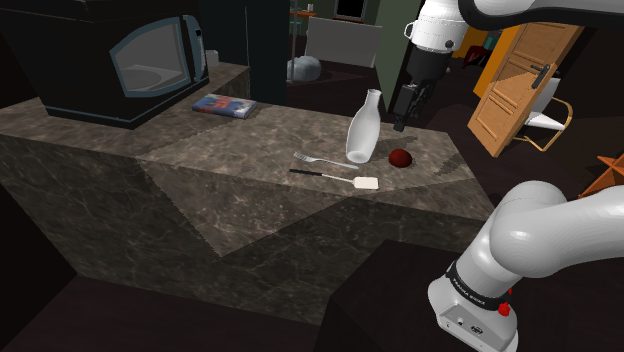} & \molmotile{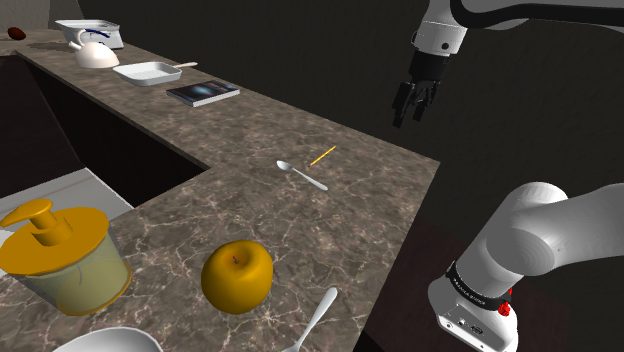} & \molmotile{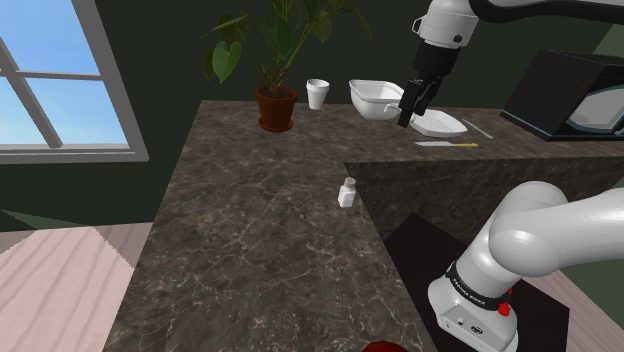} & \molmotile{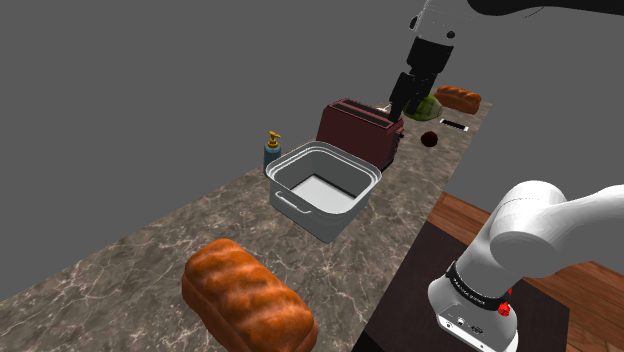} & \molmotile{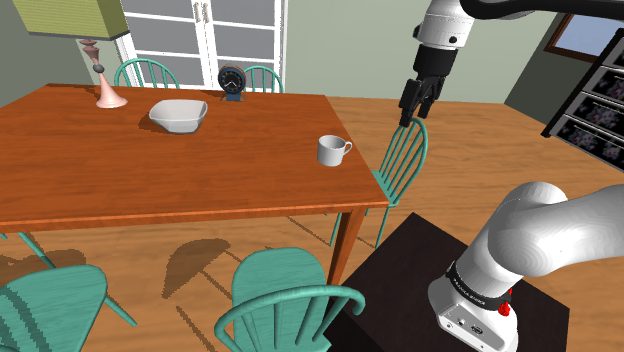} & \molmotile{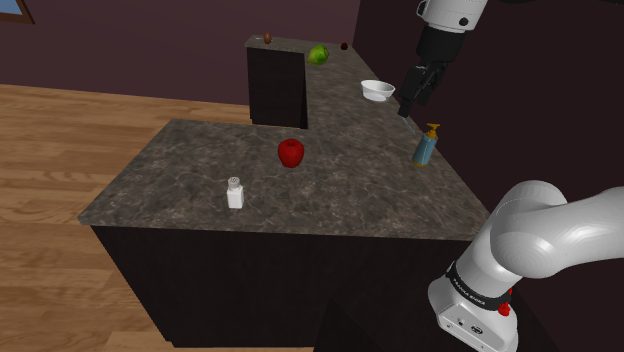} \\
    \molmotile{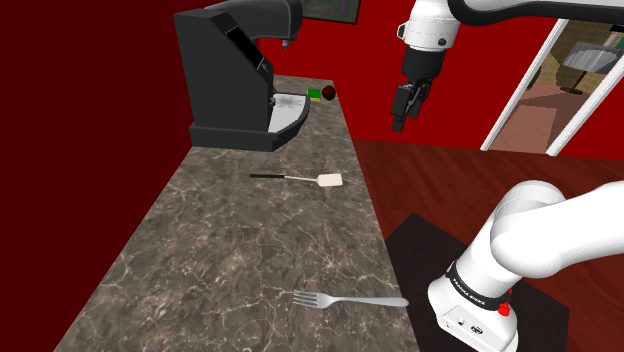} & \molmotile{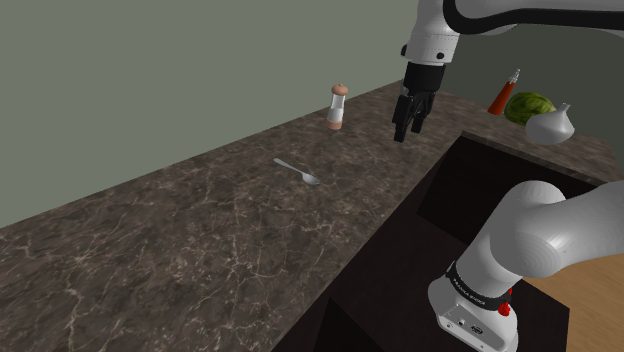} & \molmotile{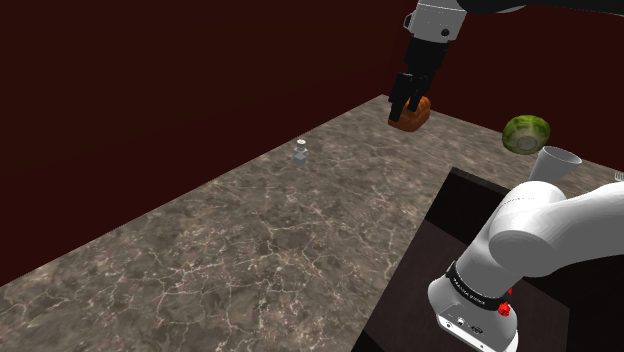} & \molmotile{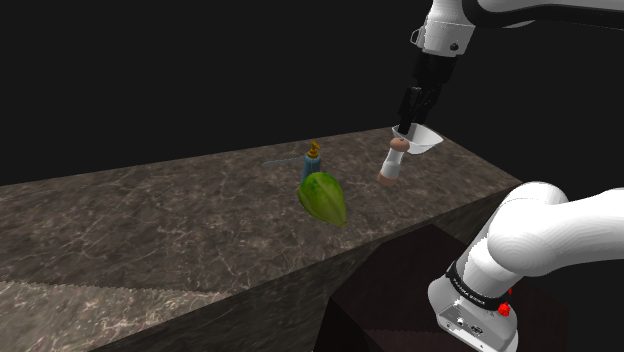} & \molmotile{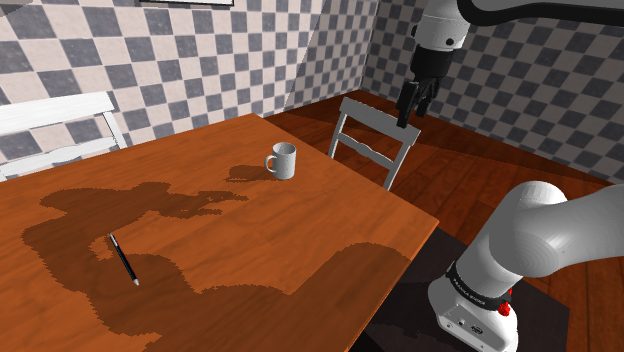} & \molmotile{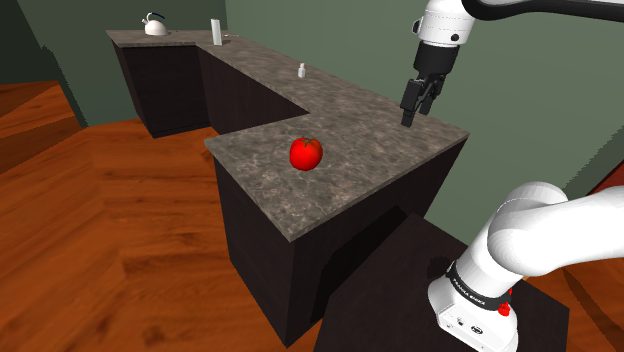} \\
    \molmotile{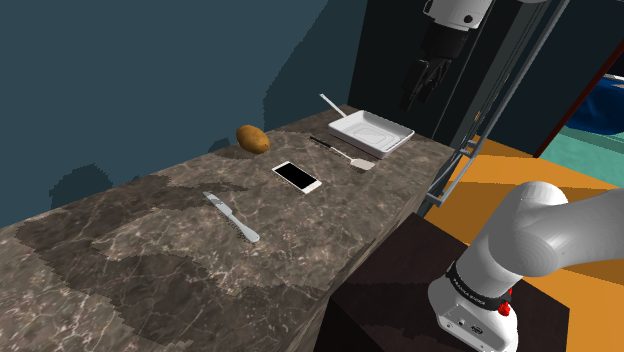} & \molmotile{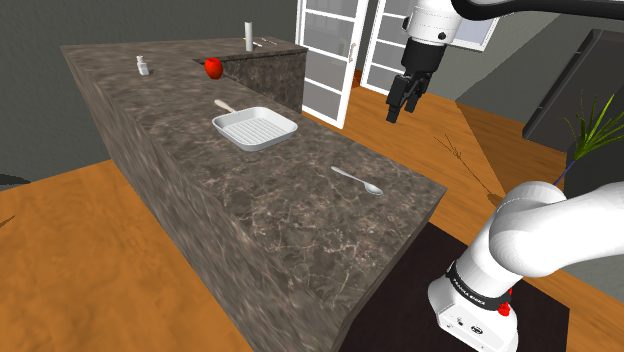} & \molmotile{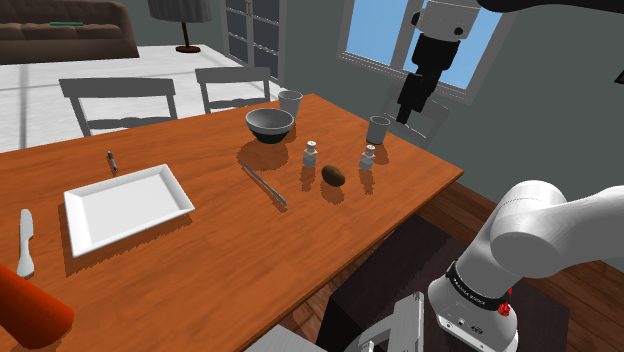} & \molmotile{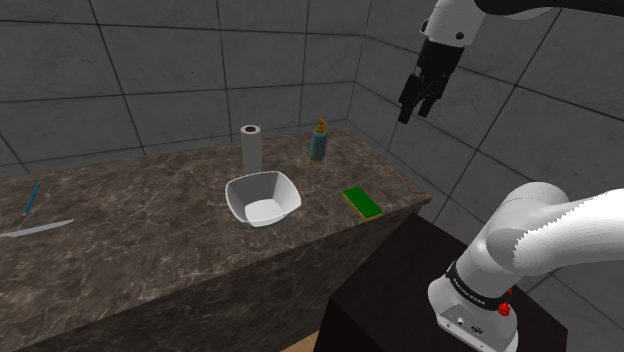} & \molmotile{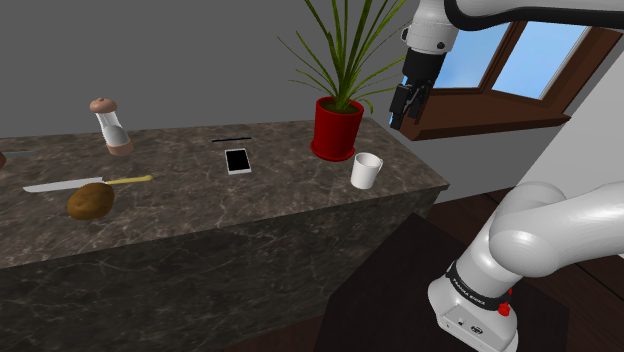} & \molmotile{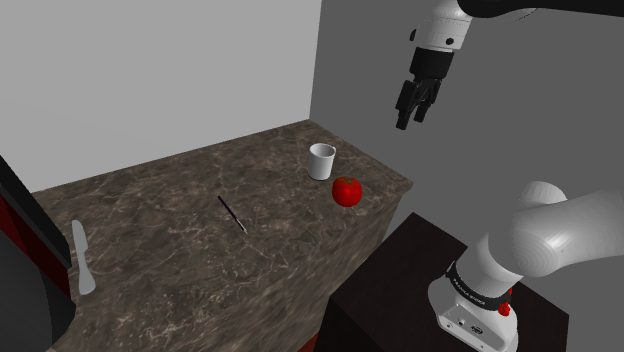} \\
  \end{tabular}

  \caption{\textbf{\texttt{MolmoSpaces} training tasks -- visualization.} Initial exterior-camera frame of every training task of the 144-task split, ordered as in Table~\ref{tab:molmo_splits}. Task IDs are shown in the lower-left corner of each frame.}
  \label{fig:molmo_visualization_training}
\end{figure}

\begin{figure}[p]
  \centering
  \setlength{\tabcolsep}{12pt}
  \renewcommand{\arraystretch}{0.5}
  \newcommand{\molmopic}{0.1\textwidth}
  \newcommand{\molmotile}[1]{\tikz[baseline=(img.south)]{%
    \node[inner sep=0pt, anchor=south west] (img) at (0,0) {\includegraphics[width=\molmopic]{figs/render/molmo_pics/molmo_#1.jpg}};%
    \node[anchor=south west, inner sep=0.8pt, fill=white, fill opacity=0.75, text opacity=1, font=\fontsize{5}{4}\selectfont] at (img.south west) {#1};}}
  \newcommand{\molmohead}[1]{\makebox[\molmopic][c]{\footnotesize{#1}}\rule[-2pt]{0pt}{0pt}}

  \begin{tabular}{@{}*{6}{c}@{}}
    \molmohead{boiler} & \molmohead{box} & \molmohead{cup} & \molmohead{ladle} & \molmohead{pot} & \molmohead{tissuepaper} \\
    [1pt]
    \molmotile{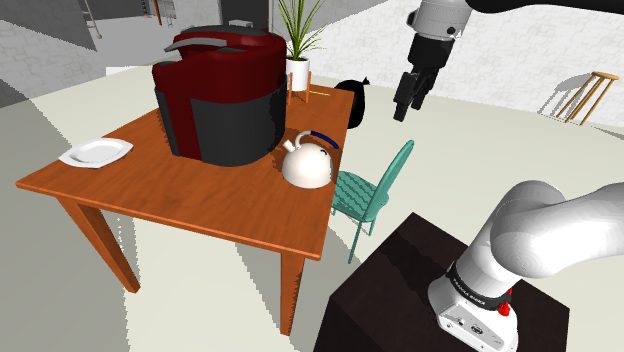} & \molmotile{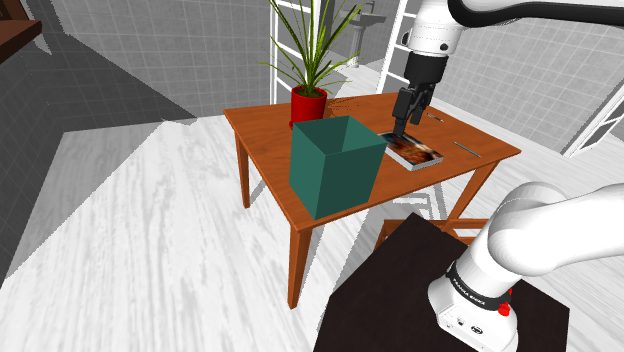} & \molmotile{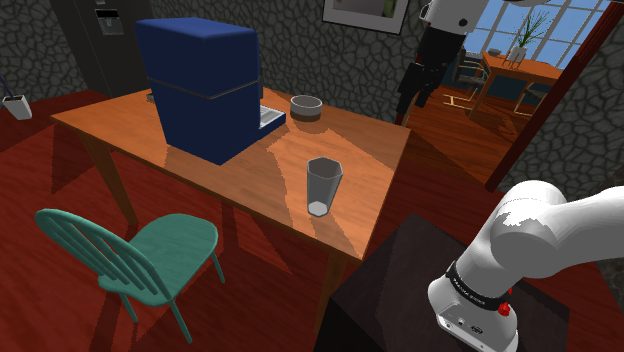} & \molmotile{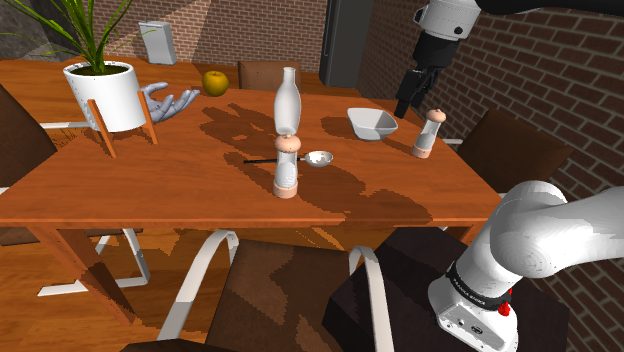} & \molmotile{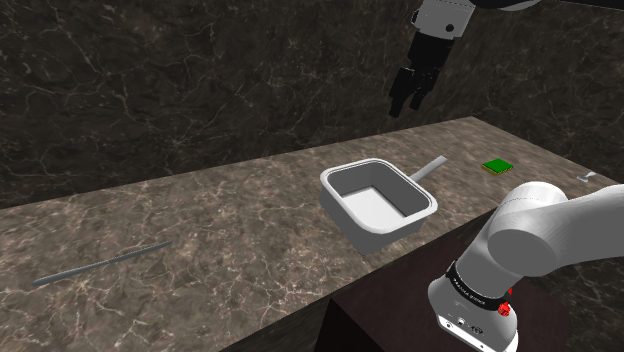} & \molmotile{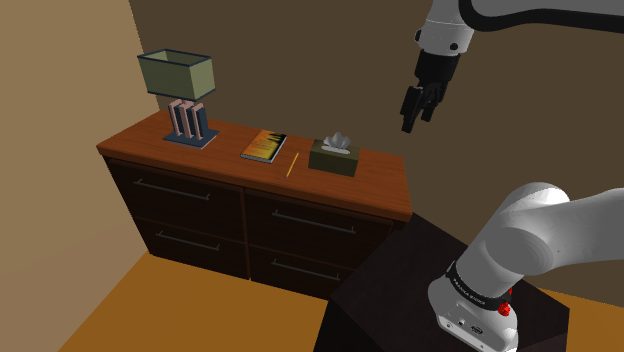} \\
    \molmotile{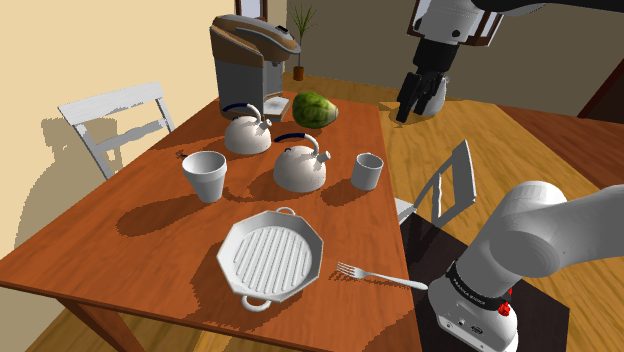} & \molmotile{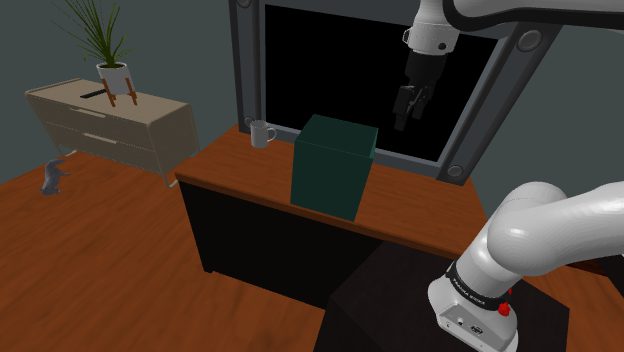} & \molmotile{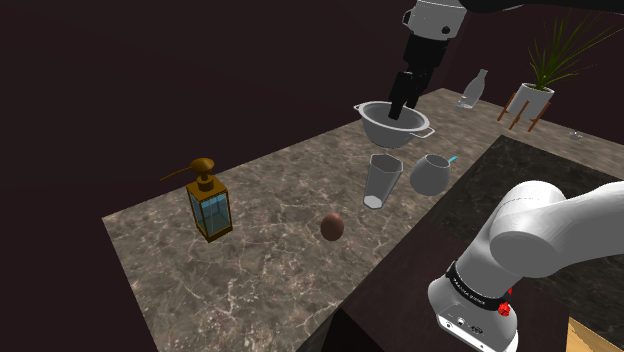} & \molmotile{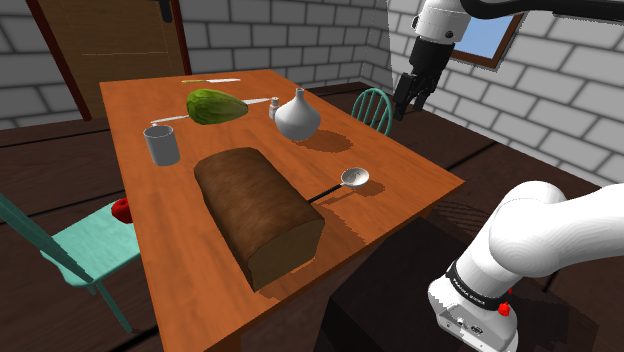} & \molmotile{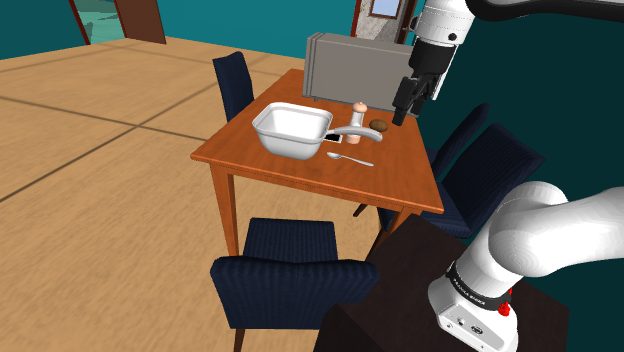} & \molmotile{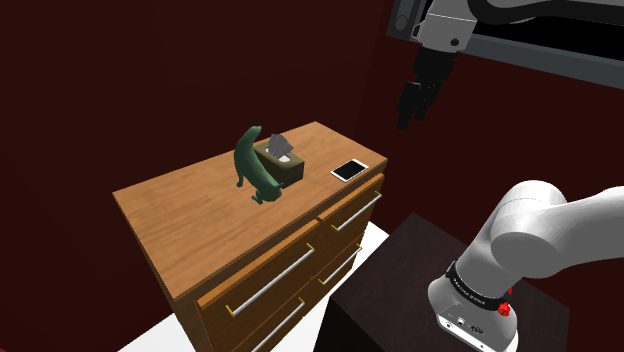} \\
    \molmotile{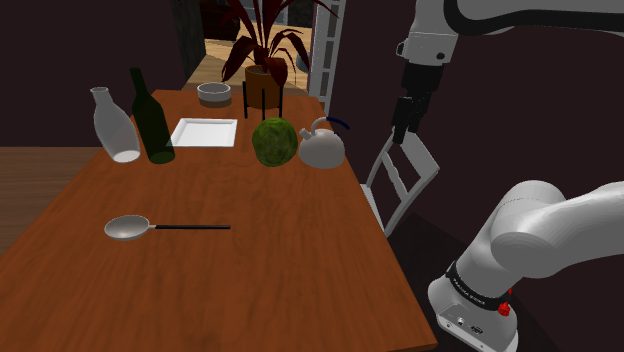} & \molmotile{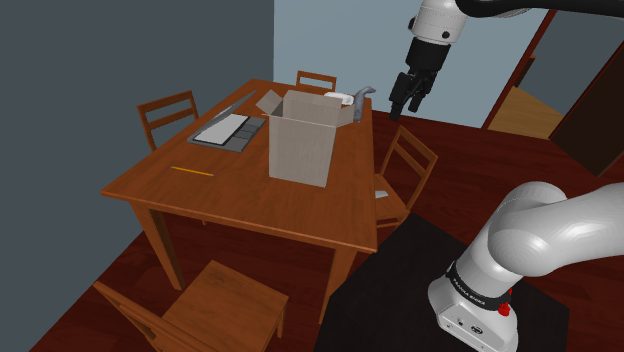} & \molmotile{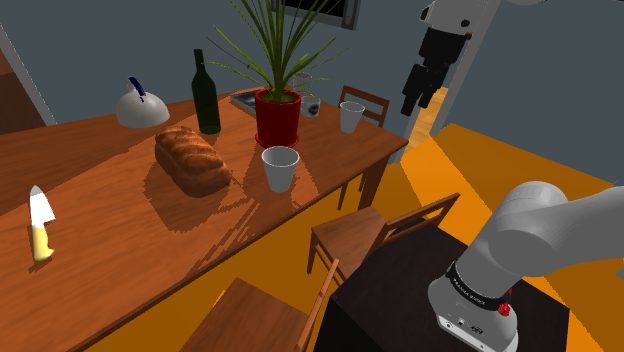} & \molmotile{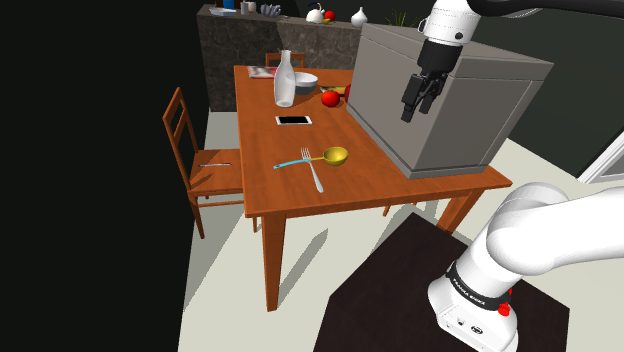} & \molmotile{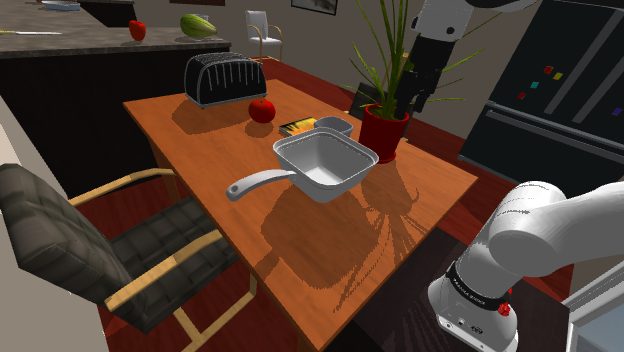} & \molmotile{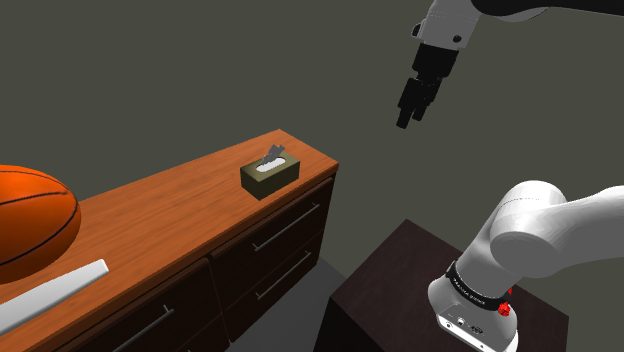} \\
    \molmotile{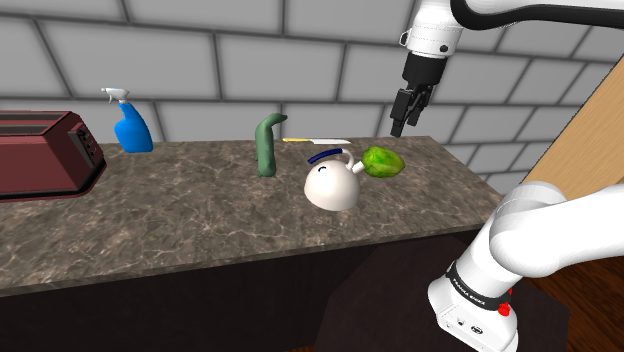} & \molmotile{273} & \molmotile{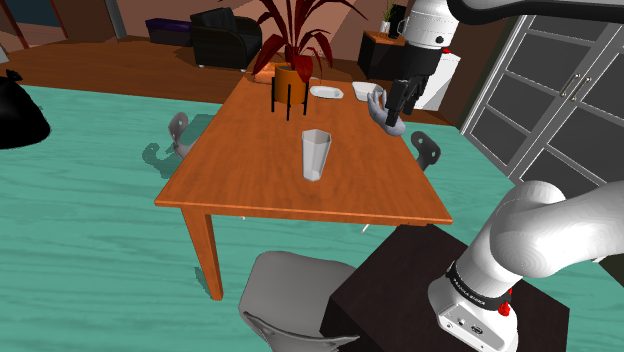} & \molmotile{73} & \molmotile{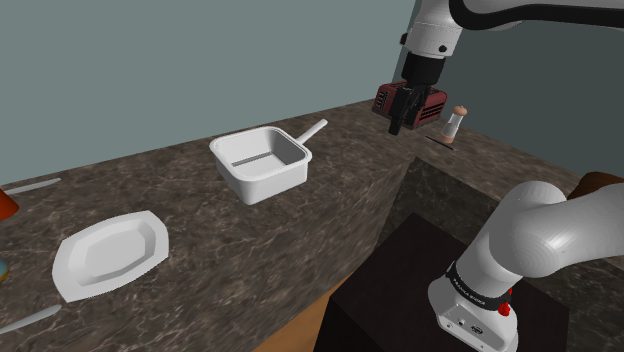} & \molmotile{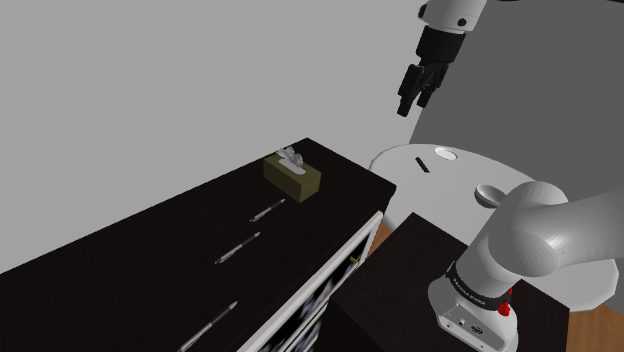} \\
    \molmotile{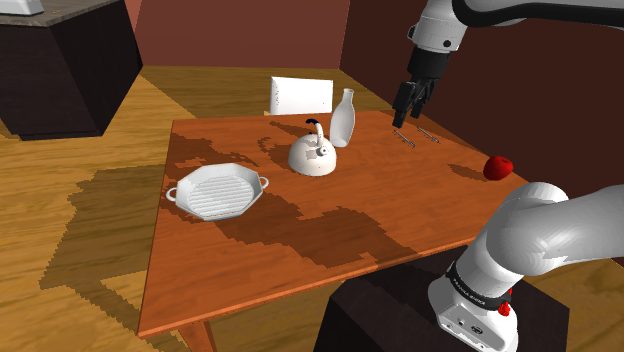} & \molmotile{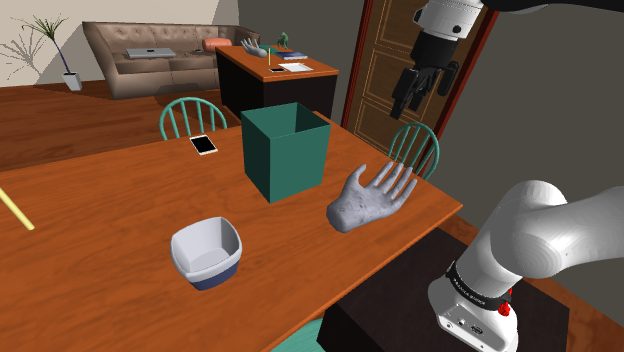} & \molmotile{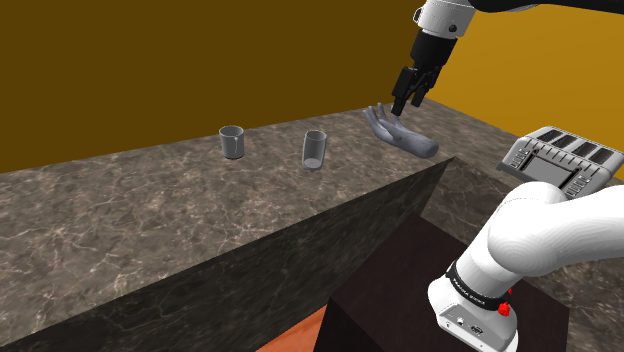} & \molmotile{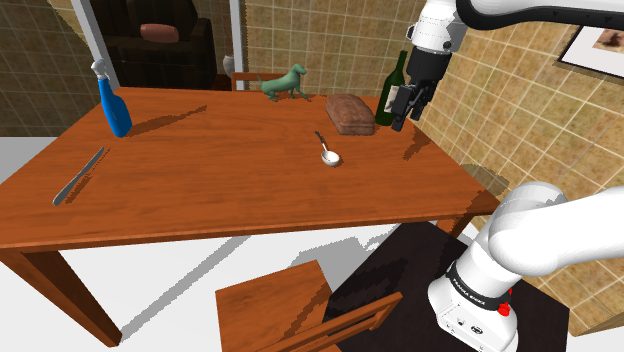} & \molmotile{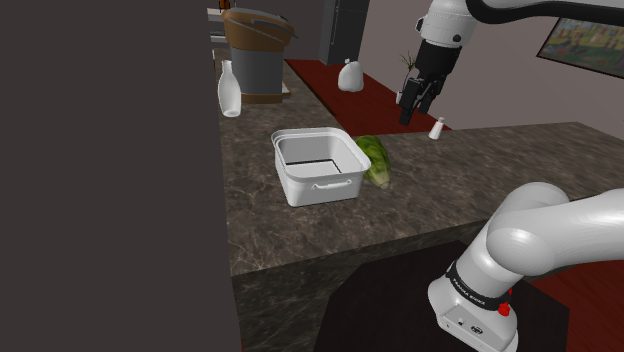} & \molmotile{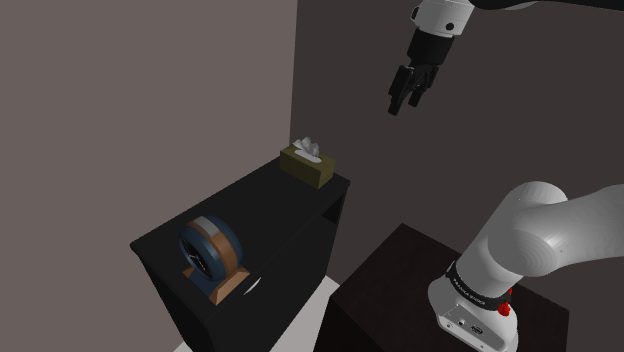} \\
    \molmotile{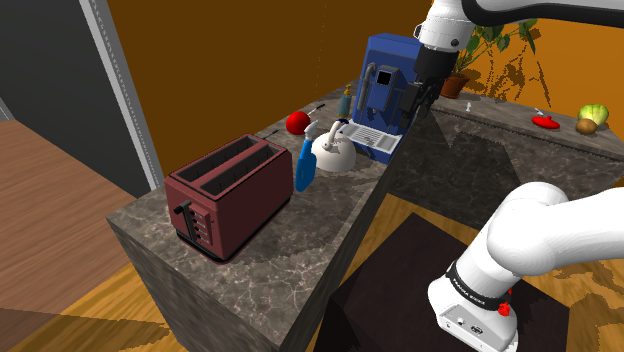} & \molmotile{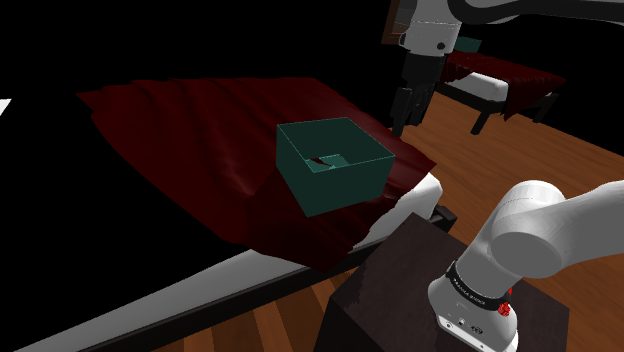} & \molmotile{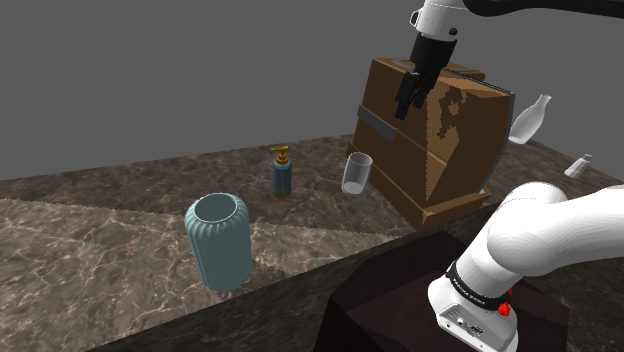} & \molmotile{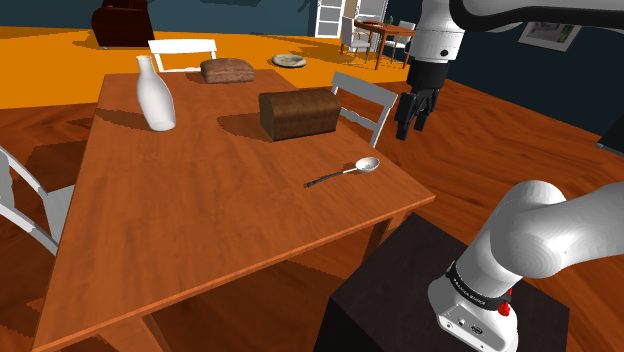} & \molmotile{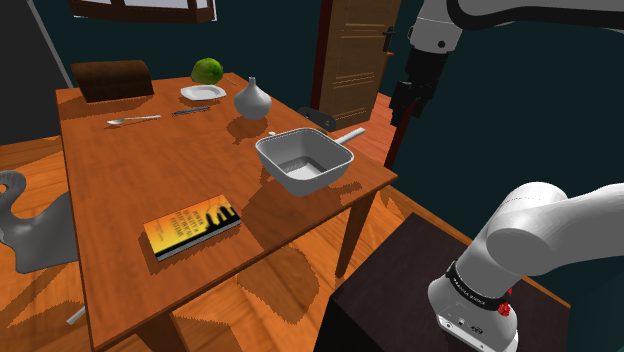} & \molmotile{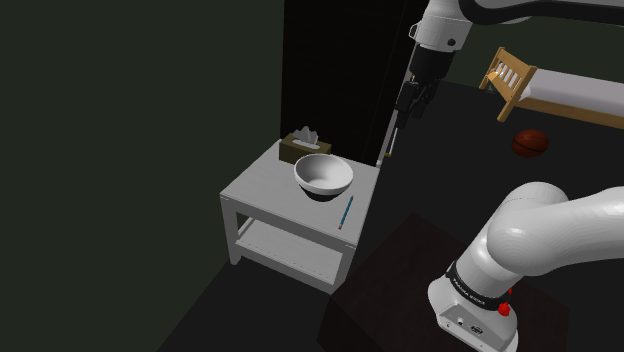} \\
    \molmotile{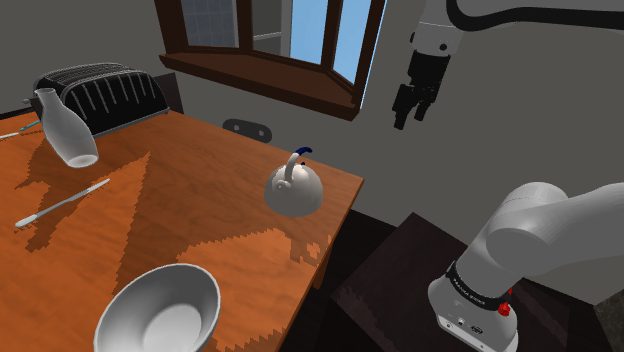} & \molmotile{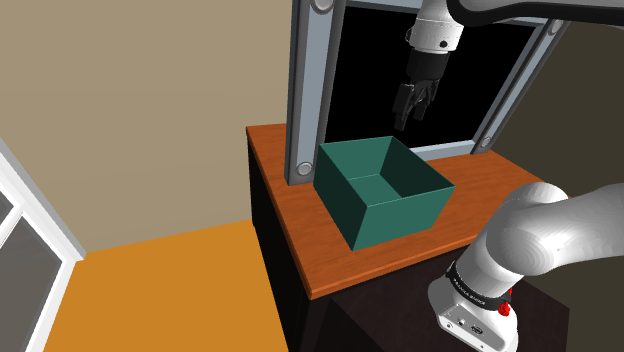} & \molmotile{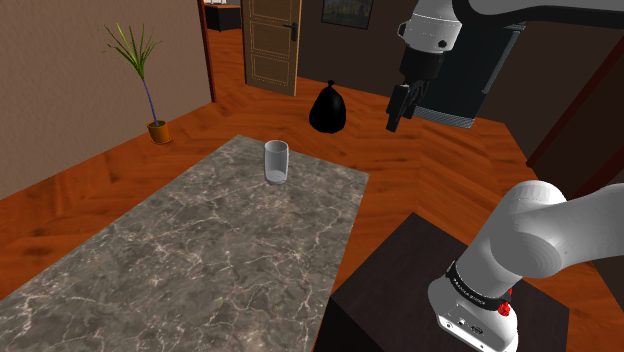} & \molmotile{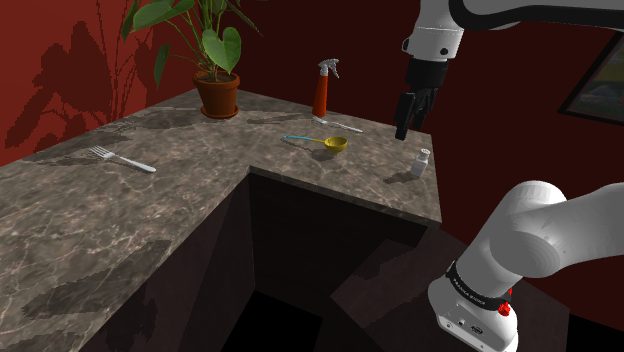} & \molmotile{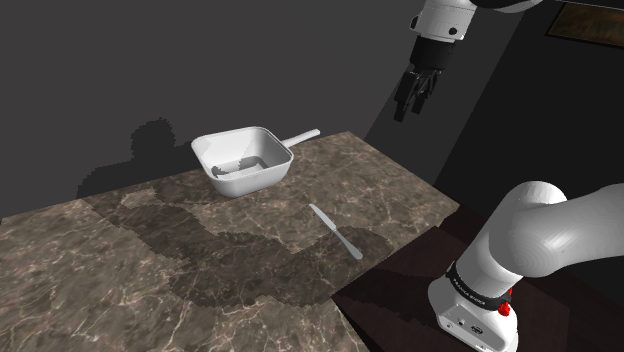} & \molmotile{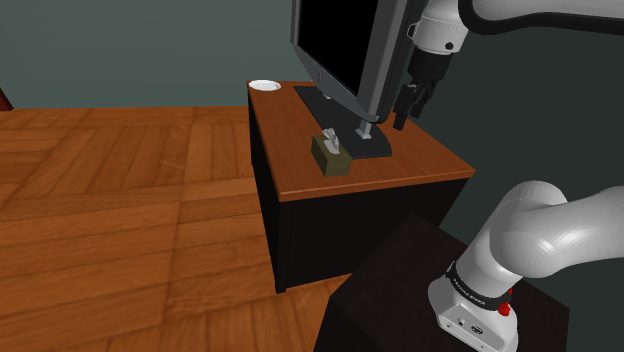} \\
    \molmotile{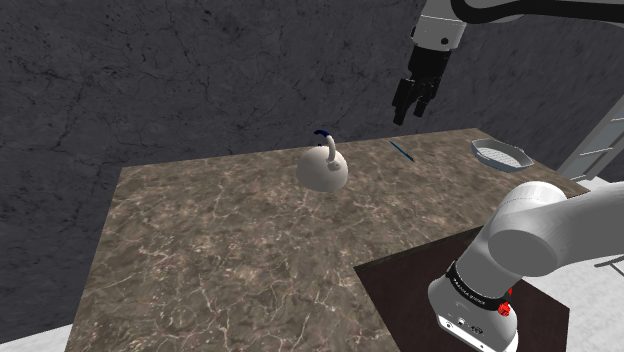} & \molmotile{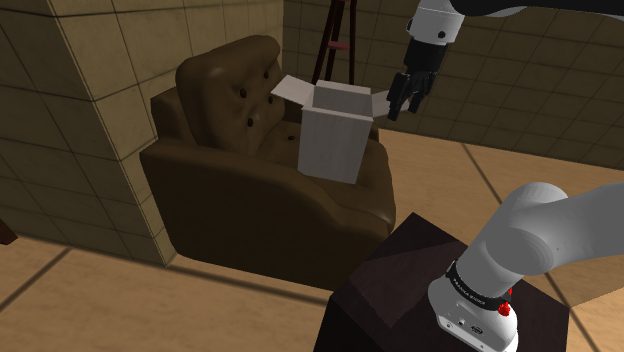} & \molmotile{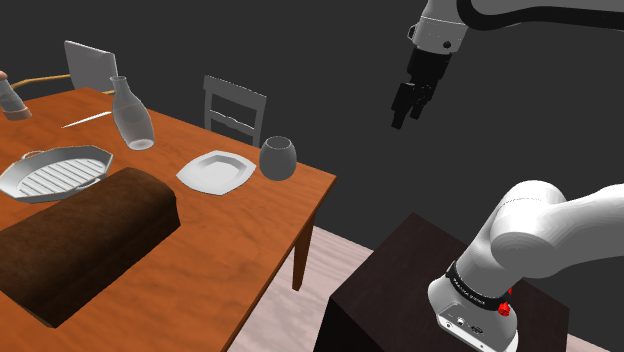} & \molmotile{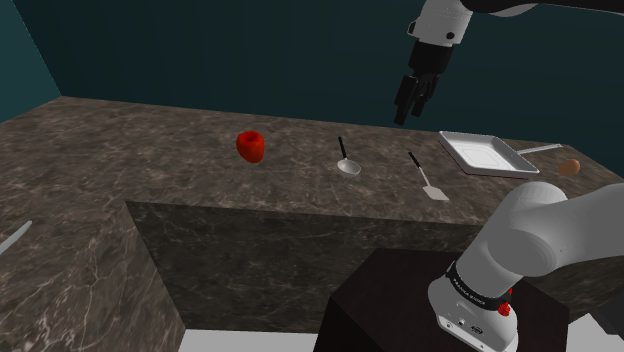} & \molmotile{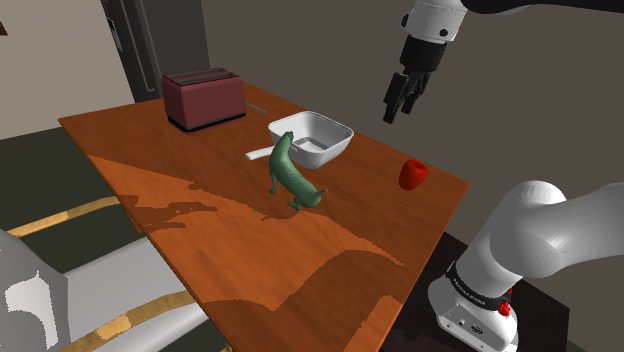} & \molmotile{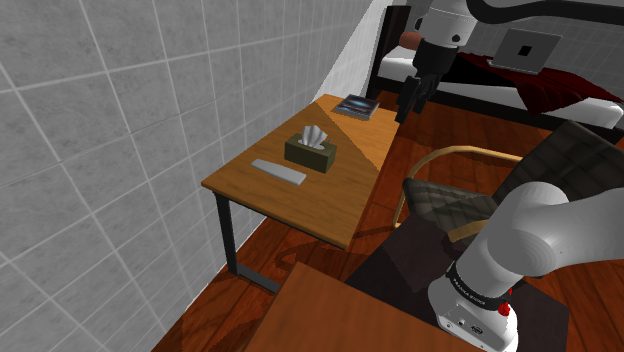} \\
  \end{tabular}
  \\
  \vspace{2mm}
  \begin{tabular}{@{}*{6}{c}@{}}
    \molmohead{spatula} & \molmohead{spoon} & \molmohead{salt/pepper shaker} & \molmohead{soapdispenser} & \molmohead{mug} & \molmohead{fruit/veg} \\
    [1pt]
    \molmotile{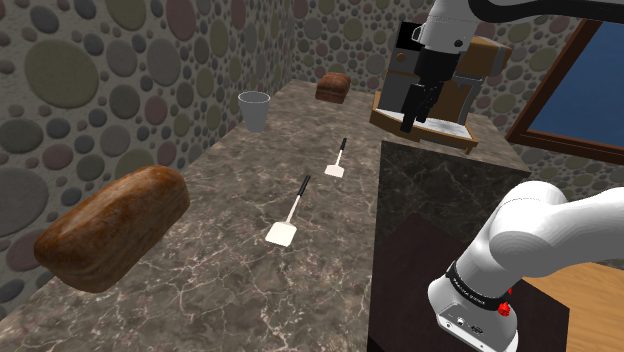} & \molmotile{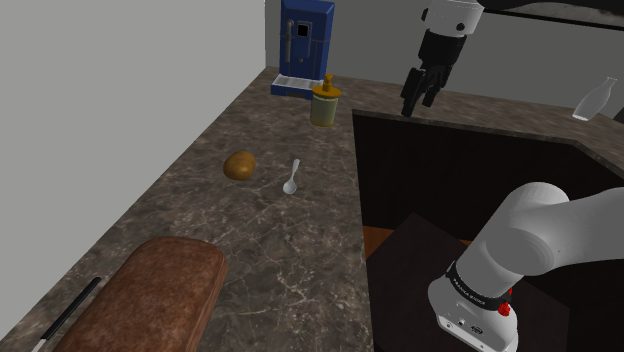} & \molmotile{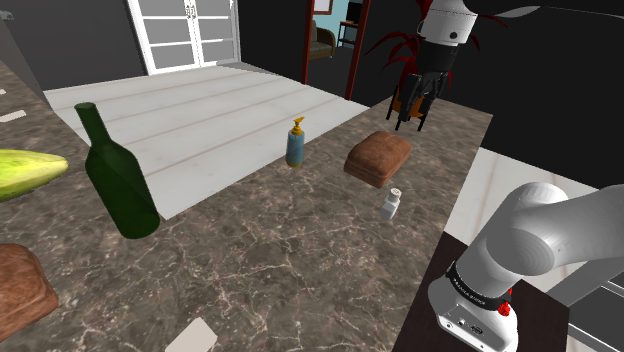} & \molmotile{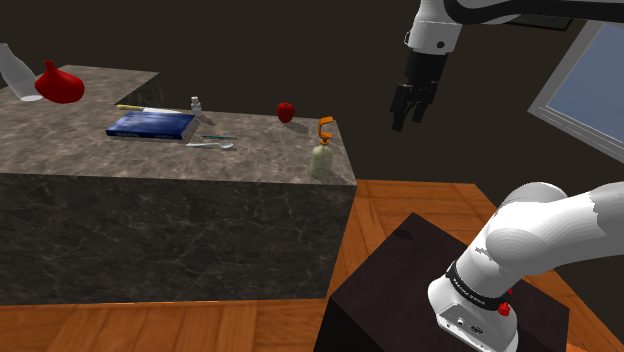} & \molmotile{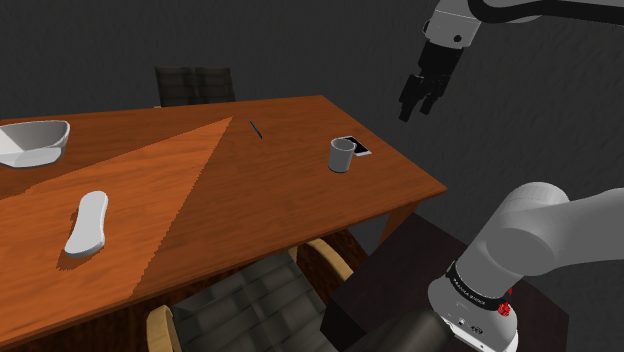} & \molmotile{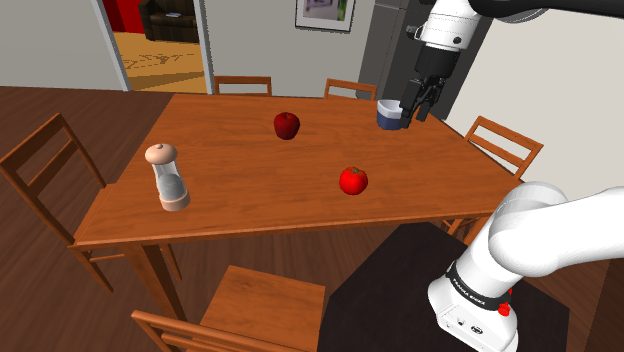} \\
    \molmotile{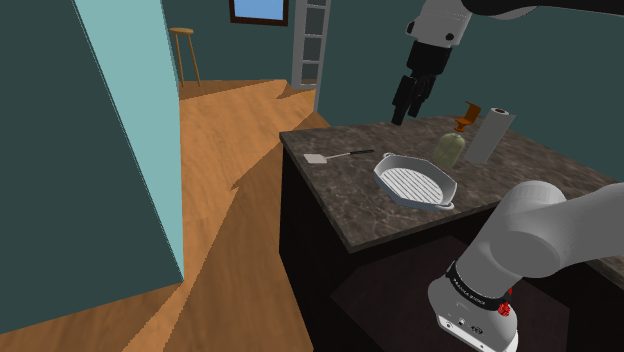} & \molmotile{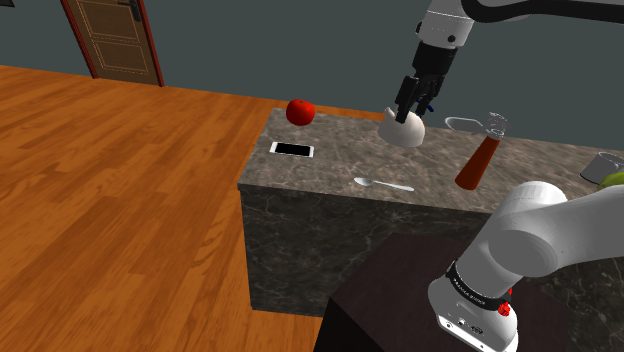} & \molmotile{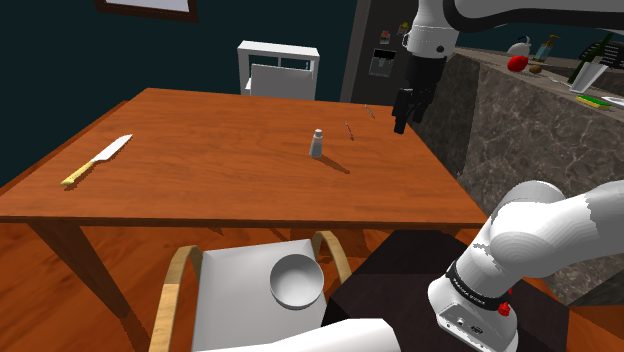} & \molmotile{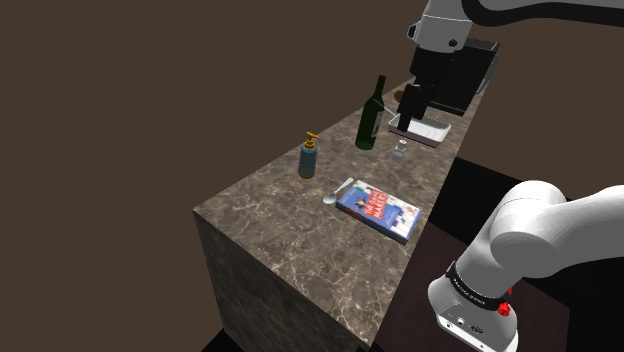} & \molmotile{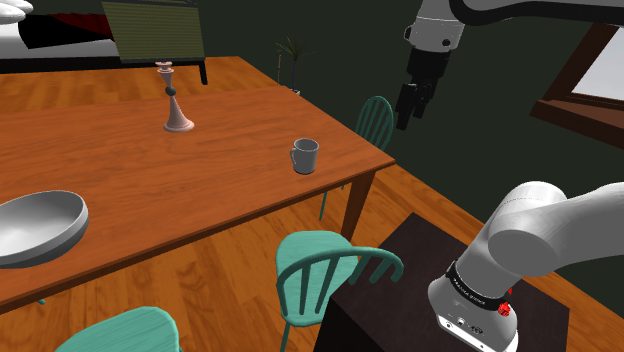} & \molmotile{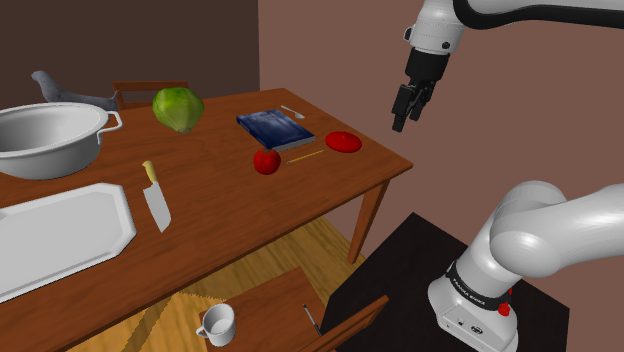} \\
    \molmotile{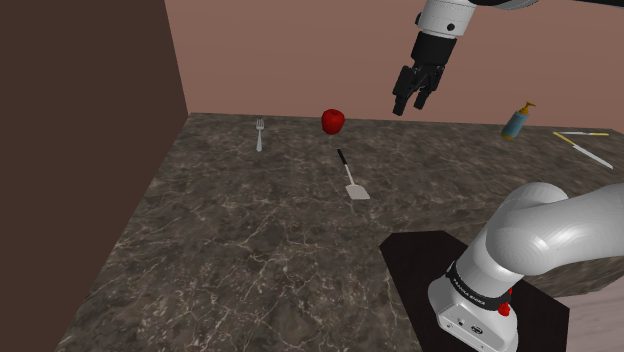} & \molmotile{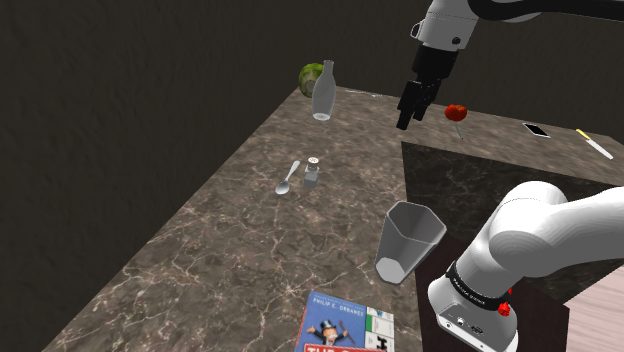} & \molmotile{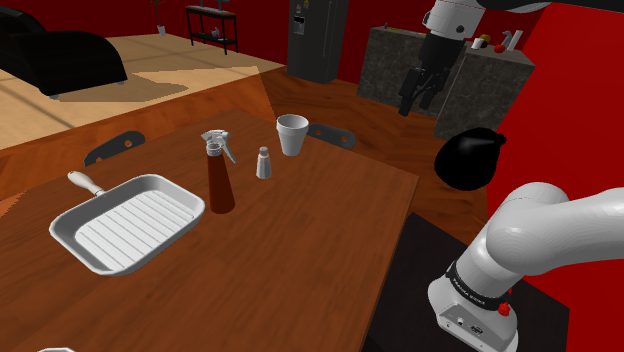} & \molmotile{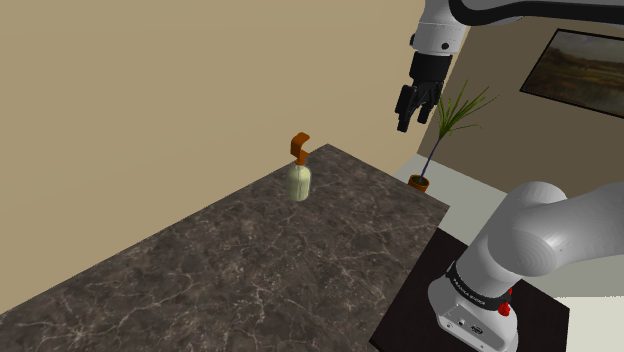} & \molmotile{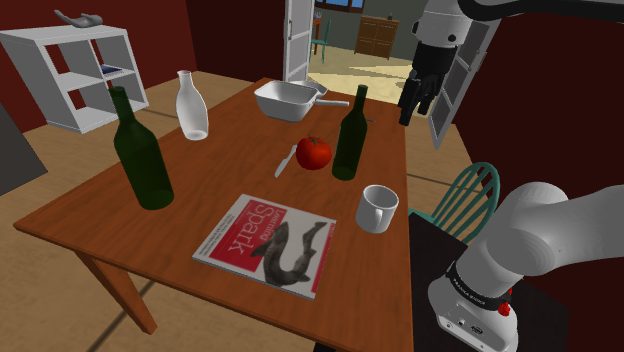} & \molmotile{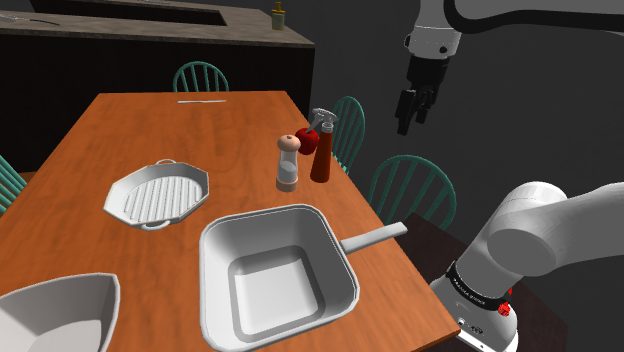} \\
    \molmotile{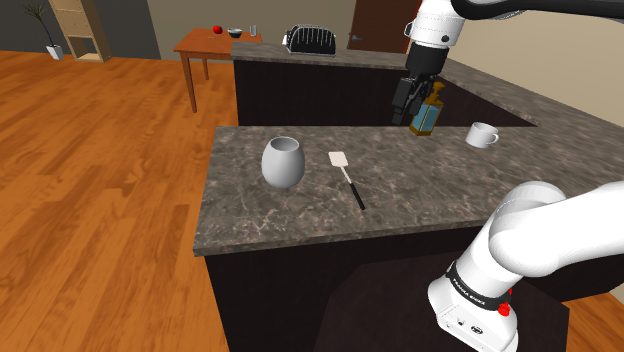} & \molmotile{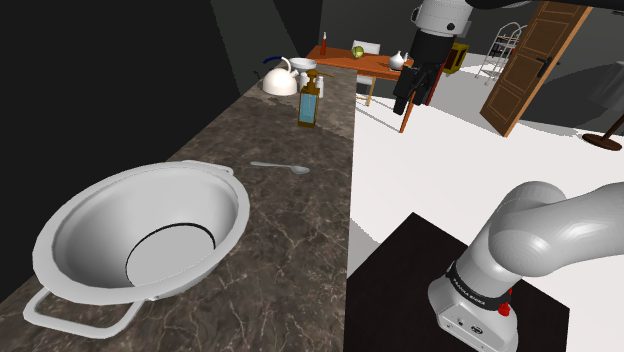} & \molmotile{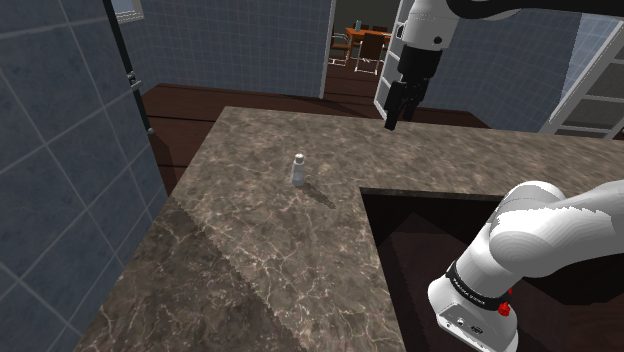} & \molmotile{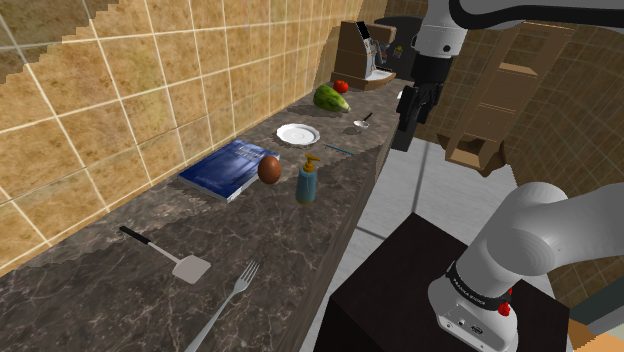} & \molmotile{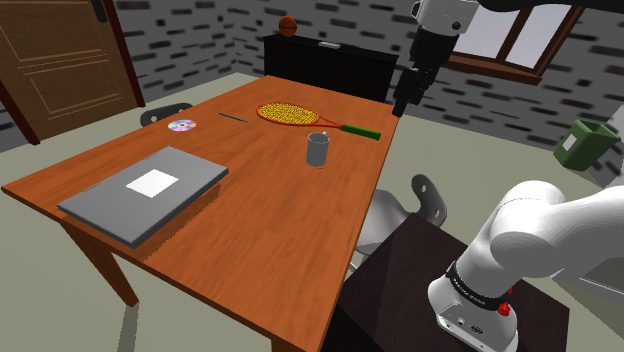} & \molmotile{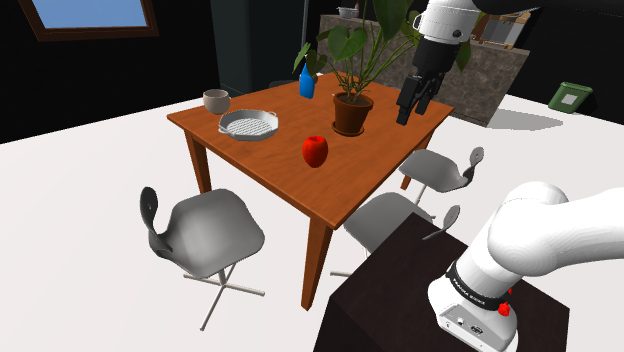} \\
    \molmotile{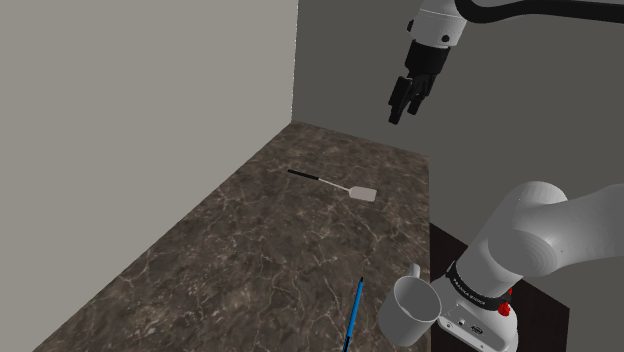} & \molmotile{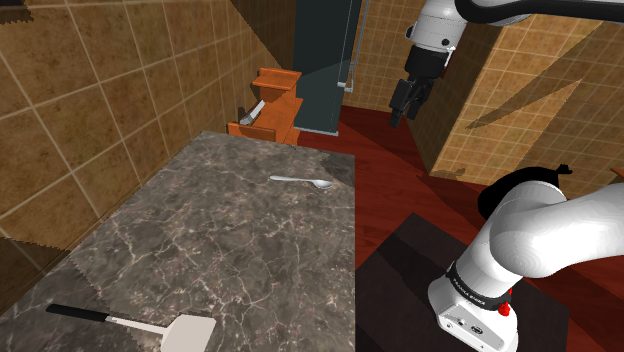} & \molmotile{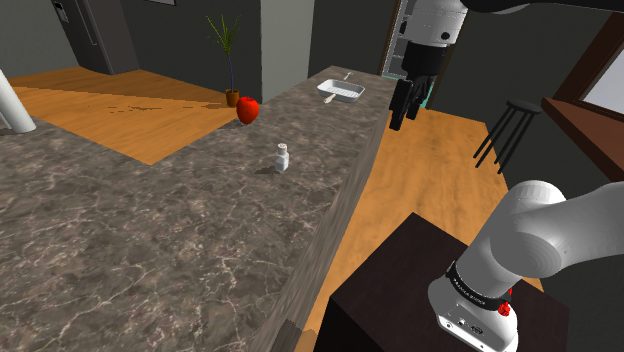} & \molmotile{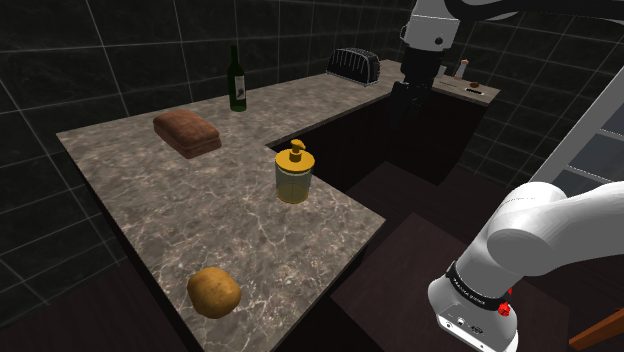} & \molmotile{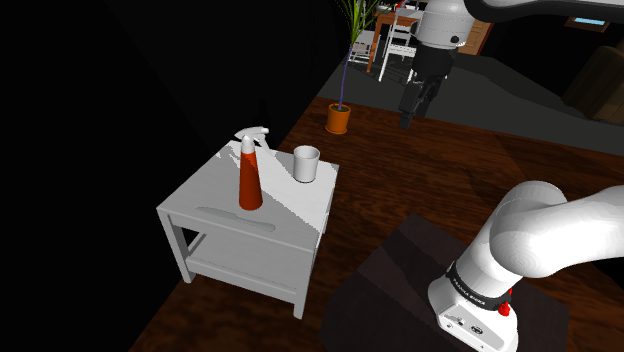} & \molmotile{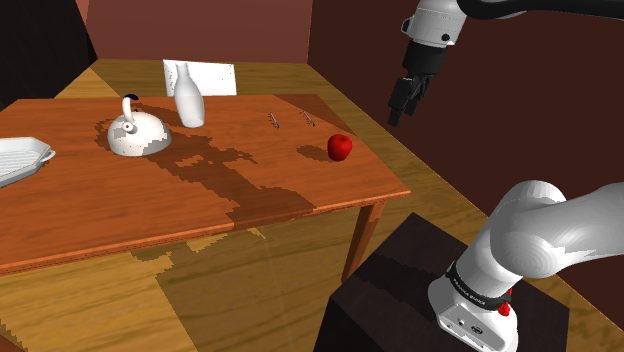} \\
    \molmotile{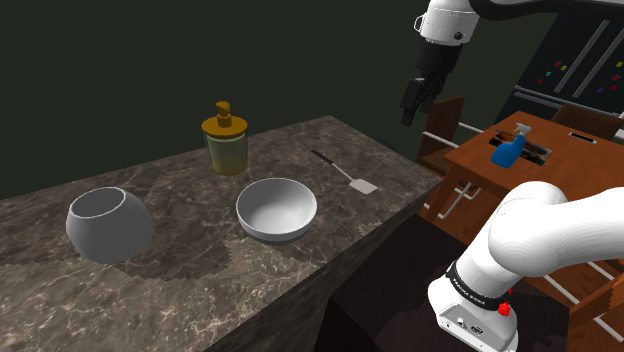} & \molmotile{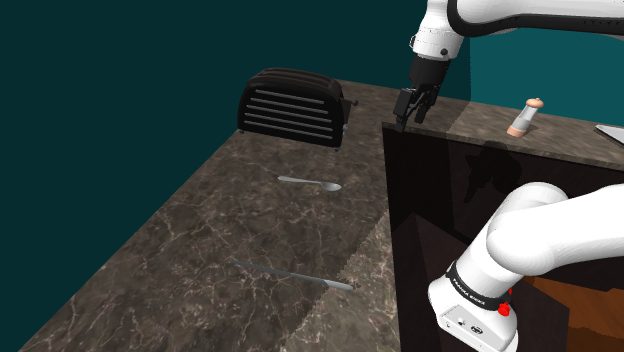} & \molmotile{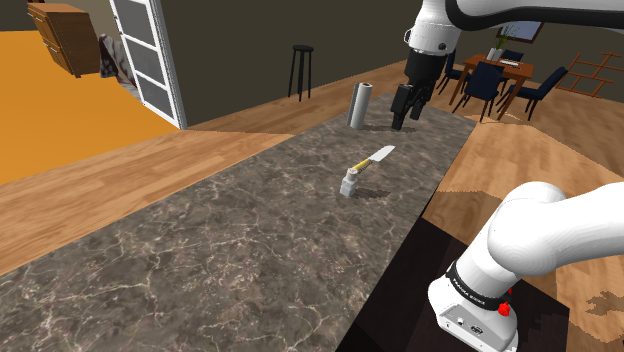} & \molmotile{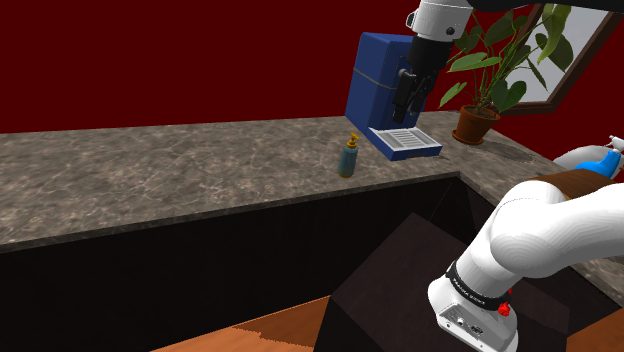} & \molmotile{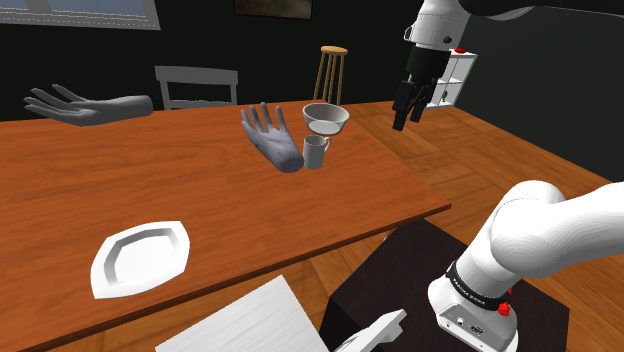} & \molmotile{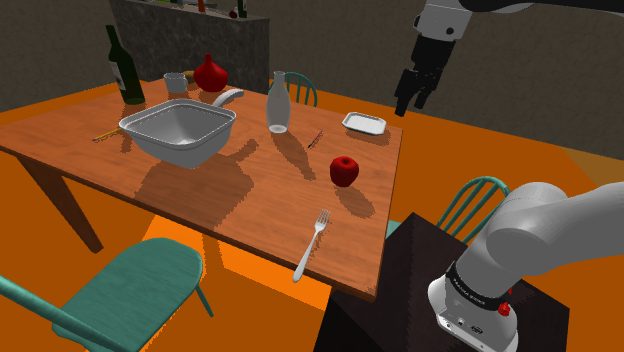} \\
    \molmotile{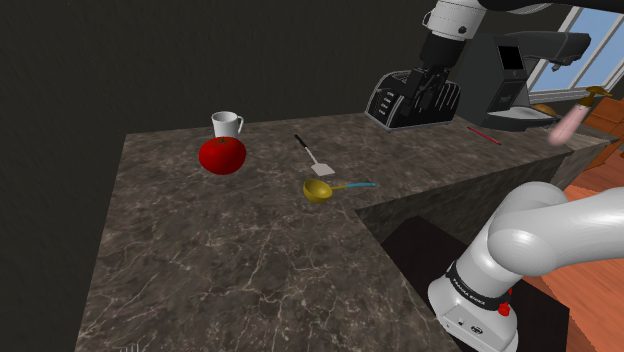} & \molmotile{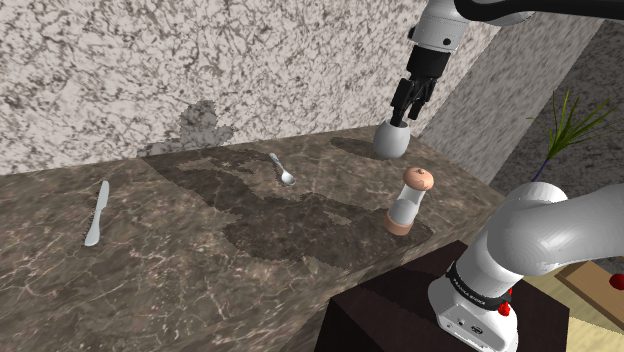} & \molmotile{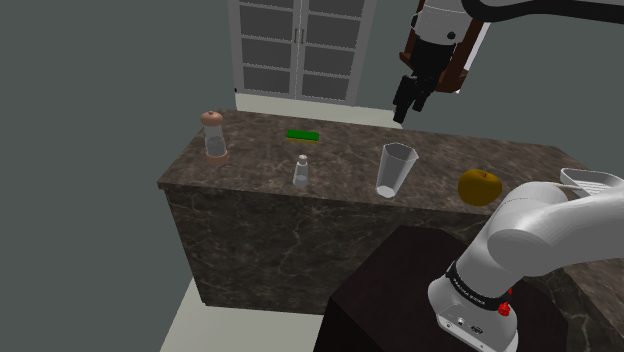} & \molmotile{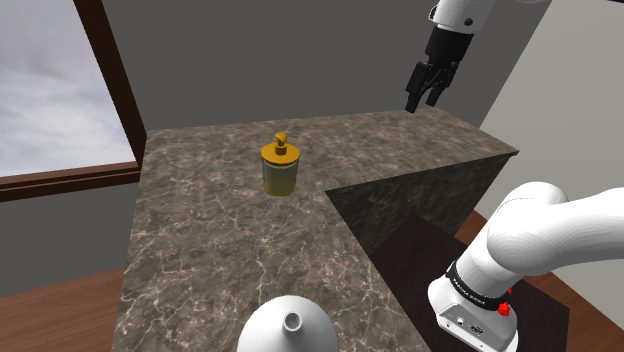} & \molmotile{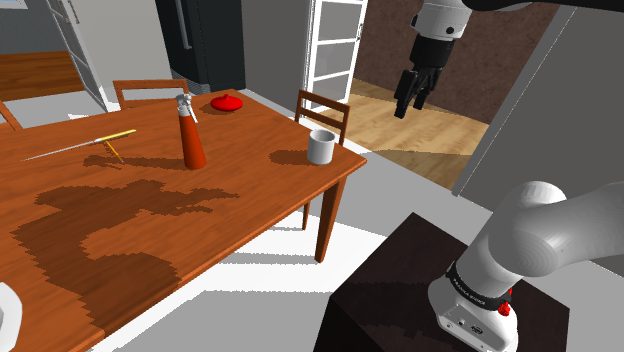} & \molmotile{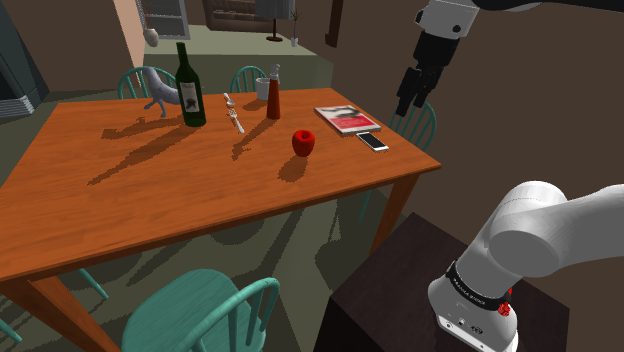} \\
    \molmotile{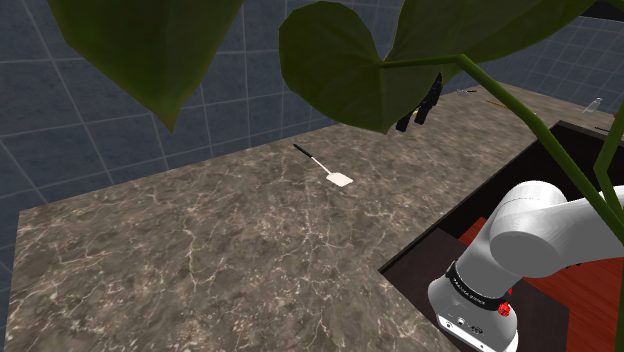} & \molmotile{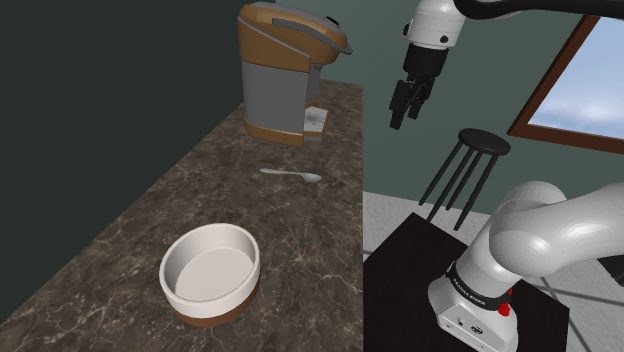} & \molmotile{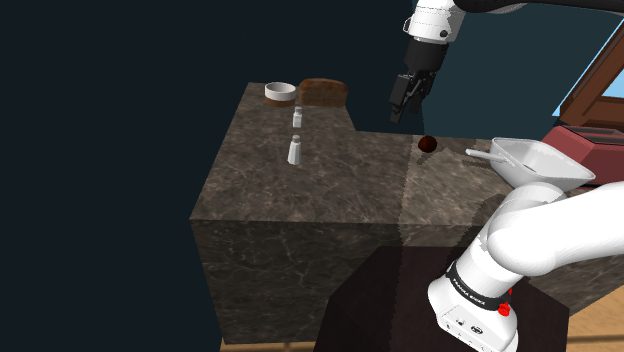} & \molmotile{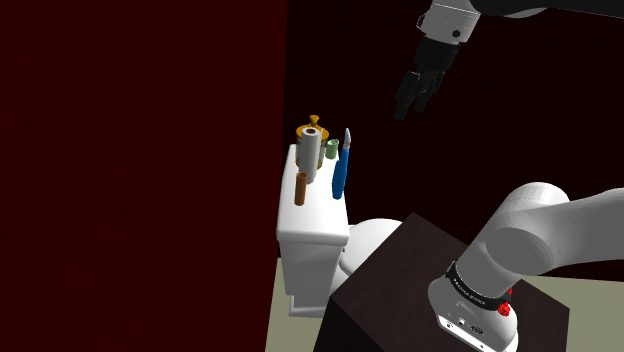} & \molmotile{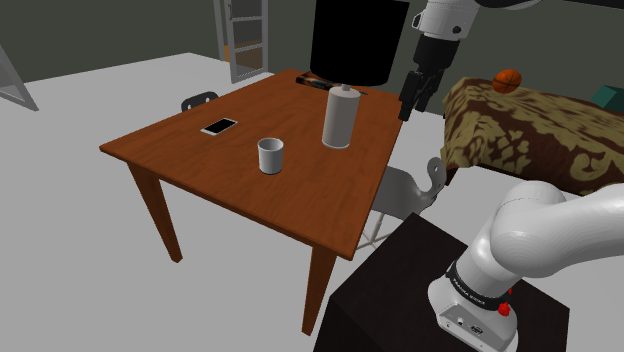} & \molmotile{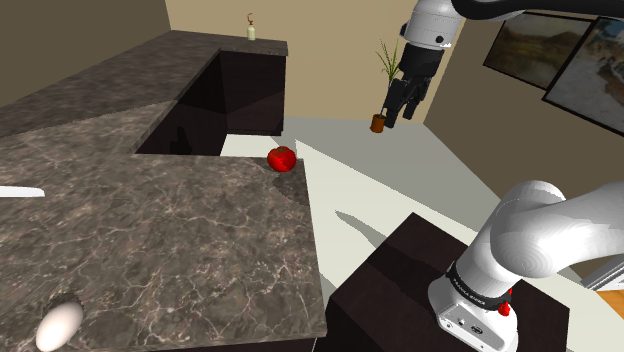} \\
  \end{tabular}

  \caption{\textbf{\texttt{MolmoSpaces} unseen scenes -- visualization.} Initial exterior-camera frame of every unseen scene task of the 144-task split, ordered as in Table~\ref{tab:molmo_splits}. Task IDs are shown in the lower-left corner of each frame.}
  \label{fig:molmo_visualization_unseen_scenes}
\end{figure}

\begin{figure}[tbp]
  \centering
  \setlength{\tabcolsep}{12pt}
  \renewcommand{\arraystretch}{0.5}
  \newcommand{\molmopic}{0.1\textwidth}
  \newcommand{\molmotile}[1]{\tikz[baseline=(img.south)]{%
    \node[inner sep=0pt, anchor=south west] (img) at (0,0) {\includegraphics[width=\molmopic]{figs/render/molmo_pics/molmo_#1.jpg}};%
    \node[anchor=south west, inner sep=0.8pt, fill=white, fill opacity=0.75, text opacity=1, font=\fontsize{5}{4}\selectfont] at (img.south west) {#1};}}
  \newcommand{\molmohead}[1]{\makebox[\molmopic][c]{\footnotesize{#1}}\rule[-2pt]{0pt}{0pt}}

  \begin{tabular}{@{}*{6}{c}@{}}
    \molmohead{atomizer} & \molmohead{bowl} & \molmohead{fork} & \molmohead{knife} & \molmohead{remotecontrol} & \molmohead{winebottle} \\
    [1pt]
    \molmotile{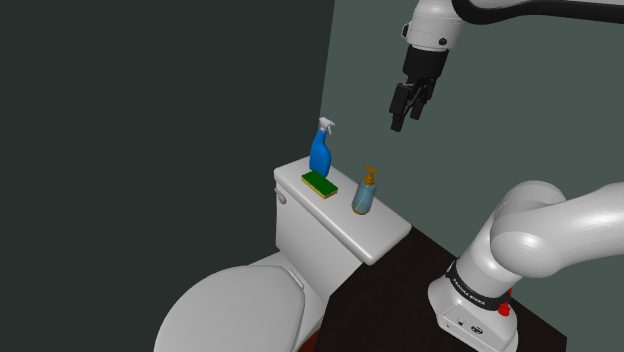} & \molmotile{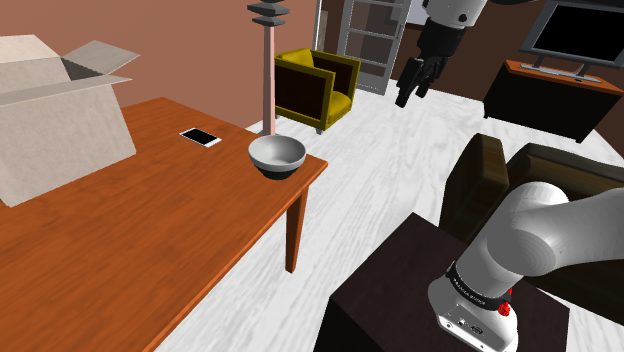} & \molmotile{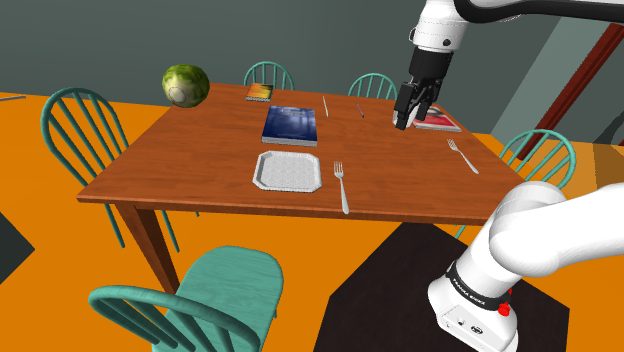} & \molmotile{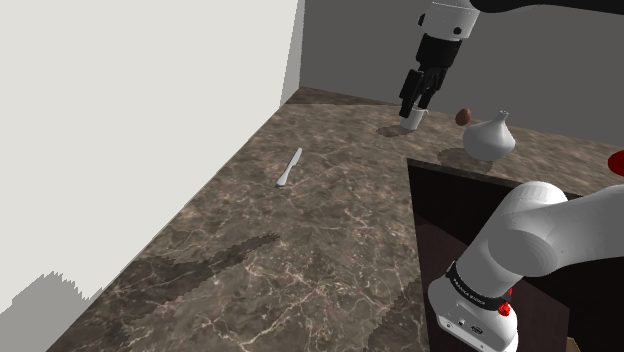} & \molmotile{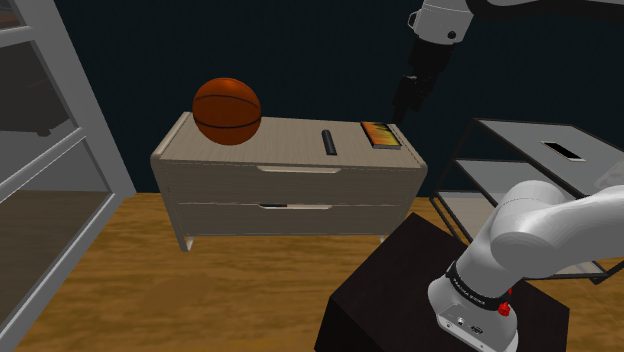} & \molmotile{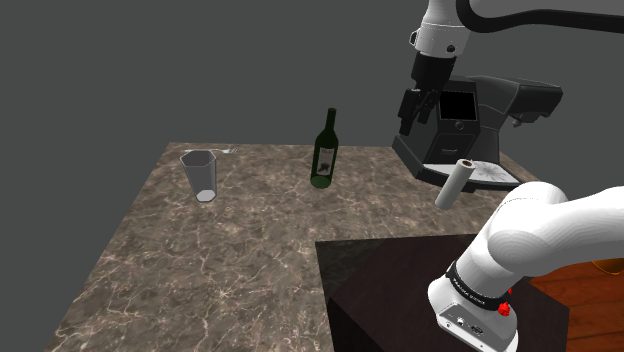} \\
    \molmotile{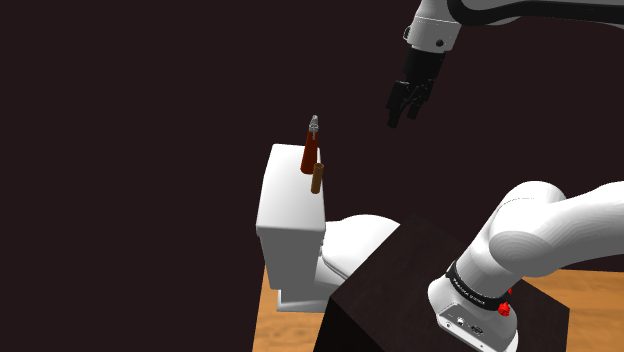} & \molmotile{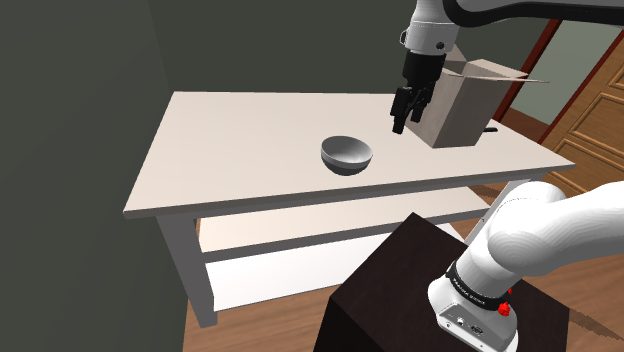} & \molmotile{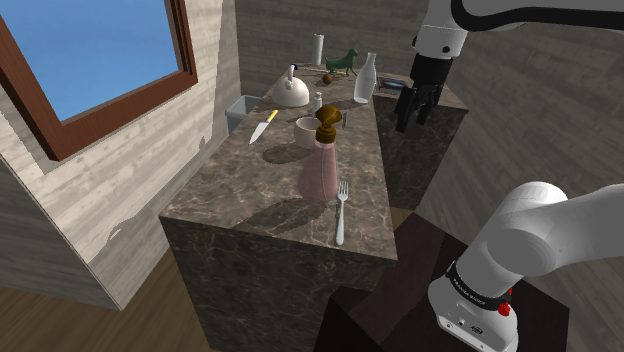} & \molmotile{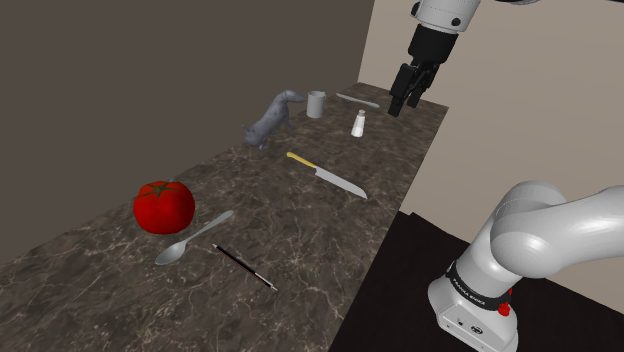} & \molmotile{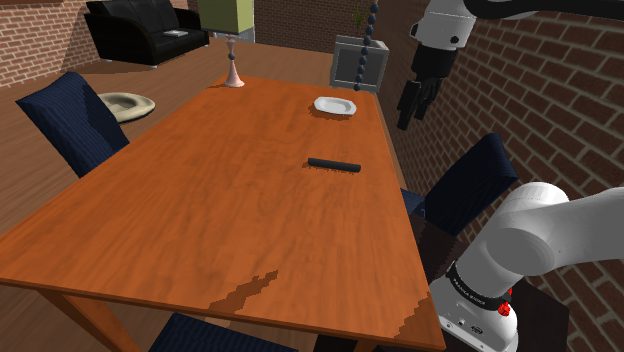} & \molmotile{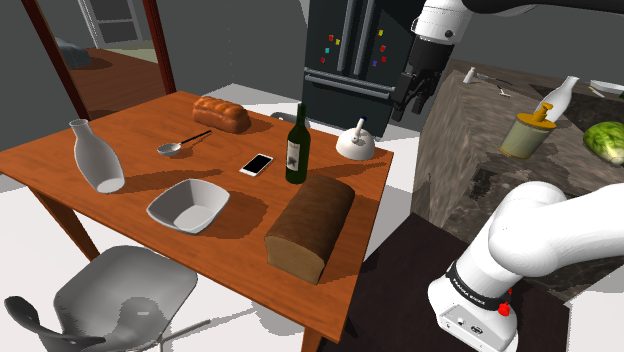} \\
    \molmotile{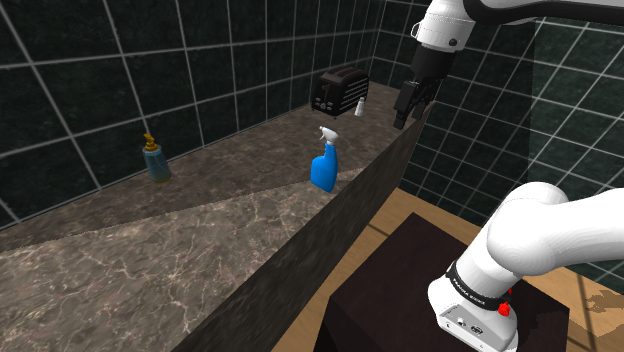} & \molmotile{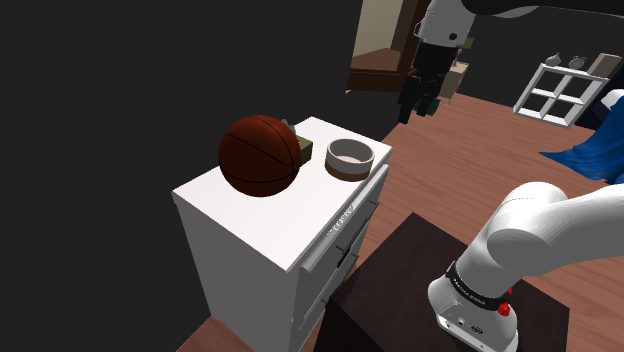} & \molmotile{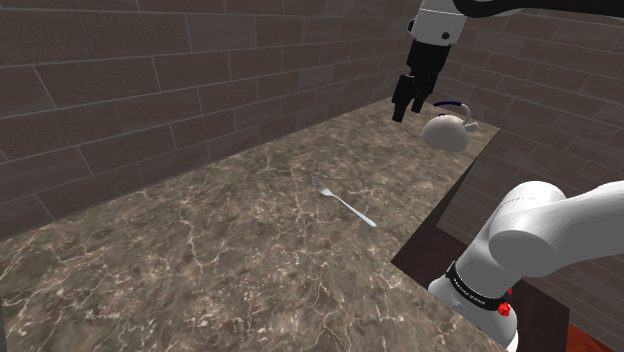} & \molmotile{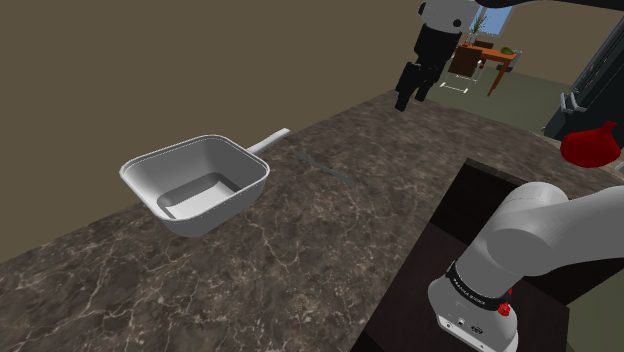} & \molmotile{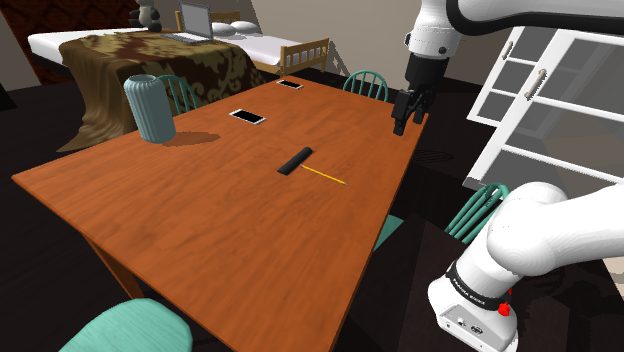} & \molmotile{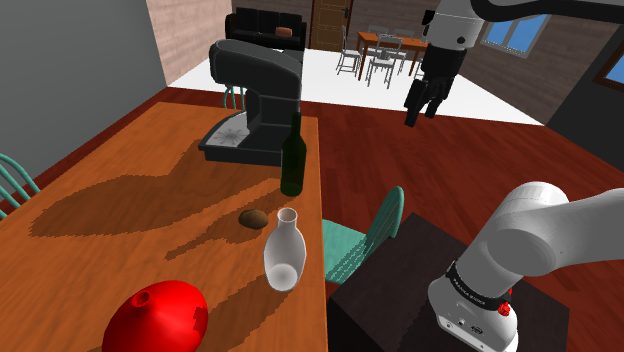} \\
    \molmotile{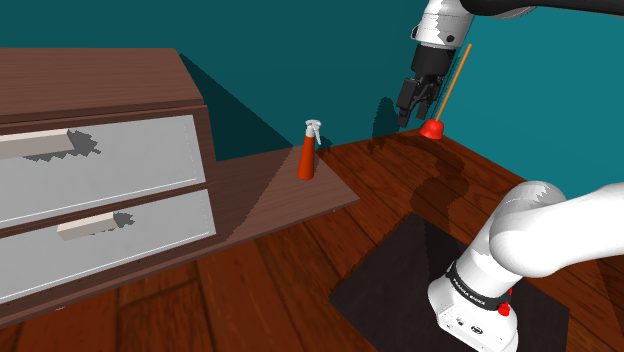} & \molmotile{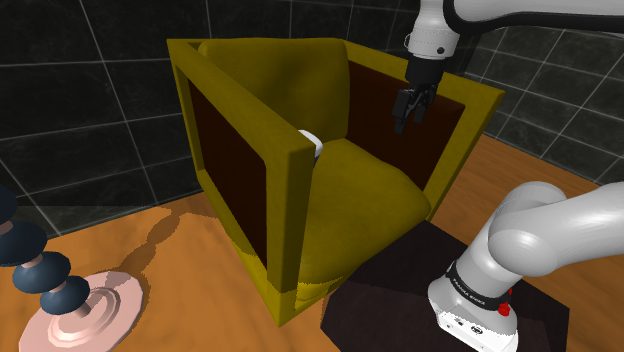} & \molmotile{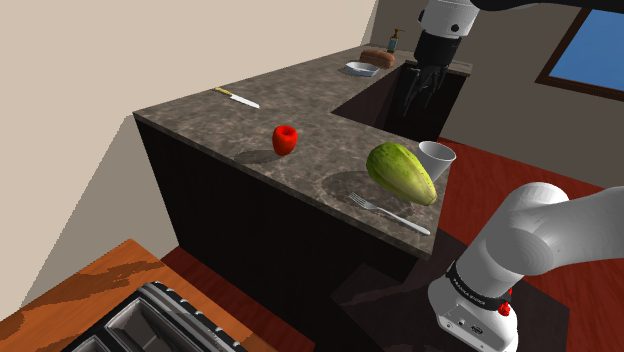} & \molmotile{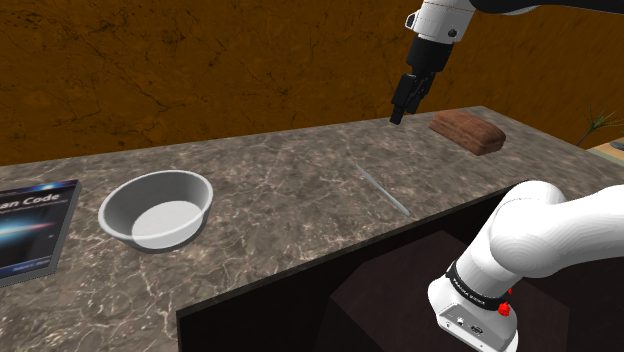} & \molmotile{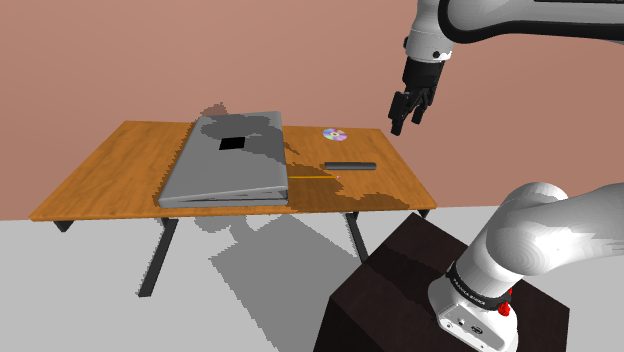} & \molmotile{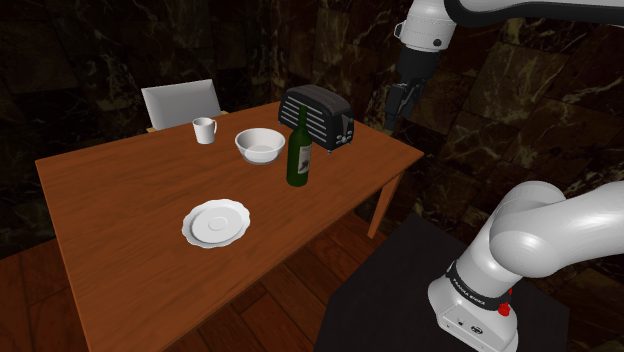} \\
    \molmotile{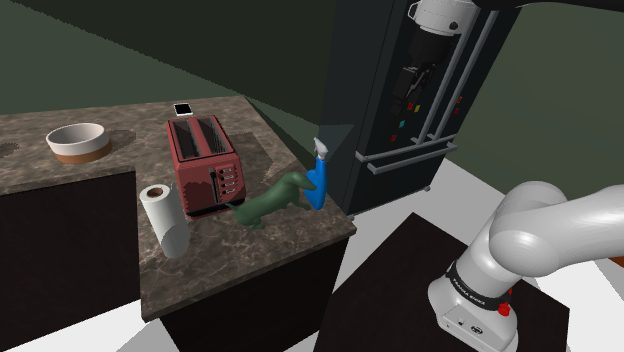} & \molmotile{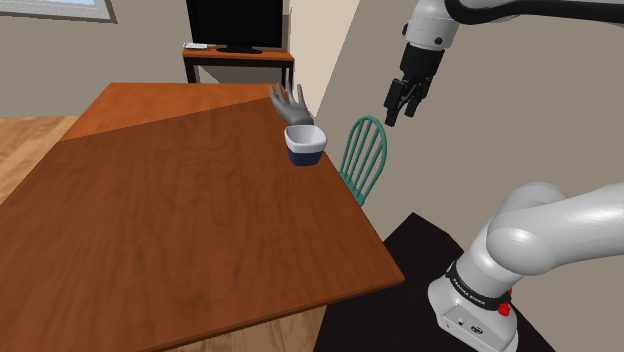} & \molmotile{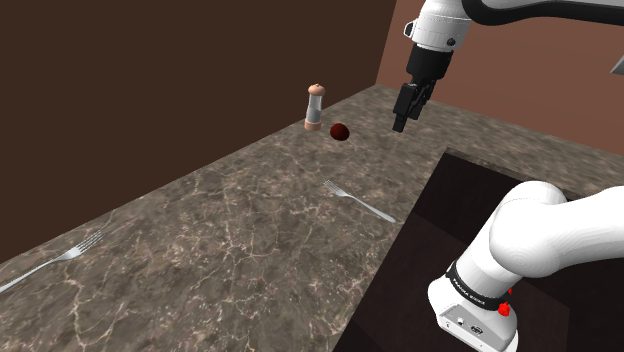} & \molmotile{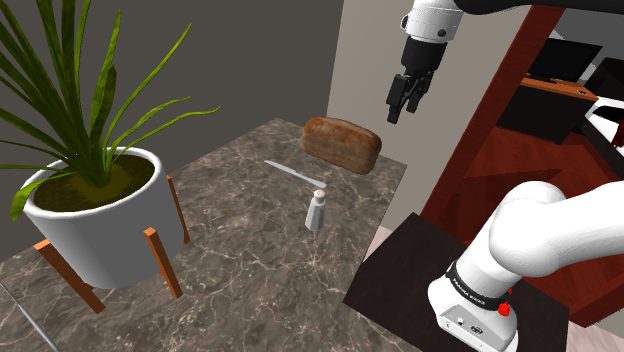} & \molmotile{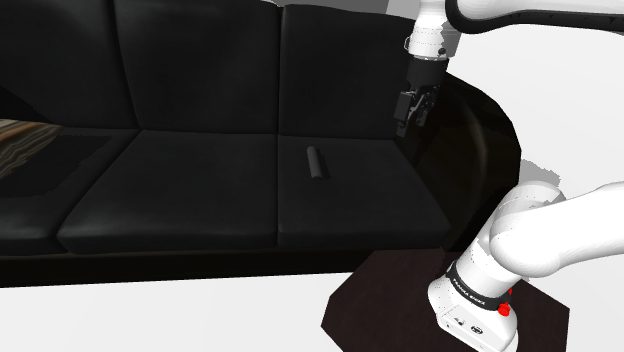} & \molmotile{226} \\
    \molmotile{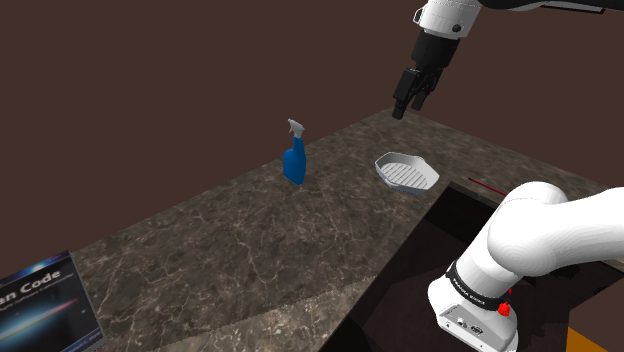} & \molmotile{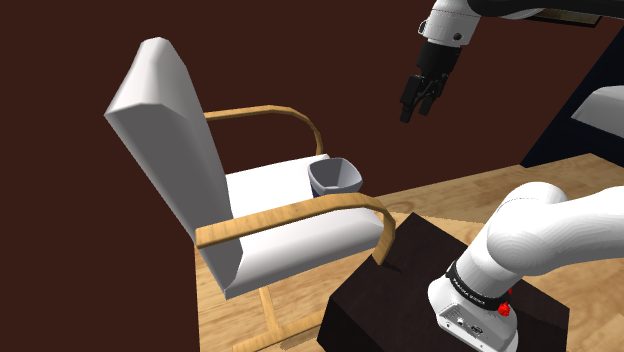} & \molmotile{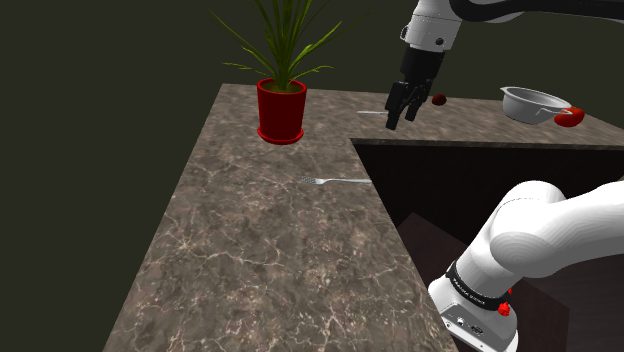} & \molmotile{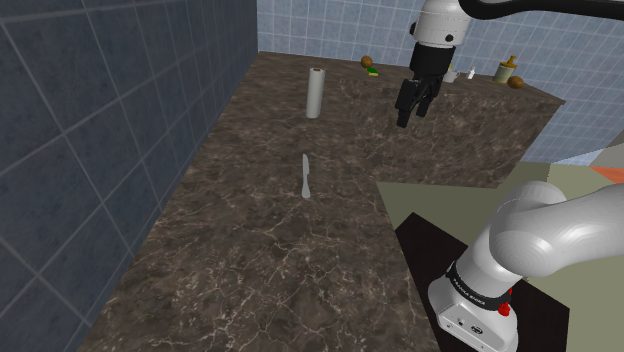} & \molmotile{357} & \molmotile{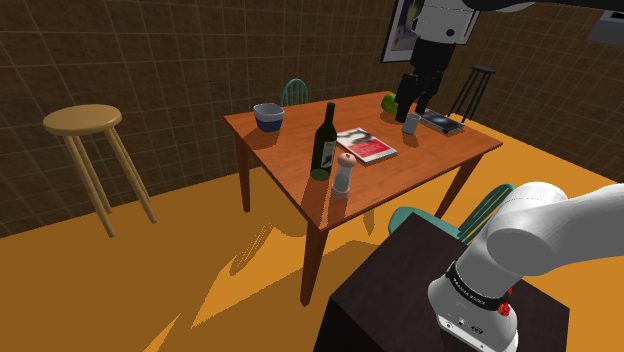} \\
    \molmotile{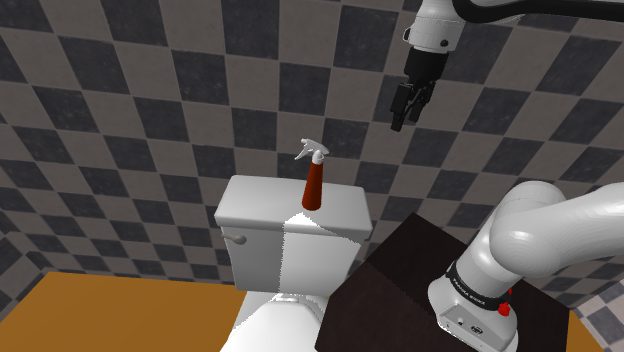} & \molmotile{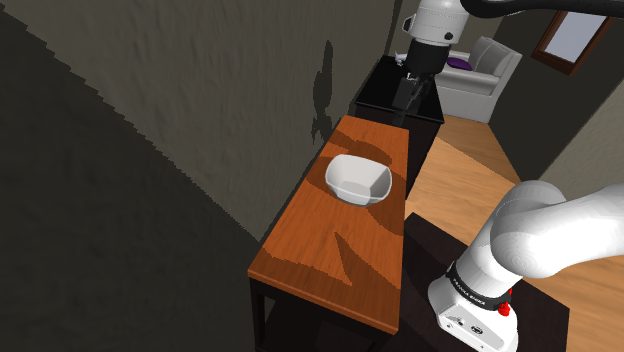} & \molmotile{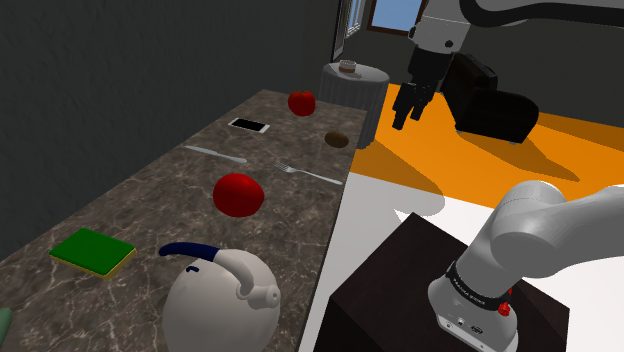} & \molmotile{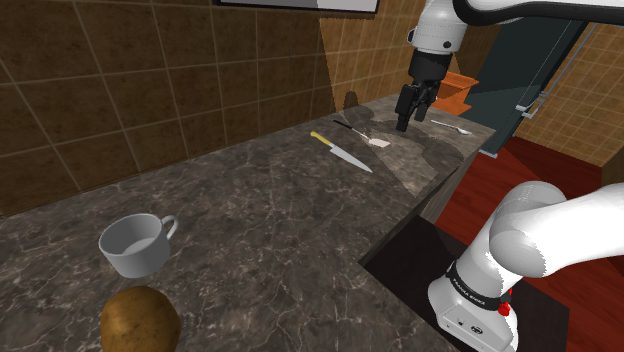} & \molmotile{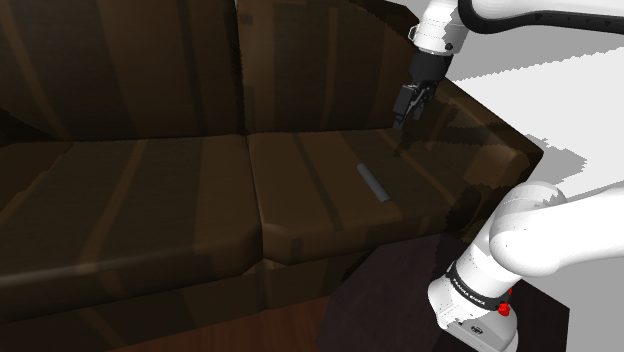} & \molmotile{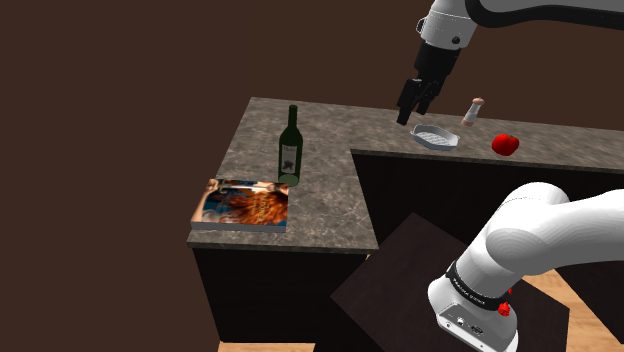} \\
    \molmotile{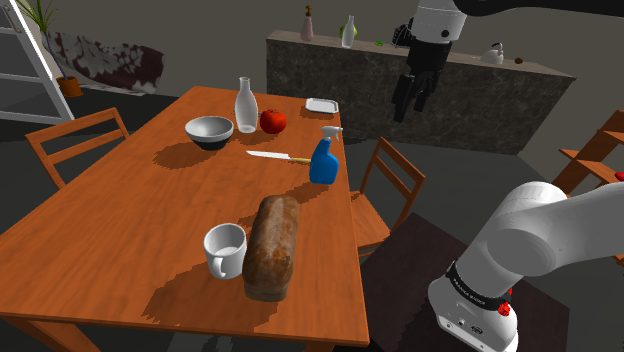} & \molmotile{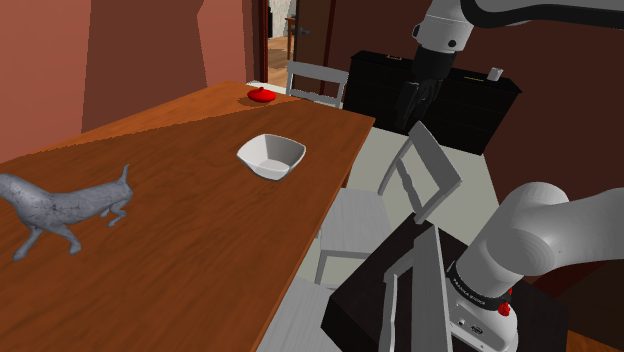} & \molmotile{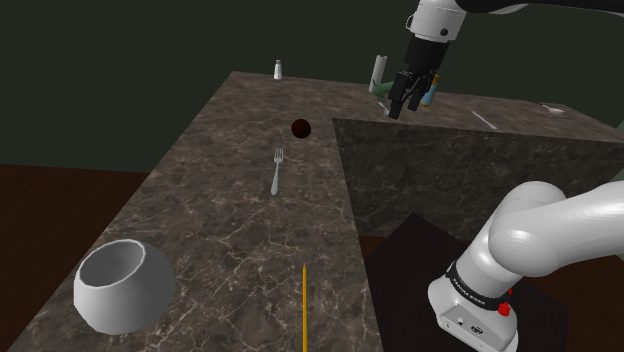} & \molmotile{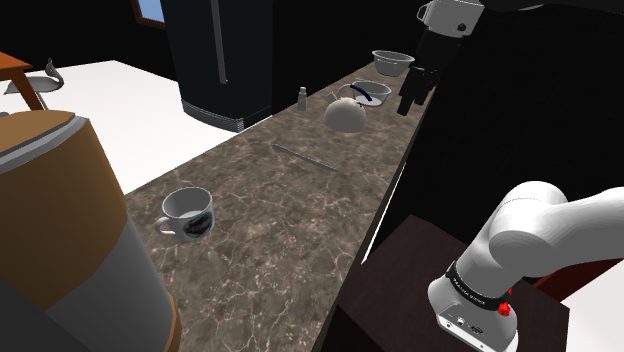} & \molmotile{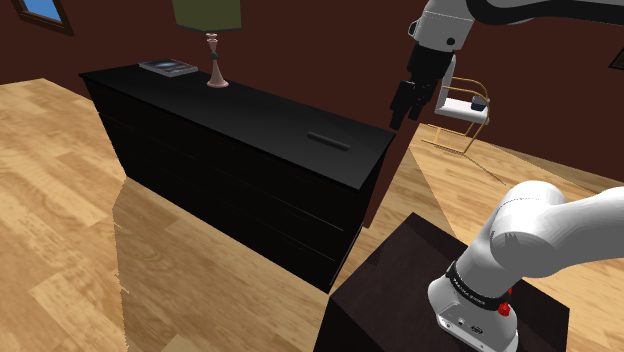} & \molmotile{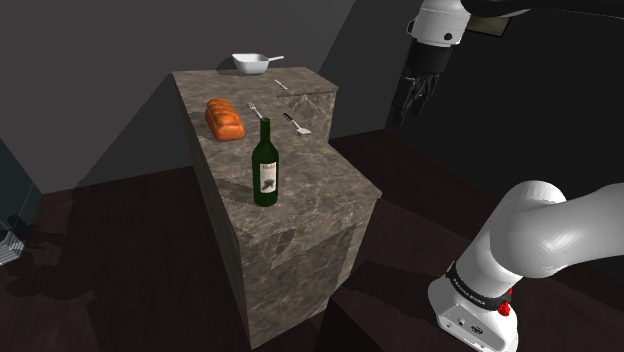} \\
  \end{tabular}

  \caption{\textbf{\texttt{MolmoSpaces} unseen objects -- visualization.} Initial exterior-camera frame of every unseen object task of the 144-task split, ordered as in Table~\ref{tab:molmo_splits}. Task IDs are shown in the lower-left corner of each frame.}
  \label{fig:molmo_visualization_unseen_objects}
\end{figure}

Studying multi-task training and generalization requires training splits of
increasing size but comparable difficulty, together with held-out tasks. For
\texttt{MolmoSpaces}, we first estimate the initial
success rate of each task by rolling out 48 episodes with the base policy. From
these we construct five training splits of 1, 4, 16, 64, and 144 tasks, each a
superset of the smaller ones, with a mean initial success rate near $30\%$. A
split of $k^2$ tasks contains $k$ objects with $k$ scenes each. We evaluate each training split on two held-out sets. The first contains
eight unseen scenes of every object the split was trained on, so each split
is tested only on new scenes of its own objects. The second, shared by all
splits, contains eight scenes each of six objects that appear in no training
split. Table~\ref{tab:molmo_splits} lists the resulting splits together with
their objects and the initial success rates, and Figures~\ref{fig:molmo_visualization_training},~\ref{fig:molmo_visualization_unseen_scenes}, and~\ref{fig:molmo_visualization_unseen_objects} visualize the tasks.

\newpage
We construct the \texttt{LIBERO} splits in the same way, estimating initial
success rates from 48 rollouts on 89 of the 90 tasks, and again running the
single-task split on four tasks. \texttt{LIBERO} is smaller than
\texttt{MolmoSpaces}, so its splits cannot be balanced as tightly: mean initial
success rate ranges from $23.4\%$ to $27.0\%$. A further 25 tasks are held out
to measure generalization. Table~\ref{tab:libero_splits} lists the splits and
their initial success rates.

\begin{table}[h]
  \centering
  \caption{\textbf{\texttt{LIBERO} task splits.} All tasks are drawn from
  \texttt{LIBERO-90}; task IDs are given without the \texttt{libero\_90\_}
  prefix. \emph{Init.\ SR} is the mean success rate of the base policy over the
  tasks of a split. The training splits are nested, and the unseen tasks are
  disjoint from all of them.}
  \label{tab:libero_splits}
  \vspace{2mm}
\begingroup
  \definecolor{mtblue}{HTML}{1D6996}
  \begin{tikzpicture}[
    x=1pt, y=1pt,
    font=\rmfamily,
    cell/.style={inner sep=0pt, anchor=base},
    lbl/.style={cell, anchor=base west},
    sub/.style={cell, anchor=base west, text width=280pt, align=left},
    head/.style={cell, text=mtblue},
    rule/.style={draw=mtblue, line width=0.9pt},
    midrule/.style={draw=mtblue!60, line width=0.5pt},
    sep/.style={draw=mtblue!25, line width=0.4pt}
  ]
  \def\W{397}

  \fill[mtblue!5] (0,-22) rectangle (\W,0);
  \draw[rule]    (0,0)   -- (\W,0);
  \draw[midrule] (0,-22) -- (\W,-22);

  \node[head, anchor=base west] at (3,-14.5) {Split};
  \node[head] at (305,-14.5) {Tasks};
  \node[head] at (368,-14.5) {Init. SR (\%)};

  \node[lbl]  at (3,-34.5) {\texttt{LIBERO} (1 task)};
  \node[cell] at (305,-34.5) {4};
  \node[cell] at (368,-34.5) {23.4};
  \node[sub] at (14,-47.5) {\emph{Task IDs}: 79, 31, 82, 38};
  \node[lbl]  at (3,-63.5) {\texttt{LIBERO} (4 tasks)};
  \node[cell] at (305,-63.5) {4};
  \node[cell] at (368,-63.5) {23.4};
  \node[sub] at (14,-76.5) {\emph{Task IDs}: 79, 31, 82, 38};
  \node[lbl]  at (3,-92.5) {\texttt{LIBERO} (16 tasks)};
  \node[cell] at (305,-92.5) {16};
  \node[cell] at (368,-92.5) {27.0};
  \node[sub] at (14,-105.5) {\emph{Task IDs}: 36, 49, 87, 86, 21, 85, 53, 64, 31, 79, 38, 82, 30, 47, 56, 24};
  \node[lbl]  at (3,-133.5) {\texttt{LIBERO} (64 tasks)};
  \node[cell] at (305,-133.5) {64};
  \node[cell] at (368,-133.5) {26.9};
  \node[sub] at (14,-146.5) {\emph{Task IDs}: 5, 32, 6, 36, 74, 39, 40, 76, 41, 43, 45, 49, 89, 65, 62, 42, 78, 8, 1, 88, 16, 17, 3, 87, 86, 58, 21, 84, 83, 25, 80, 27, 85, 73, 35, 53, 63, 14, 44, 64, 31, 79, 38, 82, 28, 70, 59, 60, 30, 12, 29, 47, 54, 56, 33, 9, 46, 50, 67, 51, 24, 20, 55, 61};
  \node[lbl]  at (3,-198.5) {Unseen tasks};
  \node[cell] at (305,-198.5) {25};
  \node[cell] at (368,-198.5) {26.9};
  \node[sub] at (14,-211.5) {\emph{Task IDs}: 34, 37, 4, 71, 66, 52, 75, 11, 18, 23, 81, 10, 69, 48, 7, 13, 0, 26, 77, 19, 72, 22, 2, 68, 57};

  \draw[sep] (0,-54.0) -- (\W,-54.0);
  \draw[sep] (0,-83.0) -- (\W,-83.0);
  \draw[sep] (0,-124.0) -- (\W,-124.0);
  \draw[sep] (0,-189.0) -- (\W,-189.0);
  \draw[rule] (0,-231.5) -- (\W,-231.5);
  \end{tikzpicture}
\endgroup
\end{table}

\newpage
\subsection{Implementation Details}
\label{app:implementation_methods}

\textbf{Common Setup. \;} All methods start from the same pretrained $\pi_{0.5}$
checkpoint, i.e., \texttt{pi05\_libero} on \texttt{LIBERO} and
\texttt{pi05\_droid\_jointpos} on \texttt{MolmoSpaces}, and share the same
data collection loop. As in the released checkpoint, the policy sees only camera images and the
language instruction; we verified that omitting proprioceptive state helps. Each iteration begins with a collection phase, in which
the policy collects 20 episodes per task into a replay buffer; training then
proceeds and differs by method. After the last iteration, a final evaluation
phase begins. Collection and evaluation both use an exponential moving average
of the policy parameters for stability. The policy outputs a chunk of 15
actions (10 actions on \texttt{LIBERO}) but executes only the first 5 before being queried again. We
nonetheless store one sample per environment step, pairing each observation with
the full action chunk executed from that step onward, stitched across consecutive
queries rather than taken from a single predicted chunk. The buffer is FIFO and sampled uniformly.

\textbf{FSFT. \;} 
FSFT stores only successful episodes and fine-tunes all policy parameters, i.e., VLM backbone, action expert, and vision encoder, with the standard $\pi_{0.5}$ flow-matching loss.

\textbf{BoN. \;} BoN keeps the VLA's parameters frozen and instead learns a
$Q$-function conditioned on observations and action chunks. At every query, it
samples $N{=}32$ candidate chunks with independent initial noise and executes
the one maximizing the minimum over the critic's ensemble, both during data
collection and at evaluation. The critic never sees raw images as it takes the
mean-pooled final-layer token embeddings of the policy's backbone, the
proprioceptive state, and the flattened action chunk as input. Since the policy is
frozen, this embedding is computed once at collection time and cached, which
makes critic training substantially faster. We use a time-to-success reward
($-1$ per step, $0$ on success), with episodes terminating only on success and
time limits treated as truncation. We train $Q$ and $V$ on all collected
episodes with $H$-step TD targets from slowly updated
target networks: the $H$ discounted rewards accumulated over the chunk, plus
$\gamma^{H}$ times the bootstrapped value of the state reached at its end.
Because $V$ regresses $Q$ at buffer actions rather than at a maximum or an upper
expectile, this is a SARSA-style evaluation of the data-collecting policy.

\textbf{OGPO. \;} Our implementation of OGPO mostly follows the design choices specified in the original paper (\citep{patil2026ogpo}). However, we make some additional design choices and modifications to adapt it to fine-tune $\pi_{0.5}$. Specifically, we make the following design choices for our runs:
\vspace{-4mm}
\begin{itemize}
    \item \textbf{Critic:} Our implementation parametrizes the state-action value function using a BroNet residual MLP and uses an ensemble of 10 critics \citep{patil2026ogpo}. The critics are aggregated by randomly sub-sampling two critics and then picking the minimum among the two. 
    \item\textbf{Critic Update:} The critic is learned by regressing on a TD target with Q-variance reduction expressed in Equation \ref{eq:ogpo_td_qvr}. At every training step, this critic receives two gradient updates: one on a batch of transitions sampled from the replay buffer, and one on a batch sampled only from successful episodes. This \textit{success oversampling} helps amplify the sparse reward signal. 
    \item \textbf{Policy Updates:} Policy updates are performed using the group relative advantages as per Equation \ref{eq:ogpo_group_advantage} and Equation \ref{eq:ogpo_policy_update}. Additionally, policy updates also use conservative advantages and BC loss on transitions from successful episodes for stability. 
    \item \textbf{BoN:} During collection rounds, we generate a set of $N=32$ candidate chunks using SDE sampling and execute the one with the highest Q-value. We aggregate the Q-function for this using the same aggregation strategy described above. \chg{Evaluation uses $N=1$.}
    \item \textbf{Training:} In experiments with OGPO, the backbone is frozen and only the action expert is trained. 
\end{itemize}
 The other design choices and exact hyperparameters used in our implementation are specified in Table~\ref{tab:hyperparameters}.

 \textbf{AWR. \;} The training pipeline of AWR is mostly the same pipeline as OGPO. The sole difference is in the policy training objective. AWR uses a weighted supervised loss (see Equation~\ref{eq:weighted_supervised}). The AWR weights are computed using the group relative advantages computed from the critic and normalize the advantages by dividing by the group's standard deviation. The BC loss on successful transitions from OGPO is retained with the same weight.

\textbf{FSFT+BoN. \;} FSFT+BoN collects data with BoN and additionally fine-tunes the expert head of the policy on the
successful transitions of the shared buffer. The critic training pipeline and collection strategy are the same as OGPO.

\newpage
\subsection{Hyperparameters}
\label{app:hyperparameters}

\begin{table}[h]
  \centering
  \caption{\textbf{Hyperparameters.} \emph{Training budget}: each run performs
  $M$ collection rounds of $K$ episodes per task, with the given number of
  gradient steps per round (single-task / multi-task).
  \emph{Tuned weights}: whether the full VLA or only the action expert is
  fine-tuned. \emph{Trained on}: whether the SFT loss uses only successful
  episodes. \emph{Update interval}: critic steps per policy update.
  \emph{Bootstrap}: SARSA uses the next action stored in the buffer, TD uses
  an action sampled from the current policy.
  Critic-based methods use a replay buffer of 1M transitions (2M for OGPO on the 144-task split). BoN and OGPO run for 10 collection rounds, except on the 144-task split, where we
extend it to 15 rounds because performance had not yet plateaued. FSFT, in contrast, plateaus early and then slightly declines, so 10 rounds was enough.}
  \label{tab:hyperparameters}
  \vspace{2mm}
\begingroup
  \definecolor{cFSFT}{HTML}{1D6996}
  \definecolor{cBoN}{HTML}{E17C05}
  \definecolor{cFSFTBoN}{HTML}{5F4690}
  \definecolor{cAWR}{HTML}{38A6A5}
  \definecolor{cOGPO}{HTML}{0F8554}
  \begin{tikzpicture}[
    font=\rmfamily,
    body/.style={inner xsep=0pt, inner ysep=2pt, outer sep=0pt, text height=7pt,
      text depth=2.5pt},
    lbl/.style={body, anchor=base west, text width=140pt, align=left},
    val/.style={body, anchor=base, minimum width=43pt},
    hd/.style={inner xsep=0pt, inner ysep=2pt, outer sep=0pt, anchor=base,
      minimum width=43pt, text height=20pt, text depth=9pt},
    group/.style={font=\rmfamily\itshape, text=black!55},
    rule/.style={draw=black!80, line width=0.9pt},
    midrule/.style={draw=black!35, line width=0.5pt}
  ]
  \matrix (m) [
    matrix of nodes,
    nodes in empty cells,
    inner sep=0pt, outer sep=0pt,
    column sep=0pt, row sep=0pt,
    column 1/.style={nodes={lbl}},
    column 2/.style={nodes={val}},
    column 3/.style={nodes={val}},
    column 4/.style={nodes={val}},
    column 5/.style={nodes={val}},
    column 6/.style={nodes={val}},
    row 1/.style={nodes={hd}},
    row 2/.style={nodes={group}},
    row 6/.style={nodes={group}},
    row 13/.style={nodes={group}},
    row 21/.style={nodes={group}},
    row 23/.style={nodes={group}},
    row 25/.style={nodes={group}},
  ] {
    |[lbl]|
      & |[text=cFSFT]| FSFT
      & |[text=cBoN]| BoN
      & |[text=cOGPO]| OGPO
      & |[text=cAWR]| AWR
      & |[text=cFSFTBoN]| \shortstack{FSFT\\+BoN} \\
    Training budget          & & & & & \\
    Collection rounds $M$    & 10 & 10/15 & 10/15 & 10 & 10 \\
    Episodes per task $K$    & 20 & 20 & 20 & 20 & 20 \\
    Steps per round          & 100/500 & 10k/50k & 1k & 1k & 1k \\
    Policy                   & & & & & \\
    Learning rate            & $2.5{\cdot}10^{-5}$ & -- & $2.5{\cdot}10^{-5}$ & $2.5{\cdot}10^{-5}$ & $2.5{\cdot}10^{-5}$ \\
    Batch size               & 256 & -- & 512 & 512 & 256 \\
    EMA decay                & 0.999 & -- & 0.995 & 0.995 & 0.995 \\
    Tuned weights            & full & -- & expert & expert & expert \\
    Trained on               & successes & -- & all & all & successes \\
    Update interval          & -- & -- & 1 & 1 & 1 \\
    Critic                   & & & & & \\
    Learning rate            & -- & $10^{-4}$ & $10^{-4}$ & $10^{-4}$ & $10^{-4}$ \\
    Architecture             & -- & BroNet & BroNet & BroNet & BroNet \\
    Hidden width             & -- & 1024 & 1024 & 1024 & 1024 \\
    Batch size               & -- & 1024 & 512 & 512 & 512 \\
    EMA decay                & -- & 0.995 & 0.95 & 0.95 & 0.95 \\
    Bootstrap                & -- & SARSA & TD & TD & TD \\
    Discount $\gamma$        & -- & 0.995 & 0.995 & 0.995 & 0.995 \\
    Replay buffer            & & & & & \\
    Capacity                 & 500k & 1M & 1M/2M & 1M & 1M  \\
    Action selection         & & & & & \\
    Candidates $N$           & -- & 32 & 32 & 32 & 32 \\
    Method-specific          & & & & & \\
    AWR temperature $\beta$  & -- & -- & -- & 1.0 & -- \\
    AWR Weight Clip          & -- & -- & -- & 20 & -- \\
    AWR/OGPO Group Size          & -- & -- & 32 & 32 & -- \\
    AWR/OGPO BC Loss Weight      & -- & -- & 1.0 & 1.0 & -- \\
    Variance Reduction $N_{vr}$  & -- & -- & 16 & 16 & 16 \\
    OGPO PPO Clip Ratio      & -- & -- & 0.01 & -- & -- \\
  };

  \path (m-1-2.base) ++(0,-5.5pt) coordinate (hdbar);
  \path (m-1-2.base) ++(0,-8.5pt) coordinate (hdmid);
  \draw[rule]    (m.north west) -- (m.north east);
  \draw[midrule] (hdmid -| m.west) -- (hdmid -| m.east);
  \draw[rule]    (m.south west) -- (m.south east);
  \foreach \r in {6,13,21,23,25}
    \draw[draw=black!15, line width=0.4pt]
      (m-\r-1.north -| m.west) -- (m-\r-1.north -| m.east);

  \foreach \c/\col in {2/cFSFT,3/cBoN,4/cOGPO,5/cAWR,6/cFSFTBoN}
    \draw[draw=\col, line width=1.8pt]
      ([xshift=-17pt]m-1-\c.base |- hdbar) -- ([xshift=17pt]m-1-\c.base |- hdbar);
  \end{tikzpicture}
\endgroup
\end{table}

\newpage
\subsection{Evaluation Protocol}
\label{app:evaluation}

In this section we describe the evaluation protocol. On both \texttt{LIBERO} and \texttt{MolmoSpaces} we evaluate each policy in distribution and on unseen tasks. For \texttt{LIBERO}, the unseen set comprises 25 tasks; for \texttt{MolmoSpaces}, eight unseen scenes of objects from the 144-task split and six unseen objects (see Appendix~\ref{app:task_splits}). Our metric throughout is success rate.

During training, the in-distribution success rate is measured on the 20 episodes the policy collects per task in each round, so no extra mid-evaluation rollouts are needed. After the last training phase, we run a separate evaluation phase of 32 episodes per task, both in distribution and on unseen tasks. We repeat every experiment with three seeds and report the mean success rate over seeds, together with 95\% stratified bootstrap confidence intervals computed with \texttt{rliable}~\citep{agarwal2021deep}.

\subsection{Computational Resources}
\label{app:compute}
Our experiments ran on nodes with four NVIDIA GH200 GPUs each. A single-task run took about 2-5 hours of wall-clock time depending on the task, and runs on the largest multi-task splits took 200 hours.

\newpage
\section{Training curves}
\label{app:training_curves}

We report success-rate curves over training for the experiments in Figures~\ref{fig:main}, ~\ref{fig:main_libero} and ~\ref{fig:policy_extraction}.

\subsection{Generalization Per Task - \texttt{MolmoSpaces}}
\label{app:generalization_per_task}

\subsubsection{\texttt{MolmoSpaces} Unseen Scenes}
\begin{figure}[h]
    \centering
    \includegraphics[width=\linewidth]{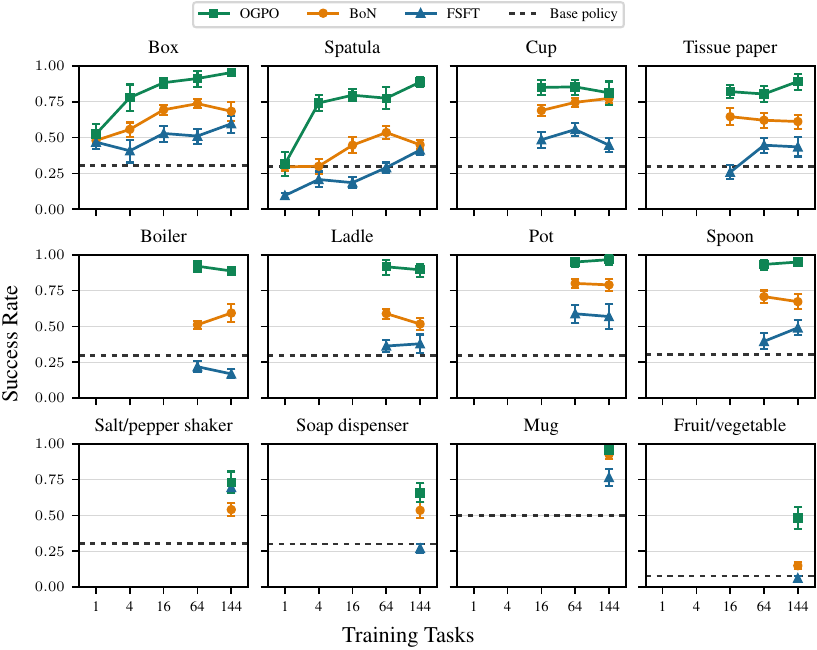}
    \caption{\texttt{MolmoSpaces}. Final success
      rate on unseen scenes of the training objects. 
      Points show the mean over three seeds with 95\% stratified bootstrap CIs.}
    \label{fig:generalization_scenes}
\end{figure}

\newpage

\subsection{Main Results — \texttt{MolmoSpaces}}
\subsubsection{\texttt{MolmoSpaces} (1 Task) In-Distribution}
\begin{figure}[h]
    \centering
    \includegraphics[width=0.54\linewidth]{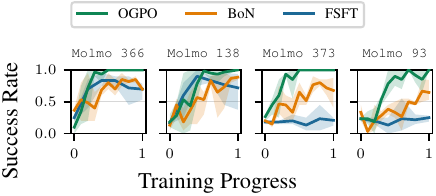}
    \caption{\textbf{\texttt{MolmoSpaces} (1 task): training tasks.} Success rate on
    the training tasks as a function of normalized training progress, where $1$
    denotes the end of post-training. Lines show the mean over three seeds. Shaded
    regions denote 95\% stratified bootstrap CIs.}
    \label{fig:curves_molmo_1}
\end{figure}

\subsubsection{\texttt{MolmoSpaces} (1 Task) Unseen Scenes}
\begin{figure}[h]
    \centering
    \includegraphics[width=\linewidth]{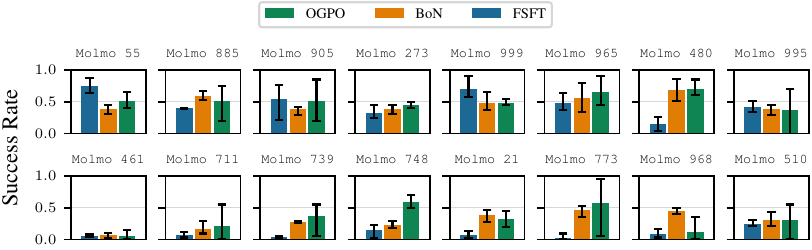}
    \caption{\textbf{\texttt{MolmoSpaces} (1 task): unseen scenes.} Final success
    rate on unseen scenes. Bars show the mean over three seeds with 95\% stratified
    bootstrap CIs.}
    \label{fig:eval_molmo_1_scenes}
\end{figure}

\newpage
\subsubsection{\texttt{MolmoSpaces} (1 Task) Unseen Objects}
\begin{figure}[h]
    \centering
    \includegraphics[width=\linewidth]{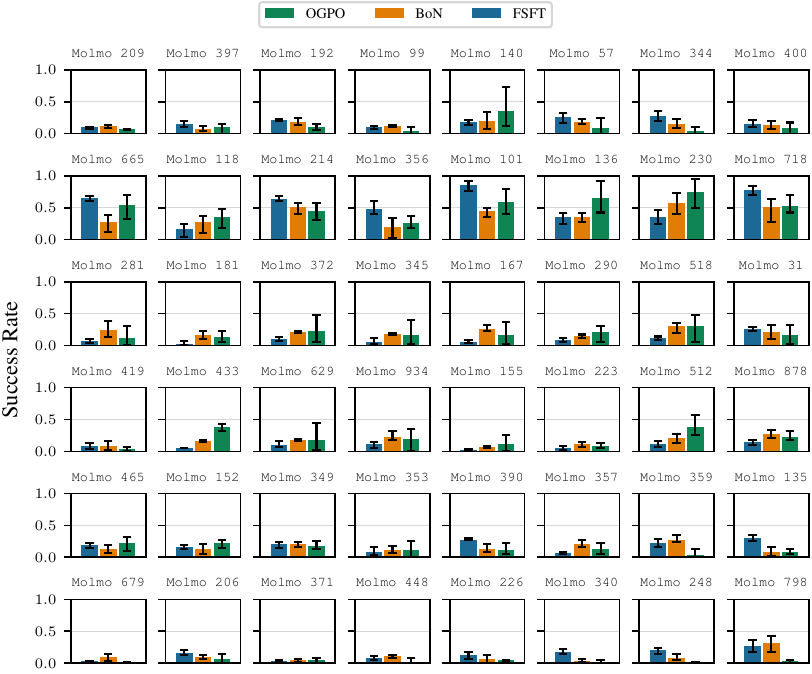}
    \caption{\textbf{\texttt{MolmoSpaces} (1 task): unseen objects.} Final success
    rate on unseen objects. Bars show the mean over three seeds with 95\% stratified
    bootstrap CIs.}
    \label{fig:eval_molmo_1_objects}
\end{figure}

\newpage
\subsubsection{\texttt{MolmoSpaces} (4 Tasks) In-Distribution}
\begin{figure}[h]
    \centering
    \includegraphics[width=0.54\linewidth]{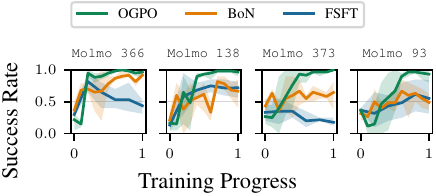}
    \caption{\textbf{\texttt{MolmoSpaces} (4 tasks): training tasks.} Success rate on
    the training tasks as a function of normalized training progress, where $1$
    denotes the end of post-training. Lines show the mean over three seeds. Shaded
    regions denote 95\% stratified bootstrap CIs.}
    \label{fig:curves_molmo_4}
\end{figure}

\subsubsection{\texttt{MolmoSpaces} (4 Tasks) Unseen Scenes}
\begin{figure}[h]
    \centering
    \includegraphics[width=\linewidth]{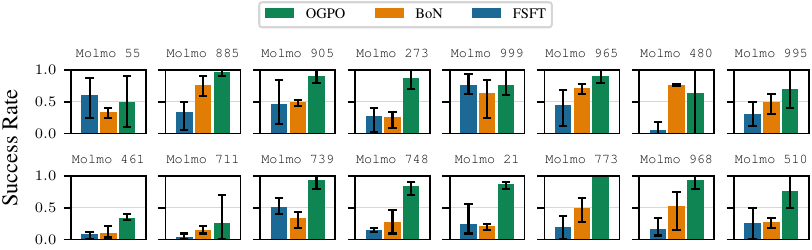}
    \caption{\textbf{\texttt{MolmoSpaces} (4 tasks): unseen scenes.} Final success
    rate on unseen scenes. Bars show the mean over three seeds with 95\% stratified
    bootstrap CIs.}
    \label{fig:eval_molmo_4_scenes}
\end{figure}

\newpage
\subsubsection{\texttt{MolmoSpaces} (4 Tasks) Unseen Objects}
\begin{figure}[h]
    \centering
    \includegraphics[width=\linewidth]{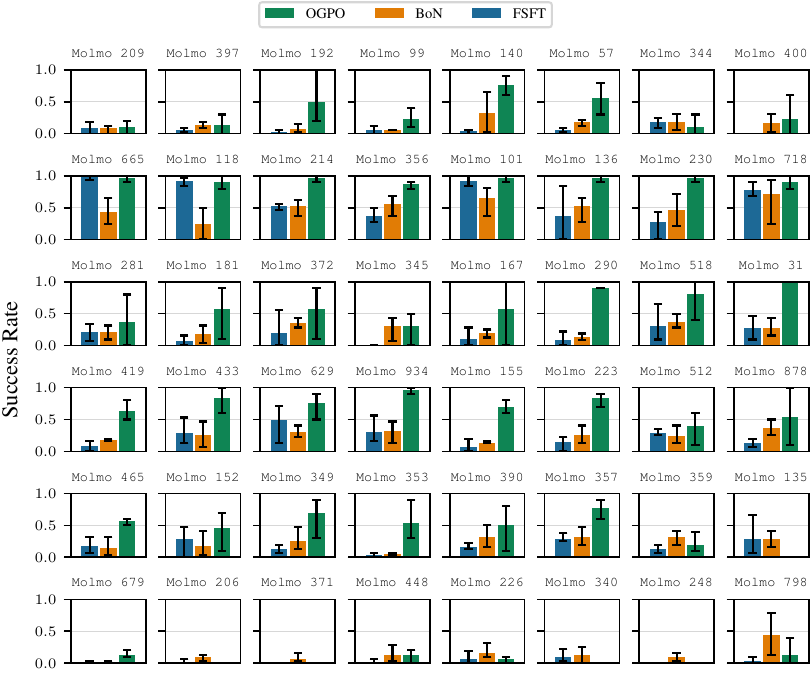}
    \caption{\textbf{\texttt{MolmoSpaces} (4 tasks): unseen objects.} Final success
    rate on unseen objects. Bars show the mean over three seeds with 95\% stratified
    bootstrap CIs.}
    \label{fig:eval_molmo_4_objects}
\end{figure}

\newpage
\subsubsection{\texttt{MolmoSpaces} (16 Tasks) In-Distribution}
\begin{figure}[h]
    \centering
    \includegraphics[width=\linewidth]{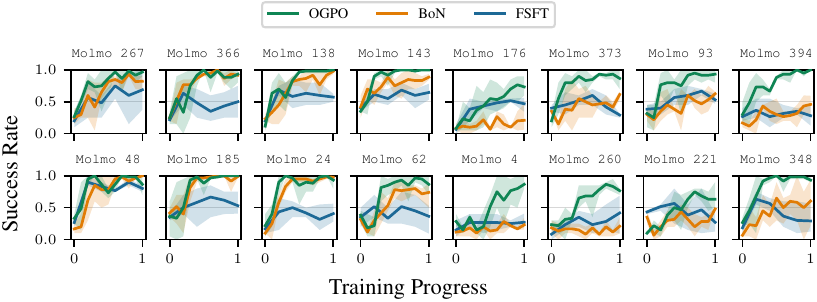}
    \caption{\textbf{\texttt{MolmoSpaces} (16 tasks): training tasks.} Success rate on
    the training tasks as a function of normalized training progress, where $1$
    denotes the end of post-training. Lines show the mean over three seeds. Shaded
    regions denote 95\% stratified bootstrap CIs.}
    \label{fig:curves_molmo_16}
\end{figure}

\subsubsection{\texttt{MolmoSpaces} (16 Tasks) Unseen Scenes}
\begin{figure}[h]
    \centering
    \includegraphics[width=\linewidth]{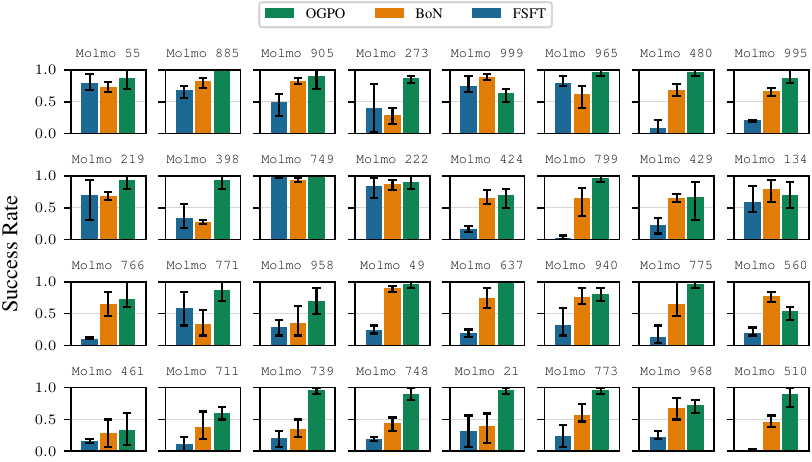}
    \caption{\textbf{\texttt{MolmoSpaces} (16 tasks): unseen scenes.} Final success
    rate on unseen scenes. Bars show the mean over three seeds with 95\% stratified
    bootstrap CIs.}
    \label{fig:eval_molmo_16_scenes}
\end{figure}

\newpage
\subsubsection{\texttt{MolmoSpaces} (16 Tasks) Unseen Objects}
\begin{figure}[h]
    \centering
    \includegraphics[width=\linewidth]{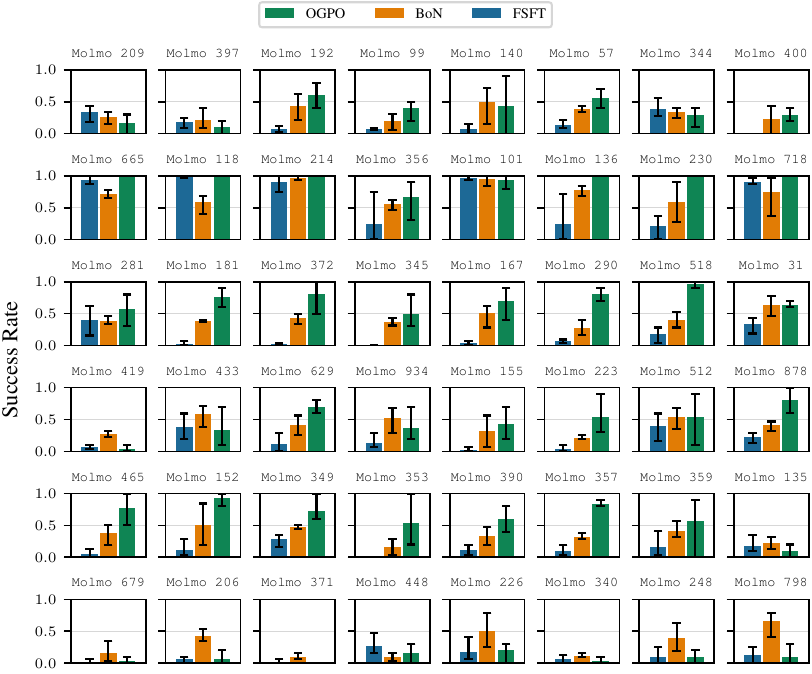}
    \caption{\textbf{\texttt{MolmoSpaces} (16 tasks): unseen objects.} Final success
    rate on unseen objects. Bars show the mean over three seeds with 95\% stratified
    bootstrap CIs.}
    \label{fig:eval_molmo_16_objects}
\end{figure}

\newpage
\subsubsection{\texttt{MolmoSpaces} (64 Tasks) In-Distribution}
\begin{figure}[h]
    \centering
    \includegraphics[width=1.0\linewidth]{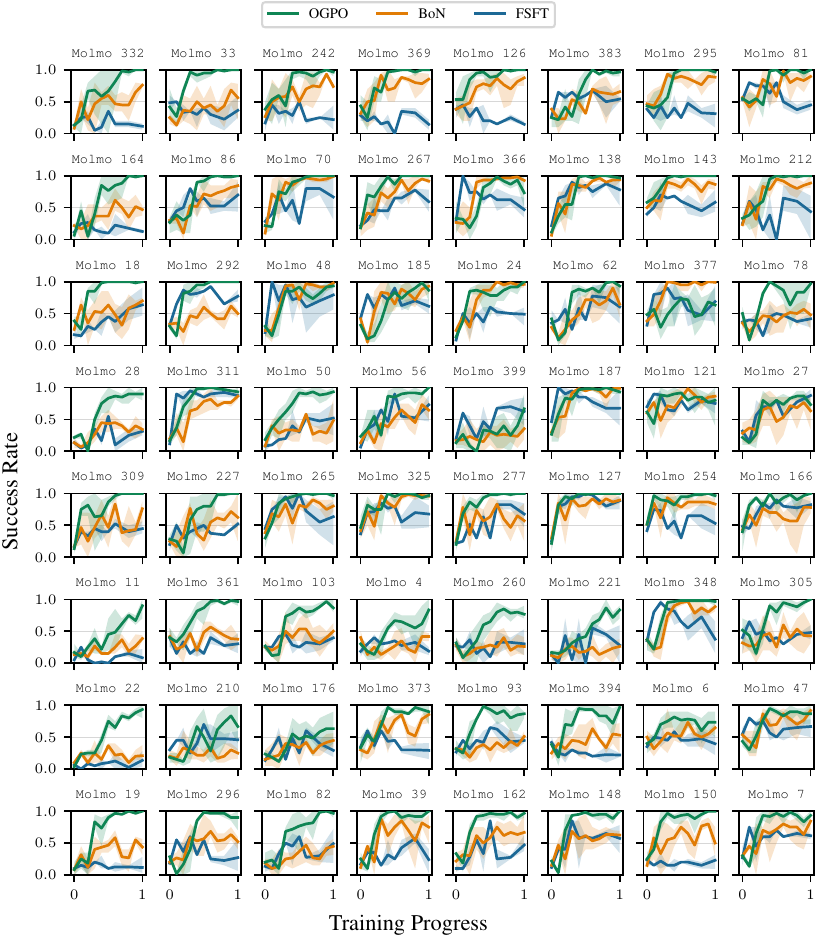}
    \caption{\textbf{\texttt{MolmoSpaces} (64 tasks): training tasks.} Success rate on
    the training tasks as a function of normalized training progress, where $1$
    denotes the end of post-training. Lines show the mean over three seeds. Shaded
    regions denote 95\% stratified bootstrap CIs.}
    \label{fig:curves_molmo_64}
\end{figure}

\newpage
\subsubsection{\texttt{MolmoSpaces} (64 Tasks) Unseen Scenes}
\begin{figure}[h]
    \centering
    \includegraphics[width=\linewidth]{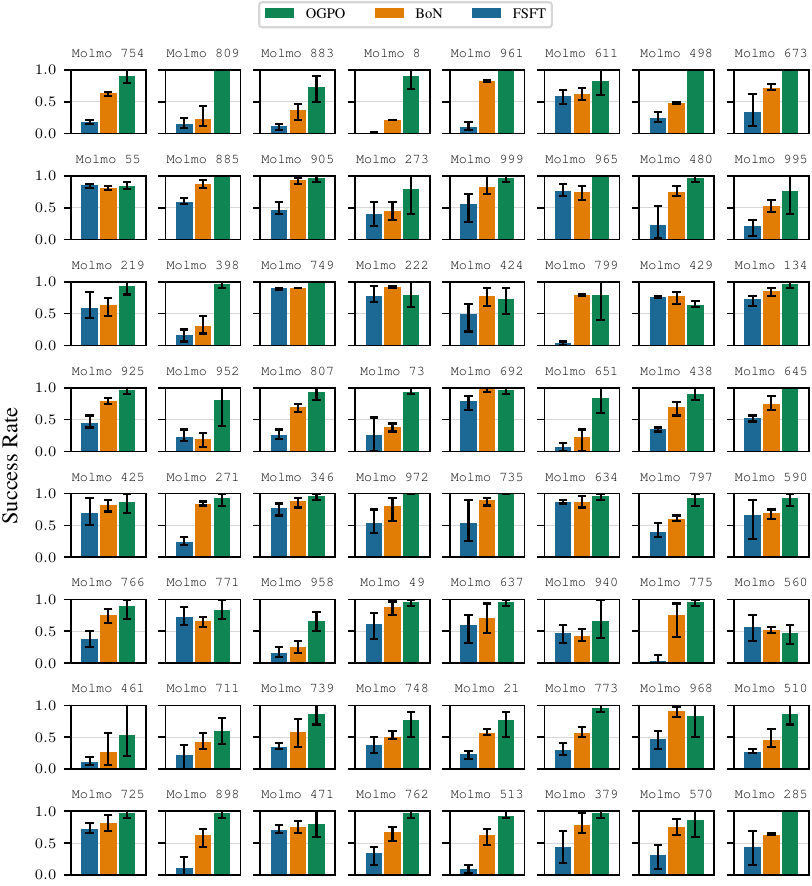}
    \caption{\textbf{\texttt{MolmoSpaces} (64 tasks): unseen scenes.} Final success
    rate on unseen scenes. Bars show the mean over three seeds with 95\% stratified
    bootstrap CIs.}
    \label{fig:eval_molmo_64_scenes}
\end{figure}

\newpage
\subsubsection{\texttt{MolmoSpaces} (64 Tasks) Unseen Objects}
\begin{figure}[h]
    \centering
    \includegraphics[width=\linewidth]{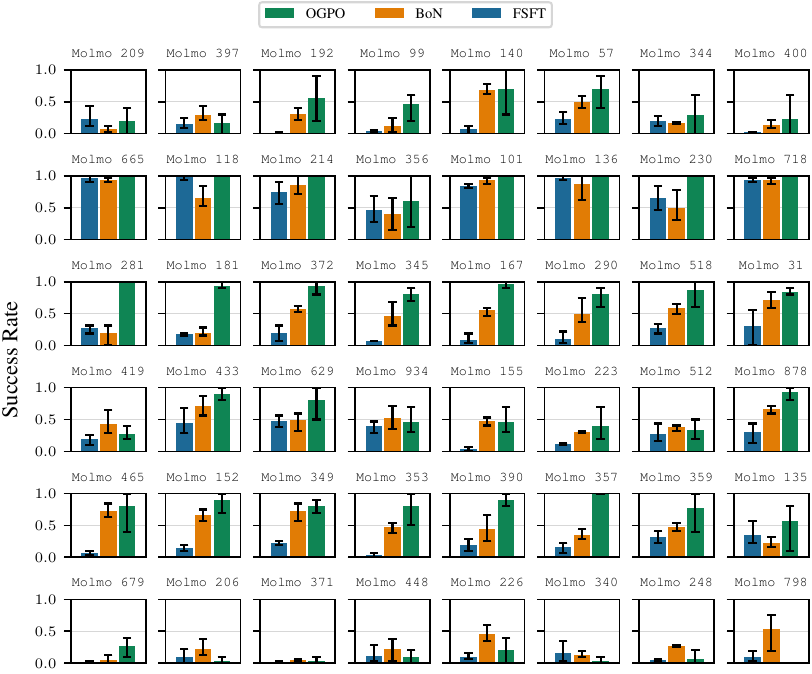}
    \caption{\textbf{\texttt{MolmoSpaces} (64 tasks): unseen objects.} Final success
    rate on unseen objects. Bars show the mean over three seeds with 95\% stratified
    bootstrap CIs.}
    \label{fig:eval_molmo_64_objects}
\end{figure}

\newpage
\subsubsection{\texttt{MolmoSpaces} (144 Tasks) In-Distribution}
\begin{figure}[h]
    \centering
    \includegraphics[width=1.0\linewidth]{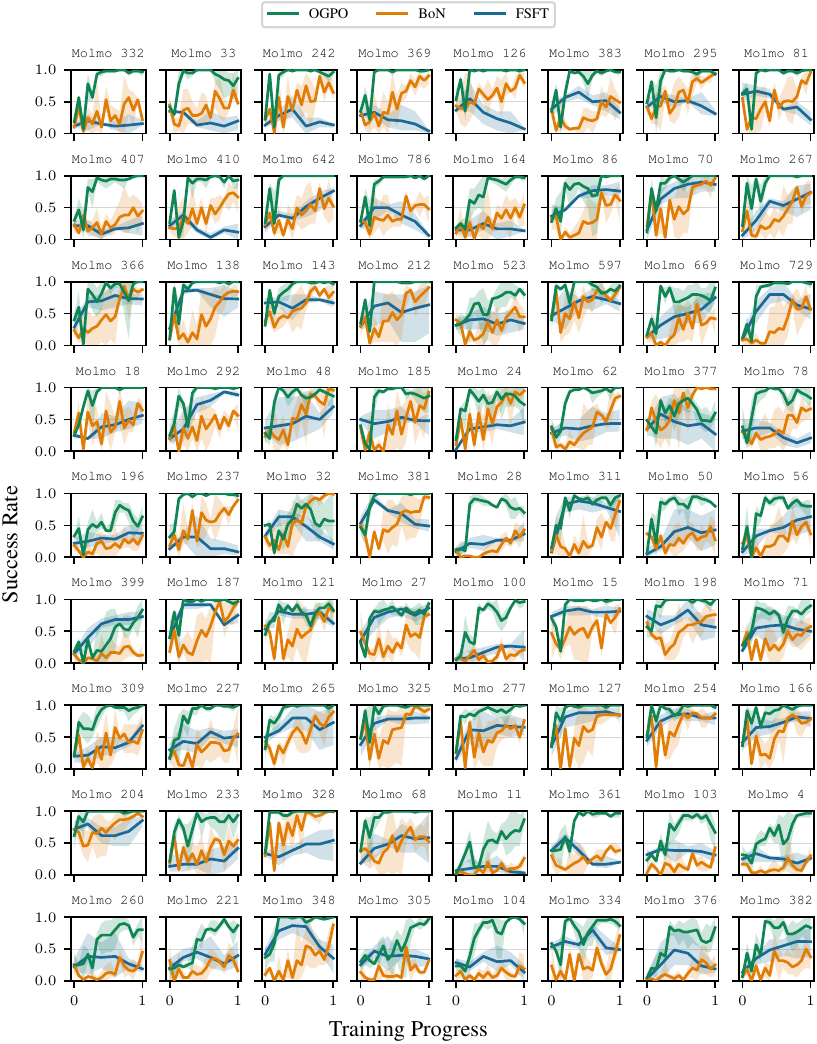}
    \caption{\textbf{\texttt{MolmoSpaces} (144 tasks, Part 1): training tasks.} Success rate on
    the training tasks as a function of normalized training progress, where $1$
    denotes the end of post-training. Lines show the mean over three seeds. Shaded
    regions denote 95\% stratified bootstrap CIs.}
    \label{fig:curves_molmo_144_part_1}
\end{figure}
\newpage
\begin{figure}[h]
    \centering
    \includegraphics[width=1.0\linewidth]{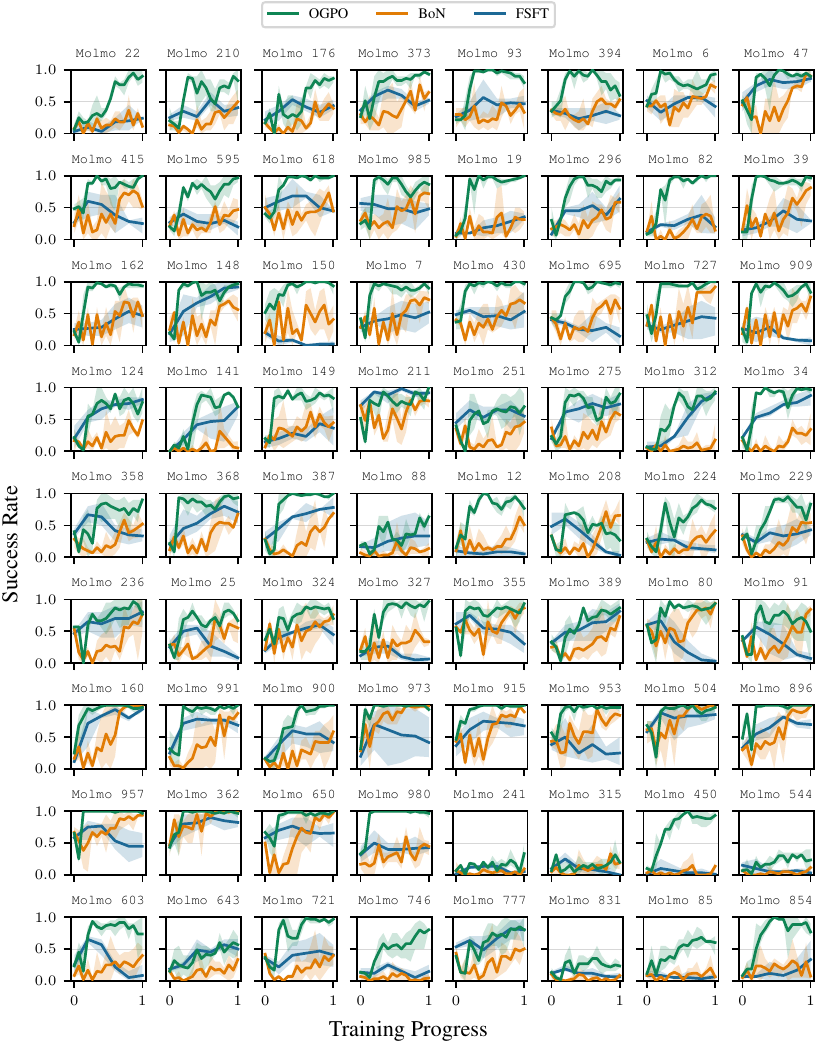}
    \caption{\textbf{\texttt{MolmoSpaces} (144 tasks, Part 2): training tasks.} Success rate on
    the training tasks as a function of normalized training progress, where $1$
    denotes the end of post-training. Lines show the mean over three seeds. Shaded
    regions denote 95\% stratified bootstrap CIs.}
    \label{fig:curves_molmo_144_part_2}
\end{figure}

\newpage
\subsubsection{\texttt{MolmoSpaces} (144 Tasks) Unseen Scenes}
\begin{figure}[h]
    \centering
    \includegraphics[width=\linewidth]{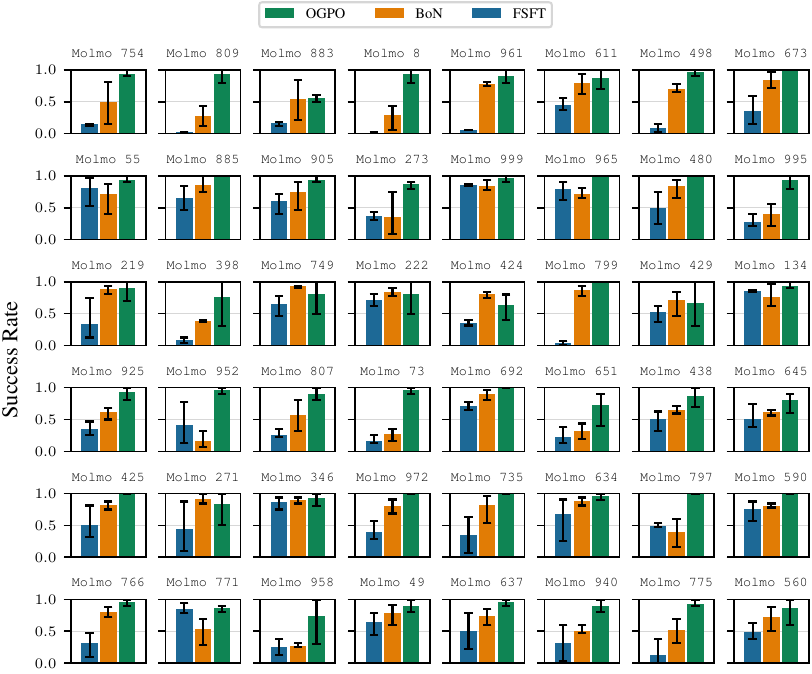}
    \caption{\textbf{\texttt{MolmoSpaces} (144 tasks, Part 1): unseen scenes.} Final success
    rate on unseen scenes. Bars show the mean over three seeds with 95\% stratified
    bootstrap CIs.}
    \label{fig:eval_molmo_144_scenes_part1}
\end{figure}

\newpage
\begin{figure}[h]
    \centering
    \includegraphics[width=\linewidth]{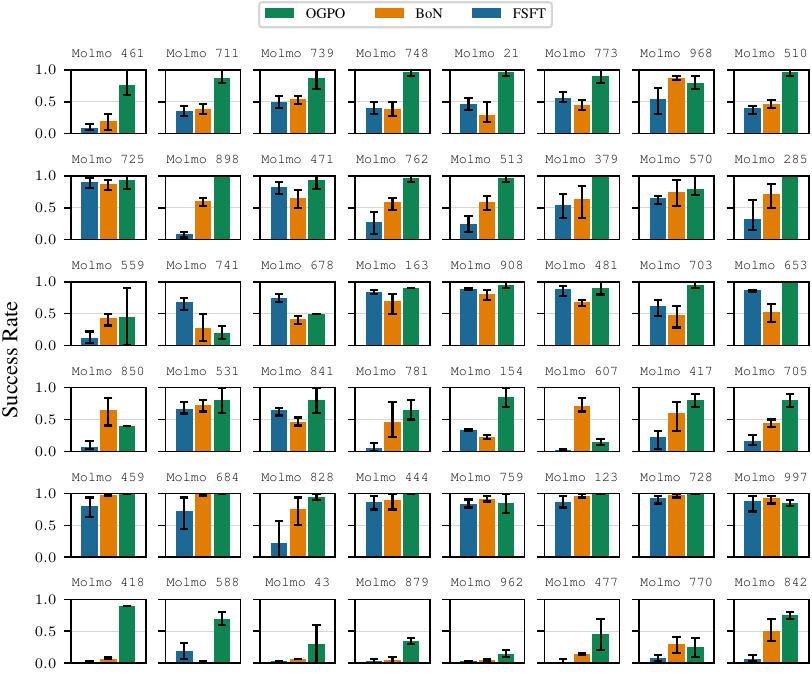}
    \caption{\textbf{\texttt{MolmoSpaces} (144 tasks, Part 2): unseen scenes.} Final success
    rate on unseen scenes. Bars show the mean over three seeds with 95\% stratified
    bootstrap CIs.}
    \label{fig:eval_molmo_144_scenes_part2}
\end{figure}
\newpage
\subsubsection{\texttt{MolmoSpaces} (144 Tasks) Unseen Objects}
\begin{figure}[h]
    \centering
    \includegraphics[width=\linewidth]{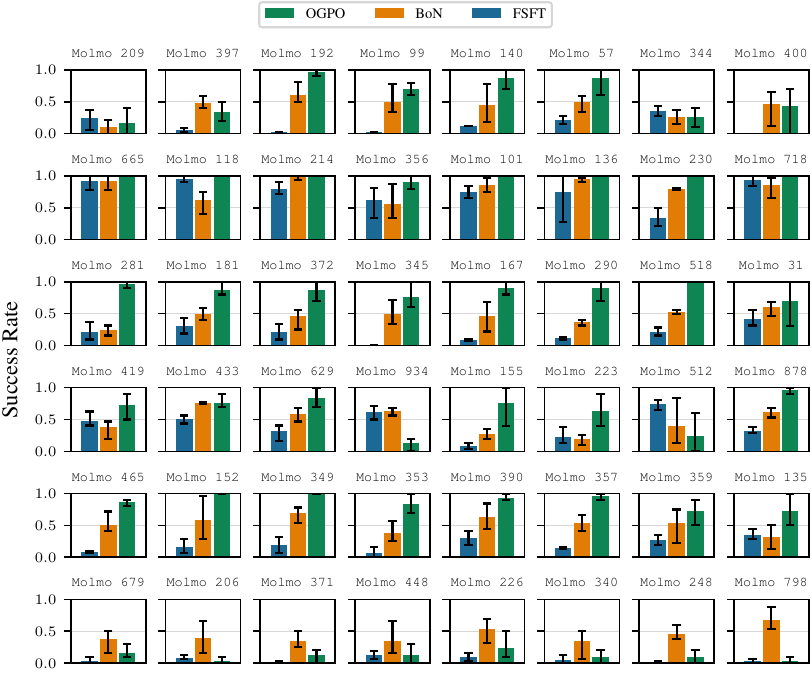}
    \caption{\textbf{\texttt{MolmoSpaces} (144 tasks): unseen objects.} Final success
    rate on unseen objects. Bars show the mean over three seeds with 95\% stratified
    bootstrap CIs.}
    \label{fig:eval_molmo_144_objects}
\end{figure}

\newpage
\subsection{Main Results — \texttt{LIBERO}}
\subsubsection{\texttt{LIBERO} (1 Task) In-Distribution}
\begin{figure}[h]
    \centering
    \includegraphics[width=0.54\linewidth]{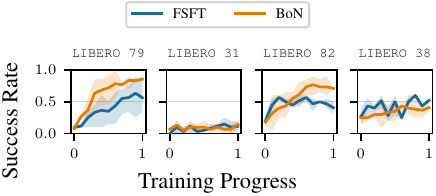}
    \caption{\textbf{\texttt{LIBERO} (1 task): training tasks.} Success rate on
    the training tasks as a function of normalized training progress, where $1$
    denotes the end of post-training. Lines show the mean over three seeds. Shaded
    regions denote 95\% stratified bootstrap CIs.}
    \label{fig:curves_libero_1}
\end{figure}

\subsubsection{\texttt{LIBERO} (1 Task) Unseen Tasks}
\begin{figure}[h]
    \centering
    \includegraphics[width=\linewidth]{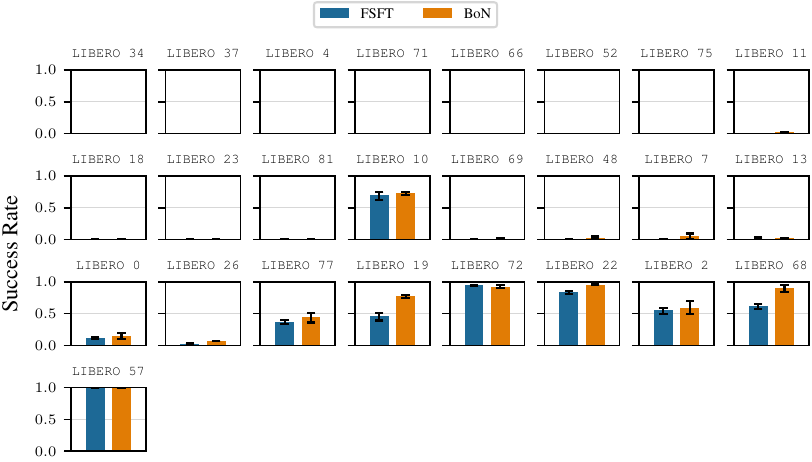}
    \caption{\textbf{\texttt{LIBERO} (1 task): unseen tasks.} Final success
    rate on unseen tasks. Bars show the mean over three seeds with 95\% stratified
    bootstrap CIs.}
    \label{fig:eval_libero_1_tasks}
\end{figure}

\newpage
\subsubsection{\texttt{LIBERO} (4 Tasks) In-Distribution}
\begin{figure}[h]
    \centering
    \includegraphics[width=0.54\linewidth]{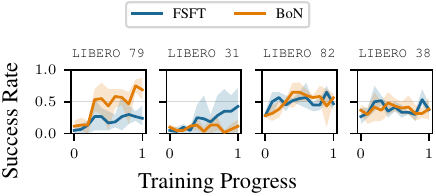}
    \caption{\textbf{\texttt{LIBERO} (4 tasks): training tasks.} Success rate on
    the training tasks as a function of normalized training progress, where $1$
    denotes the end of post-training. Lines show the mean over three seeds. Shaded
    regions denote 95\% stratified bootstrap CIs.}
    \label{fig:curves_libero_4}
\end{figure}

\subsubsection{\texttt{LIBERO} (4 Tasks) Unseen Tasks}
\begin{figure}[h]
    \centering
    \includegraphics[width=\linewidth]{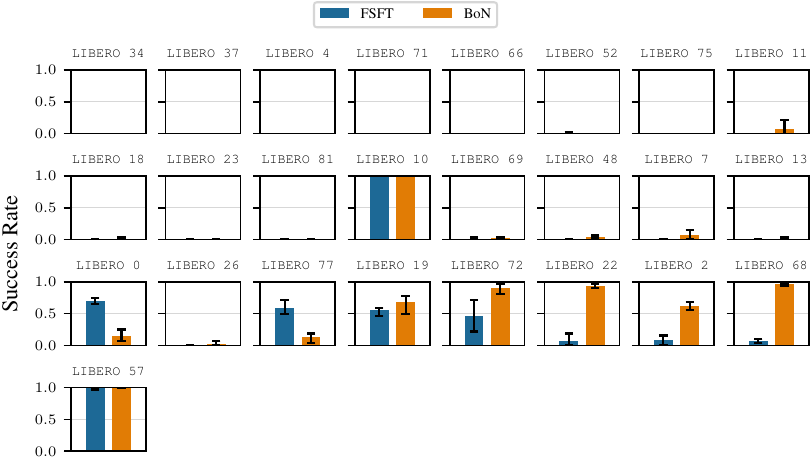}
    \caption{\textbf{\texttt{LIBERO} (4 tasks): unseen tasks.} Final success
    rate on unseen tasks. Bars show the mean over three seeds with 95\% stratified
    bootstrap CIs.}
    \label{fig:eval_libero_4_tasks}
\end{figure}

\newpage
\subsubsection{\texttt{LIBERO} (16 Tasks) In-Distribution}
\begin{figure}[h]
    \centering
    \includegraphics[width=\linewidth]{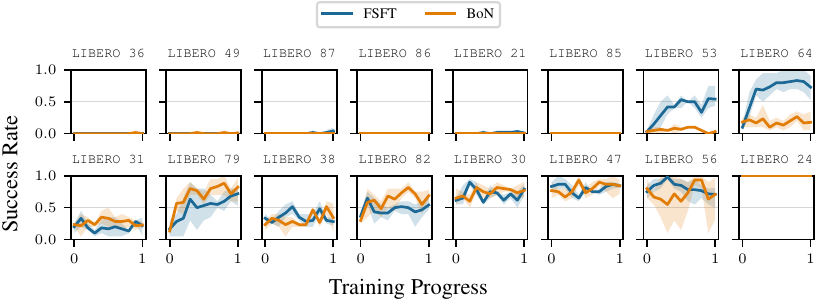}
    \caption{\textbf{\texttt{LIBERO} (16 tasks): training tasks.} Success rate on
    the training tasks as a function of normalized training progress, where $1$
    denotes the end of post-training. Lines show the mean over three seeds. Shaded
    regions denote 95\% stratified bootstrap CIs.}
    \label{fig:curves_libero_16}
\end{figure}

\subsubsection{\texttt{LIBERO} (16 Tasks) Unseen Tasks}
\begin{figure}[h]
    \centering
    \includegraphics[width=\linewidth]{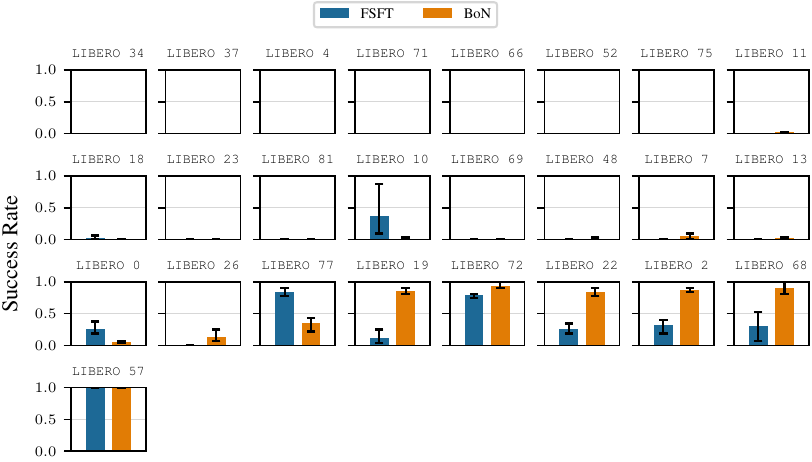}
    \caption{\textbf{\texttt{LIBERO} (16 tasks): unseen tasks.} Final success
    rate on unseen tasks. Bars show the mean over three seeds with 95\% stratified
    bootstrap CIs.}
    \label{fig:eval_libero_16_tasks}
\end{figure}

\newpage
\subsubsection{\texttt{LIBERO} (64 Tasks) In-Distribution}
\begin{figure}[h]
    \centering
    \includegraphics[width=1.0\linewidth]{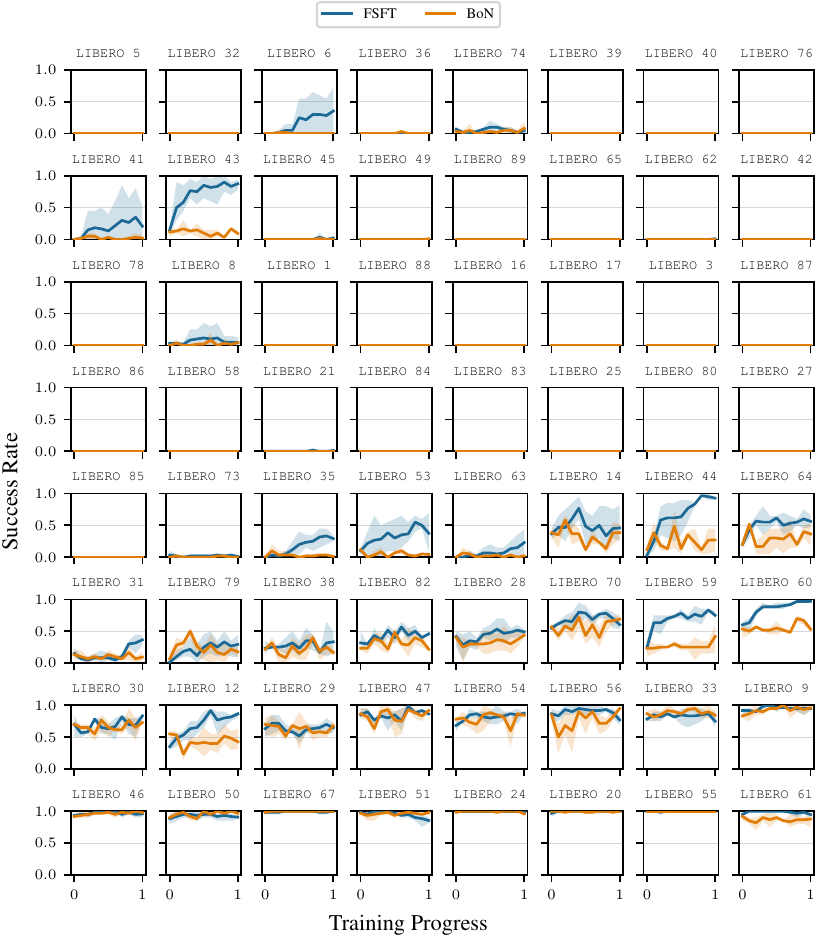}
    \caption{\textbf{\texttt{LIBERO} (64 tasks): training tasks.} Success rate on
    the training tasks as a function of normalized training progress, where $1$
    denotes the end of post-training. Lines show the mean over three seeds. Shaded
    regions denote 95\% stratified bootstrap CIs.}
    \label{fig:curves_libero_64}
\end{figure}

\newpage
\subsubsection{\texttt{LIBERO} (64 Tasks) Unseen Tasks}
\begin{figure}[h]
    \centering
    \includegraphics[width=1.0\linewidth]{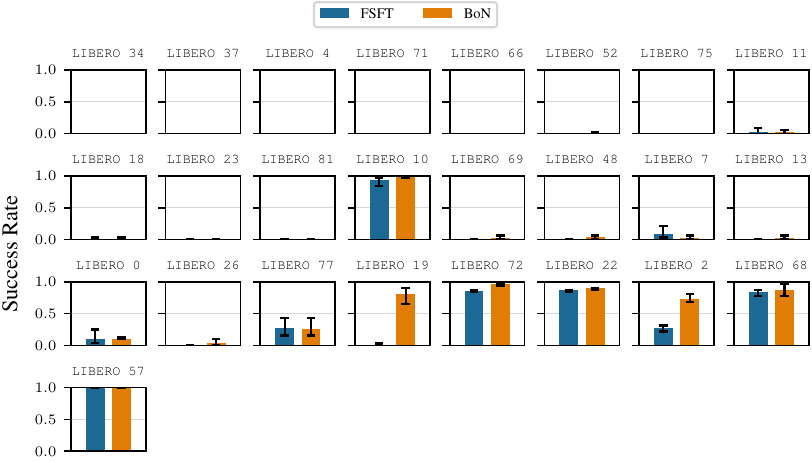}
    \caption{\textbf{\texttt{LIBERO} (64 tasks): unseen tasks.} Final success
    rate on unseen tasks. Bars show the mean over three seeds with 95\% stratified
    bootstrap CIs.}
    \label{fig:eval_libero_64_tasks}
\end{figure}

\newpage
\subsection{Recipes for Policy Extraction Results}
\label{app:actor_critic}

\subsubsection{\texttt{MolmoSpaces} (16 Tasks)}
\begin{figure}[h]
    \centering
    \includegraphics[width=1.0\linewidth]{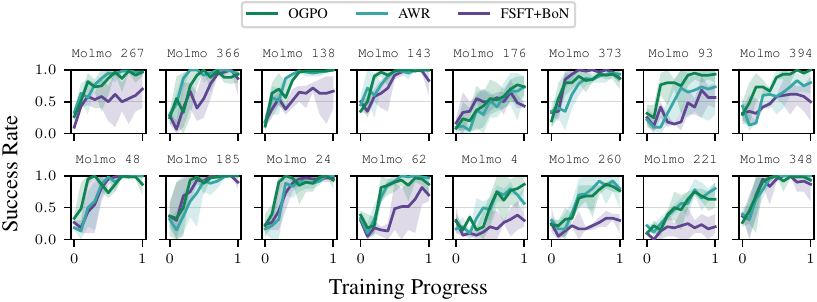}
    \caption{\textbf{\texttt{MolmoSpaces} (16 tasks): recipes for policy extraction.} Success rate on
    the training tasks as a function of normalized training progress, where $1$
    denotes the end of post-training. Lines show the mean over three seeds. Shaded
    regions denote 95\% stratified bootstrap CIs.}
    \label{fig:curves_policy_extraction}
\end{figure}

\end{document}